\documentclass[final,3p,times]{elsarticle}
\usepackage{mathtools}
\usepackage{amssymb}
\usepackage{stmaryrd}
\usepackage{color}
\usepackage{pdflscape}
\usepackage{epsfig}

\usepackage{epstopdf}
\usepackage{multirow}
\usepackage{hhline}
\usepackage{txfonts}
\usepackage{mathrsfs}
\usepackage{color}
\usepackage{algorithmic}

\usepackage{mathrsfs}
\usepackage{sidecap}
\usepackage{graphicx, array}
\usepackage{makecell}
\usepackage{mathabx}

\usepackage{lineno,hyperref}
\modulolinenumbers[5]

\newtheorem{thm}{Theorem}[section]

\newtheorem{rem}[thm]{Remark}
\newtheorem{exm}[thm]{Example}

\numberwithin{equation}{section}

\allowdisplaybreaks

\journal{Journal of Computer Methods in Applied Mechanics and Engineering}

\begin{document}

\begin{frontmatter}

\title{\lowercase{\textit{hp}}-VPINNs: Variational Physics-Informed Neural Networks \\ With Domain Decomposition}

\author[Brownaddress]{Ehsan Kharazmi\corref{mycorrespondingauthor}}
\cortext[mycorrespondingauthor]{Corresponding author}
\ead{ehsan\_kharazmi@brown.edu}

\author[WPIaddress]{Zhongqiang Zhang}
\author[Brownaddress,PNNLaddress]{George Em Karniadakis}

\address[Brownaddress]{Division of Applied Mathematics, Brown University, 170 Hope St, Providence, RI 02906, USA.}
\address[WPIaddress]{Department of Mathematical Sciences,	
		Worcester Polytechnic Institute, 100 Institute Rd, Worcester , MA 01609, USA.}
\address[PNNLaddress]{Pacific Northwest National Laboratory, Richland, WA 99354, USA.}



\begin{abstract}
We formulate a general framework for \textit{hp}-variational physics-informed neural networks (\textit{hp}-VPINNs) based on the nonlinear approximation of shallow and deep neural networks and \textit{hp}-refinement via domain decomposition and projection onto space of high-order polynomials. The {\em trial space} is the space of neural network, which is defined globally over the whole computational domain, while the {\em test space} contains the piecewise polynomials. Specifically in this study, the \textit{hp}-refinement corresponds to a \textit{global approximation with local learning} algorithm that can efficiently localize the network parameter optimization. We demonstrate the advantages of \textit{hp}-VPINNs in accuracy and training cost for several numerical examples of function approximation and solving differential equations.
\end{abstract}


\end{frontmatter}

\thispagestyle{plain}
%
\section{Introduction}
\label{Sec: Introduction NN}
%
Neural networks (NN) have gained a lot of attention more recently in solving differential equations, see e.g. \cite{BerNys18,weinan2018deep,raissi2019physics,mao2020physics,khoo2019solving,SamANe20,liao2019deep,LiTTL19}. They offer a nonlinear approximant via the composition of hidden layers in a variety of network structures and activation functions, and their universal approximation properties provide an alternative approach for solving differential equations. In general, the nonlinear approximation \cite{devore1998nonlinear,devore2009nonlinear} extends the approximants to reside to a nonlinear space and does not limit the approximation to linear spaces; it contains different approaches such as wavelet analysis \cite{daubechies1992ten}, dictionary learning \cite{tariyal2016greedy}, adaptive pursuit and compressed sensing \cite{davis1994adaptive, ohlsson2013nonlinear, candes2006compressive, candes2008introduction}, adaptive splines \cite{devore1998nonlinear}, radial basis functions \cite{devore2010approximation}, Gaussian kernels \cite{hangelbroek2010nonlinear}, and neural networks \cite{mhaskar1992approximation, mhaskar2019function, daubechies2019nonlinear}. 

Due to the nature of nonlinear approximations of neural  networks, solving differential equations using NNs is formulated as optimization problems where it is crucial to design appropriate loss functions to optimize the quantities of interests. Based on the method of variational/weighted residuals \cite{FinScr66}, several solvers have been developed, such as deep Galerkin method (DGM) \cite{sirignano2018dgm} based on the least squares,  physics-informed neural networks (PINNs) \cite{raissi2019physics,yang2018physics} based on the collocation methods, and variational physics-informed neural networks (VPINNs) \cite{kharazmi2019variational,khodayi2019varnet} based on the Galerkin method. Along this path, 
we develop a method which is called  \textit{\textit{hp}-Variational Physics Informed Neural Networks} (\textit{hp}-VPINNs) 
based on the sub-domain Petrov-Galerkin method. The  neural network still serves as the trial space but, compared to all the aforementioned works, the sub-domain Petrov-Galerkin methods allow \textit{hp}-refinement via domain decomposition as \textit{h}-refinement and projection onto space of high order polynomials as \textit{p}-refinement.

In this work, we consider the following problem 
\begin{align}
\label{Eq: Strong form }
\mathcal{L}^\textbf{q} u(\textbf{x},t) & = f(\textbf{x},t), \quad (\textbf{x},t)\in \Omega \times (0,T],
\\
\label{Eq: Strong form BC}
u(\textbf{x}, t)   & = h(\textbf{x}, t),  \quad (\textbf{x},t)\in \partial \Omega \times [0,T],
\quad u(\textbf{x}, 0)   = g(\textbf{x}), \quad \textbf{x}\in \Omega,
\end{align}
where the bounded domain $\Omega \subset \mathbb{R}^d$ with boundaries $\partial \Omega$, $T>0$, and $u(\textbf{x}, t): \Omega \times [0,T] \rightarrow \mathbb{R}$ describes the underlying physical phenomena modeled by the above governing equation. The operator $\mathcal{L}^\textbf{q}$ is usually comprised of the identity and the differential operators with some parameters $\textbf{q}$. We assume that $ \tilde{u}(\textbf{x}, t;\textbf{W} , \textbf{b})$ is a NN approximation (trial solution) of $u(\textbf{x}, t)$ in \eqref{Eq: Strong form }-\eqref{Eq: Strong form BC}. Specifically, the NN is comprised of $\ell$ hidden layers with $\mathcal{N}_i$ neurons in each layer and activation function $\sigma$ that takes the following form
\begin{align}
\label{Eq: deep NN}
%
u_{NN}(\textbf{x},t; \textbf{W} , \textbf{b}) = g \circ T^{(\ell)} \circ T^{(\ell-1)} \circ \cdots \circ T^{(1)}(\textbf{x}) .
\end{align} 
In the output layer, the linear mapping is $g: \mathbb{R}^{\mathcal{N}_\ell} \rightarrow \mathbb{R}$, and in each hidden layer $i=1,2,\cdots,\ell$, the nonlinear mapping is $T^{(i)}(\cdot) = \sigma(\textbf{W}_i \times \cdot + \textbf{b}_i)$ with weights $\textbf{W}_i \in \mathbb{R}^{\mathcal{N}_i \times \mathcal{N}_{i-1}}$ and biases $\textbf{b}_i \in \mathbb{R}^{\mathcal{N}_i}$, where $\mathcal{N}_0 = d$ is the input dimension.
Then, we define the \textit{strong-form residual} $r(\tilde{u})$, the boundary residual $ r_b(\tilde{u})$, and the initial residual  $ r_0(\tilde{u}) $ as~
\begin{align}
\label{Eq: residue strong form}
%
r(\tilde{u}) 
&= \mathcal{L}^\textbf{q} \tilde{u} - f, \quad \forall (\textbf{x}, t) \in \,\,\,\Omega \times (0,T],
\\ \nonumber
r_b(\tilde{u}) 
&= \tilde{u} - h , \,\,\quad\quad \forall (\textbf{x}, t) \in \partial\Omega \times [0,T],
\\ \nonumber
r_0(\tilde{u}) 
&= \tilde{u} - g , \quad\quad\,\, \forall (\textbf{x}, t) \in \,\,\, \Omega \times \{  t=0 \}.
\end{align}
The residuals are measures to the extent to which the approximation $\tilde{u}$ satisfies the equations \eqref{Eq: Strong form }-\eqref{Eq: Strong form BC}. Ideally, the exact solution is recovered when all the residuals are identically zero. The weighted integrals of the residuals are obtained by projecting them onto a properly chosen space of test (weighting) functions $V$ and then set to zero; this leads to the variational form of the problem. Specifically, we choose some test functions $v_j$ such that 
\begin{eqnarray}\label{eqs:variational-residuals}
\mathcal{R}_j(\tilde{u}) &=&
\int_{\Omega\times (0,T]}  r(\tilde{u}) v_j \,dx\,dt 
=0,
\\
\mathcal{R}_{b,j}(\tilde{u}) &=&
\int_{\partial \Omega\times (0,T]} r_b(\tilde{u}) v_j \,dx\,dt 
=0,
\\
\mathcal{R}_{0,j}(\tilde{u}) &=&
\int_{\Omega } r_0(\tilde{u}) v_j \,dx  
=0.
\end{eqnarray}
To solve the nonlinear system resulting from these equations, we formulate it as the following minimization problem:
\begin{equation} 
\min_{\textbf{W},\textbf{b}} \mathcal{J}(\tilde{u},v),
\end{equation}
where
\begin{equation}
\label{Eq: energy functional}
\mathcal{J}(\tilde{u},v) 
= 
w \, \sum_{j=1}^{N_r} \mathcal{R}_{j}^2(\tilde{u}) 
+ 
w_{b} \, \sum_{j=1}^{N_b} \mathcal{R}^2_{b,j}(\tilde{u}) 
+  
w_{0} \, \sum_{j=1}^{N_0} \mathcal{R}^2_{0,j}(\tilde{u}) .
\end{equation}
The parameters $\{w, w_b, w_0 \}$ denote the weight coefficients in the loss function. They may be user-specified or tuned manually or automatically, e.g., in practice based on the numerical experiment in each problem; their optimal bound, however, is still an open problem in the literature \cite{wang2020understanding}.

\begin{table}[!ht]
\center
\caption{ \scriptsize{ \label{Table: MWR} Various numerical methods based on different approximation and test functions. }}
%
\scalebox{0.8}{
\begin{tabular}
{| c || c | @{\hspace*{.30in}}c@{\hspace*{.30in}} |} 
\multicolumn{1}{c}{\textbf{test function}}  & \multicolumn{2}{c}{\textbf{trial function}} \\
\hline
$v$ & $\tilde{u}$ & $\tilde{u}=$DNNs \\ \Xhline{3\arrayrulewidth}
delta Dirac & collocation  & PINNs \cite{raissi2019physics}\\ \hline
$v =  r(\tilde{u})$ & least square & DGM \cite{sirignano2018dgm}    \\ \hline
polynomials (global) & Petrov-Galerkin   & VPINNs \cite{kharazmi2019variational}   \\ \hline
polynomials (piece wise) & Petrov-Galerkin   & VarNet \cite{khodayi2019varnet}  \\ \hline
nonoverlapping \eqref{Eq: subdomain} & sub-domain Petrov-Galerkin & \textit{hp}-VPINNs   \\ \hline
%
\end{tabular}
}
\end{table}

Different choices of trial function $\tilde{u}$ and test function $v_j$ in \eqref{Eq: energy functional} correspond to various numerical methods. Most of these methods are well established and analyzed in the literature when linear approximations are used. Here, we focus on the nonlinear approximation of $\tilde{u}$ and various choices of test functions, and briefly discuss some choices of test functions when the trial function is shallow/deep NN; see Table \ref{Table: MWR} for comparison.

The \textit{Dirac delta test function}s, $v(\textbf{x}, t) = \delta(\textbf{x} - \textbf{x}_r) \delta(t - t_r),$ correspond to the collocation method. These test functions project the residuals onto a finite set of collocation points, making the equation to be satisfied at these points. The collocation formulation is used in 
\cite{BerNys18} and PINNs \cite{raissi2019physics}. The PINN formulation has been recently successfully employed in many physical problems such as discovering turbulence models from scattered/noisy measurements \cite{raissi2019deep}, high speed flows \cite{mao2020physics}, stochastic differential equation by generative adversarial networks \cite{yang2018physics}, fractional differential equations \cite{pang2019fpinns}, and adaptive activation functions \cite{jagtap2019adaptive,jagtap2019locallyadaptive}. Specifically, PINNs use the following functional  
\begin{align}
\label{Eq: loss PINN - 1}
L^{\mathfrak{s}} 
& = \frac{1}{N_r} \sum_{i = 1}^{N_r} \Big|r(\textbf{x}^i_{r}, t^i_r)\Big|^2 
+ \tau_b \, \frac{1}{N_b} \sum_{i = 1}^{N_b} \Big|r_b(\textbf{x}^i_{b}, t^i_b) \Big|^2
+ \tau_0 \, \frac{1}{N_0} \sum_{i = 1}^{N_0} \Big|r_0(\textbf{x}^i_{0}) \Big|^2,
\end{align}
where the residuals $r$, $r_b$ and $r_0$ are given in \eqref{Eq: residue strong form} and 
$\{(\textbf{x}^i_r, t^i_r )\}_{i=1}^{N_r}$, $\{(\textbf{x}^i_b, t^i_b)\}_{i=1}^{N_b}$ and $\{\textbf{x}^i_r\}_{i=1}^{N_0}$ are collocation points in their domains. We use the superscript $\mathfrak{s}$ to refer to the loss function associated with the strong-form of the residual. The \textit{deep Galerkin method} \cite{sirignano2018dgm, al2019applications} also employs the nonlinear approximation of NNs, however, it takes the test functions to be $v = r(\tilde{u})$ and essentially forms a least square method. Other formulations include the deep Ritz method \cite{weinan2018deep} and its extension to deep Nitsche method \cite{liao2019deep} with essential boundary conditions.

The variational formulation of PINNs, namely VPINNs \cite{kharazmi2019variational}, takes the nonlinear approximation of DNN as the approximation function. It projects the residuals onto the space of polynomials and thus  forms a Petrov-Galerkin method. It has been shown in \cite{kharazmi2019variational} that in VPINNs the \textit{variational residuals} can be obtained analytically for the case of shallow networks. Specifically, the VPINN formulation uses the following functional 
\begin{align}
\label{Eq: loss weak - 1}
&
L^{\mathfrak{v}} = \frac{1}{K} \sum_{j = 1}^{K} \Big| \mathcal{R}_j  \Big|^2 
+ \tau_b \, \frac{1}{N_b} \sum_{i = 1}^{N_b} \Big|r_b(\textbf{x}^i_{b}, t^i_b) \Big|^2
+ \tau_0 \, \frac{1}{N_0} \sum_{i = 1}^{N_0} \Big|r_0(\textbf{x}^i_{0}) \Big|^2,
\end{align}
that takes the test functions $v_j$ ($1\leq j\leq K$) from orthogonal polynomials. The superscript $\mathfrak{v}$ refers to the loss function associated with the variational-form of the residual. Other formulations based on the variational form of the problem have been developed. VarNet \cite{khodayi2019varnet, khodayi2018deep} takes the test functions to be the piece-wise linear shape functions of finite element method, D3M \cite{LiTTL19} formulation includes the reformulation of problem \eqref{Eq: Strong form }-\eqref{Eq: Strong form BC} into a system of first-order equations, and WAN \cite{zang2019weak,bao2020numerical} develops an adversarial framework by taking the test function to be a separate network.

In this paper, we develop the \textit{hp}-VPINNs by taking a different set of test functions, which are non-overlapping on each sub-domain of the domain; see the next section for more details. Our formulation has the flexibility to construct both local and global approximations with locally/globally defined test functions. We show the comparison with other methods that use NN in Table \ref{Table: GL approximation}. The flexibility of \textit{hp}-VPINN formulation allows us to accommodate the singularities, steep solution, and sharp changes; see Section \ref{Sec: VPINN Poisson examples 2d} for an example with corner singularity. Moreover, it allows us to adaptively select the orthogonal polynomials over the sub-domains with smooth solution. Thus, we may better balance the training cost by mainly focusing on optimizing the network parameters based on sub-domains with less regular solutions, leading to a localized learning process. For the integrals in \eqref{eqs:variational-residuals} and \eqref{Eq: energy functional}, we may perform integration by parts to reduce the regularity requirement before applying numerical quadrature rules for further discretization. For elliptic problems, we test the effects of performing no integration by parts or performing it once or twice; see Sections \ref{Sec: VPINN Poisson examples} and \ref{Sec: VPINN Poisson examples 2d}.

\begin{table}[!t]
	\center
	\caption{ \scriptsize{ \label{Table: GL approximation} Local/Global trial function versus local/global test function. }}
	%
	\scalebox{0.85}{
		\begin{tabular}
			{| l | c | c |}
			\hline
			&  local trial functions                          &   global trial functions           \\ \hline
			local test functions   & conservative VPINNs \cite{jagtap2020CPINN}  &   \textit{hp}-VPINNs       \\ \hline
			global test functions  & --   &    VPINNs \cite{kharazmi2019variational} 
			varNet \cite{khodayi2019varnet}, D3M \cite{LiTTL19}
			\\ \hline
			%
		\end{tabular}
	}
\end{table}

The rest of the paper is organized as follows. In Section \ref{Sec: vPINN}, we present the construction of \textit{hp}-VPINNs. Then, we examine the efficiency of the proposed method in approximating several functions in Section \ref{sec:vnn-for-approx}. In Sections \ref{Sec: VPINN Poisson examples} and \ref{Sec: VPINN Poisson examples 2d}, we present the details of calculations for elliptic problems in one- and two-dimensions. In Section \ref{Sec: VPINN Diff examples}, we show how the proposed method is modified to solve an inverse problem using a linear advection-diffusion equation.

\section{\textit{hp}-Variational Physics-Informed Neural Network (\textit{hp}-VPINN)}
\label{Sec: vPINN}
The \textit{hp}-VPINN formulation is based on the following localized  test functions, defined over nonoverlapping sub-domains $V_j, \,\, j = 1, 2, \cdots, N_{sd}$ of partition of the set $V$ ($\Omega\times(0,T]$ or $\Omega$ in this work). The test function defined on 
a subset $V_j \subset V$ reads
\begin{align}
\label{Eq: subdomain}
v_j = \begin{cases}
\bar{v} \neq 0, & \text{over } V_j,
\\
0, & \text{over } V_i^c,
\end{cases}
\quad V_j \cup V_j^c = V,
\end{align}
that leads to a sub-domain method. The non-vanishing test function $\bar{v}$ is a polynomial of order to be chosen in practice.

We define the \emph{elemental variational residual} as  
\begin{align}
\label{Eq: residue var form elemental}
\mathcal{R}^{(e)}
= \left( \mathcal{L}^\textbf{q} u_{NN}- f, v \right)_{\Omega_e \times \Gamma_e},
\end{align}
which is enforced for the admissible local test function within element $e$. Subsequently, we define the \emph{variational loss function} as
\begin{align}
\label{Eq: loss var elemental}
&
L^{\mathfrak{v}} = 
\sum_{e=1}^{N_{el}} \, \frac{1}{K^{(e)}} \sum_{k = 1}^{K^{(e)}} \Big| \mathcal{R}^{(e)}_k  \Big|^2
+ \tau_b \, \frac{1}{N_b} \sum_{i = 1}^{N_b} \Big|r_b(\textbf{x}^i_{b}, t^i_b) \Big|^2
+ \tau_0 \, \frac{1}{N_0} \sum_{i = 1}^{N_0} \Big|r_0(\textbf{x}^i_{0}) \Big|^2,
\end{align}
where $K^{(e)}$ is the total number of test functions in element $e$, the term $\mathcal{R}^{(e)}_k$ is the $k$-th entry of the corresponding tensor associated with element $e$, and $r_b$ and $r_0$ have the same form as in \eqref{Eq: residue strong form}. We refer to Sections \ref{Sec: VPINN Poisson examples} and \ref{Sec: VPINN Poisson examples 2d} for detailed derivation of \textit{hp}-VPINNs for one- and two-dimensional problems, respectively. 

The projection of strong-form residuals onto test functions additionally adds two major truncation and numerical integration errors into the existing approximation and generalization errors of DNNs. Increasing the number of test functions in order to eliminate the truncation error may further complicate the loss function and thus increase the chance of optimization failure in practice. In the case of shallow networks, the variational residual is obtained analytically \cite{kharazmi2019variational} that completely removes the numerical integration error. However, the compositional structure of hidden layers in DNNs makes it almost impossible to analytically compute the integrals in the variational loss function. Hence, we need to employ proper numerical integration techniques in the case of deep networks, which opens up new problems on developing and analyzing numerical integration methods for functions represented by DNNs. In this work, we adopt the Gauss quadrature rules. To avoid the curse of dimensionality in high-dimensional problems, we can employ numerical approaches such as quasi-Monte Carlo integration \cite{morokoff1995quasi} or sparse grid quadratures \cite{Smolyak63,novak1996high}.

\begin{rem}
Domain decomposition can provide the opportunity to assign the network optimization in each sub-domain to a specific computer node. We note that, however, as in the current \textit{hp}-VPINN formulation, even though we decompose the domain into several sub-domain, we still employ a single DNN to approximate the solution over the whole computational domain. In this setting, the parallelization may not be trivial as we only have a single loss function associated with DNN. 
\end{rem}

\begin{rem}
The activation function $\sigma$ has similar forms in \eqref{Eq: deep NN} for each neuron. However, it may also have different domain and image dimensionality based on the structure of network \cite{jagtap2019locallyadaptive, jagtap2019adaptive}. An adaptive basis viewpoint of DNNs is also given in \cite{cyr2019robust}.

\end{rem}

\section{Variational Neural Networks (VNNs) for Function Approximation}\label{sec:vnn-for-approx}
%
Let us consider the problem of approximating the target function $u(\textbf{x}): \Omega \rightarrow \mathbb{R}$ by $u_{NN}(\textbf{x})$.
We define the approximation residual as $r(\textbf{x}) = u(\textbf{x}) -  u_{NN}(\textbf{x})$. The setup can be viewed as follows. 
We let $w_{b}$ and $w_{0}$ be zero and thus define the corresponding loss function 
\begin{align}
\label{Eq: total var loss}
\text{L}^{\mathfrak{v}} 
= \sum_{e=1}^{N_{el}} \,
\frac{1}{K^{(e)}} \sum_{k = 1}^{K^{(e)}} \Big| \mathcal{R}^{(e)}_k \Big|^2,\quad \mathcal{R}_k^{(e)} = \int_{\Omega_e} \, \left(u_{NN}(\textbf{x}) - u(\textbf{x}) \right) \, v_k^{(e)}(\textbf{x}) \, d\Omega_e,
\end{align} 
where $K^{(e)}$ is the number of test functions employed in the element $e$. The VNN formulation with the loss function \eqref{Eq: total var loss} inherits all of the advantages \textit{hp}-VPINNs, i.e. \textit{hp}-refinement, employing different test functions in each element $e$, and the flexibility of adaptively choosing the proper number of test functions in each element $e$. 

We construct a fully connected network with $\ell$ hidden layers, each with $\mathcal{N} $ neurons and tanh activation functions (if not specifically mentioned otherwise). We use Legendre polynomials as test functions, i.e. $v_k(x) = P_{k-1}(x), \,\, k=1,2,\cdots,K$. We also use the Gauss quadrature rule with $Q$ quadrature points to compute the integrals. We consider two different approaches to approximate the function: \textit{i}) global or single element VNN where $N_{el}=1$ and $v_k^{(e)}$'s are smooth functions $v_k$'s, defined over the single element; \textit{ii}) local or elemental VNN where $N_{el}>1$ and $v_k^{(e)}$'s are locally defined but only one of them is non-zero; \textit{iii}) multi-elemental VNN where $N_{el}>1$ and $v_k^{(e)}$'s are locally defined and all are non-zero. In approach (\textit{ii}), the network captures the target function only on the restricted local elements, where $v_k^{(e)}$ is non-zero.

\begin{figure}[!ht]
	\center
	\begin{tabular}{l  l  l}
		\multicolumn{3}{l}{\textbf{A} \quad\qquad\qquad\qquad\qquad global (single element) VNN with global test functions} \\  [-1 pt] 
		\includegraphics[clip, trim=2.25cm 0cm 0.5cm 0cm, width=0.3\linewidth]{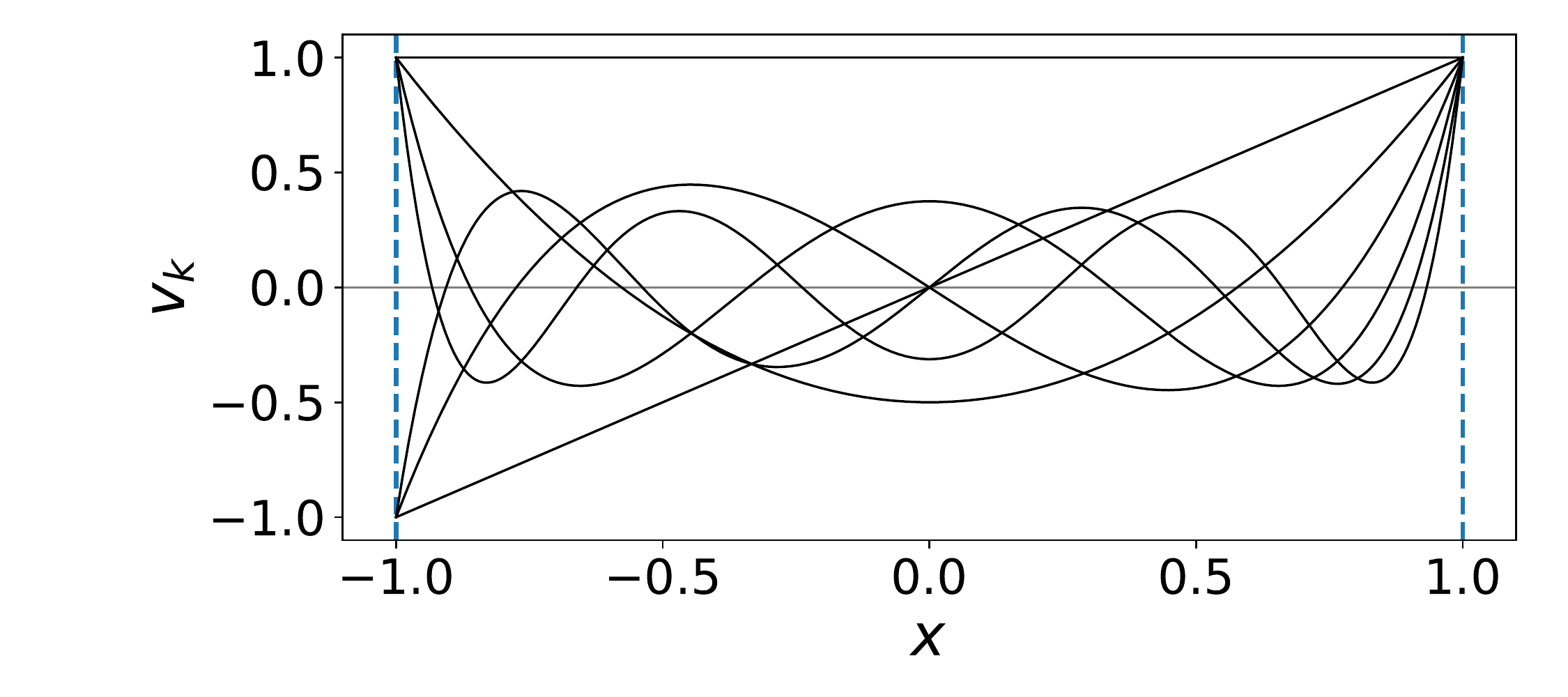}
		&
		\includegraphics[clip, trim=2.3cm 0cm 0.5cm 0cm, width=0.3\linewidth]{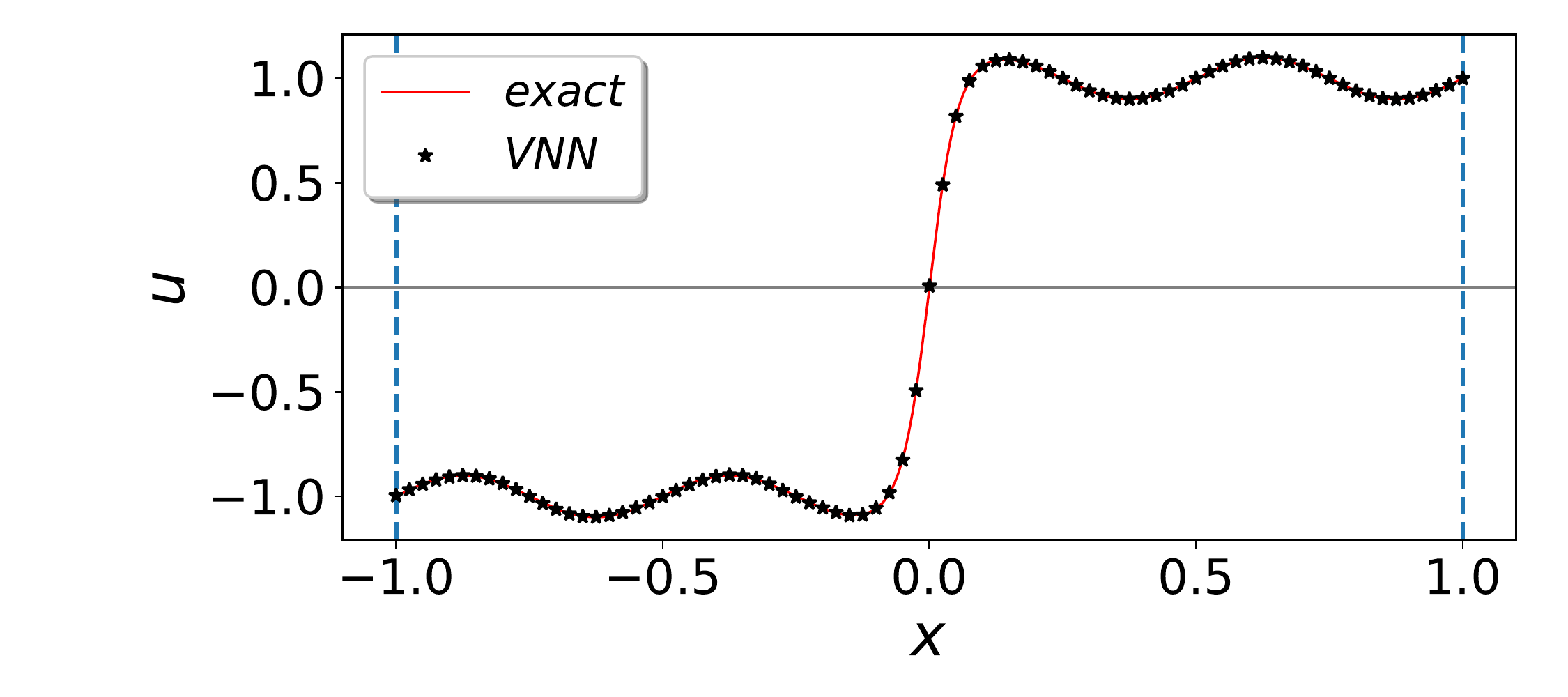}
		&
		\includegraphics[clip, trim=2cm 0cm 0.5cm 0cm, width=0.3\linewidth]{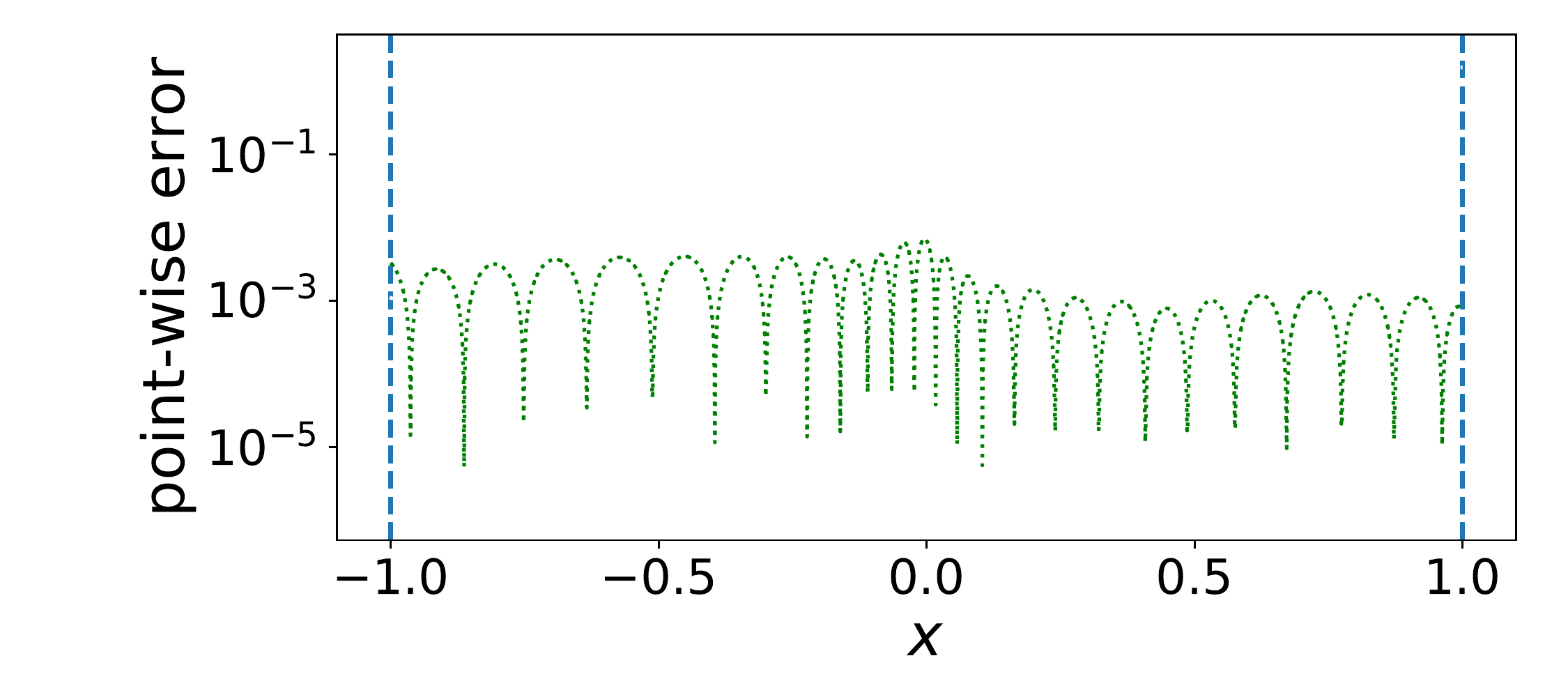}
		\\
		\multicolumn{3}{l}{\textbf{B} \quad\qquad\qquad\qquad\qquad\qquad local (elemental) VNN with local test functions} 
		\\  [-1 pt] 
		\includegraphics[clip, trim=2.25cm 0cm 0.5cm 0cm, width=0.3\textwidth]{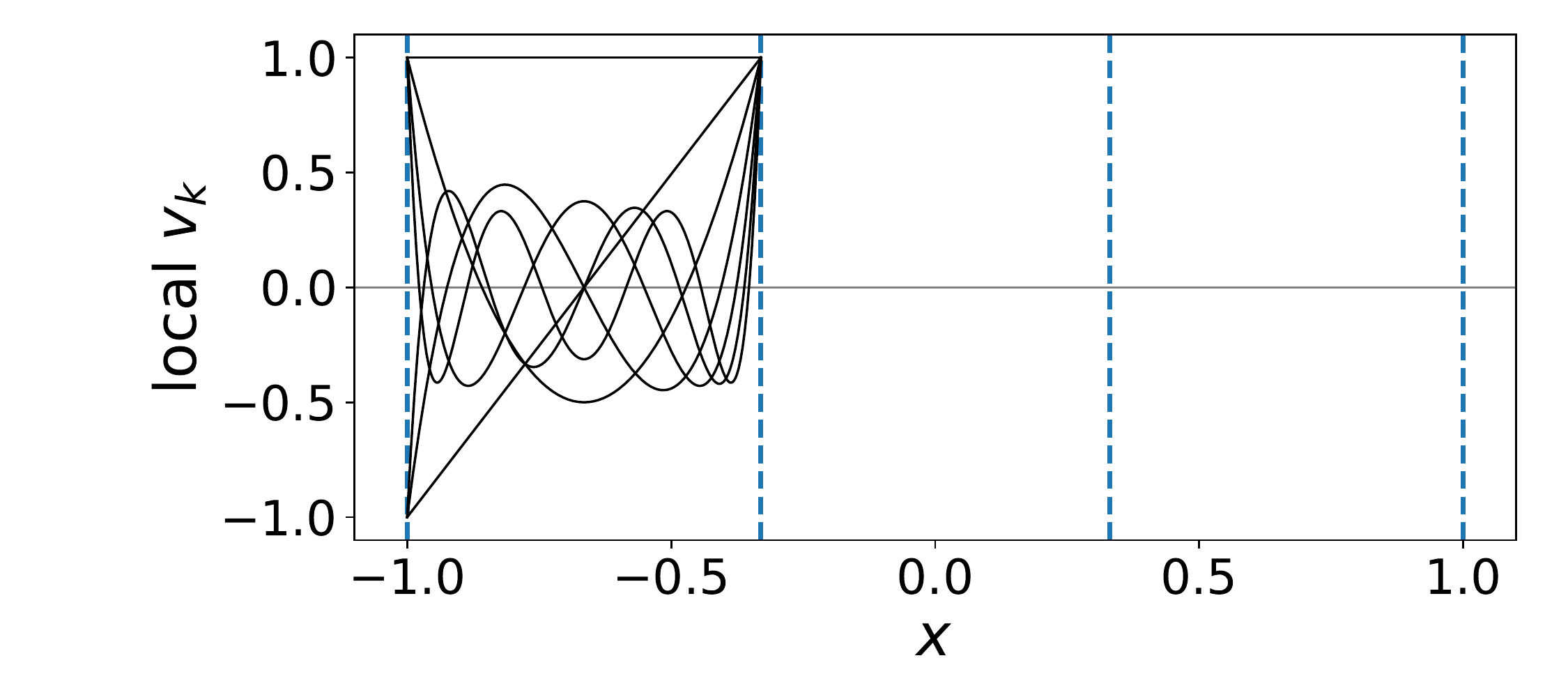}
		&
		\includegraphics[clip, trim=2.3cm 0cm 0.5cm 0cm, width=0.3\textwidth]{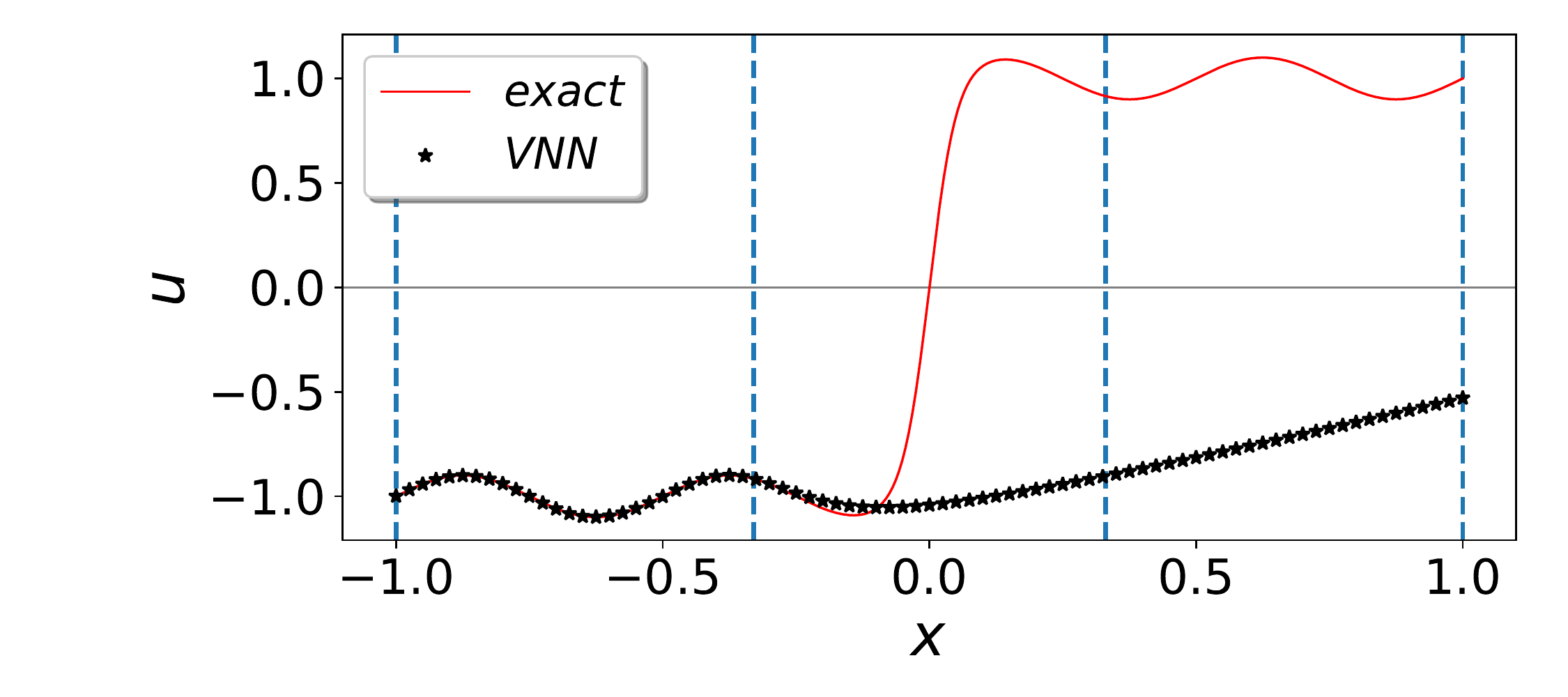}
		&
		\includegraphics[clip, trim=2cm 0cm 0.5cm 0cm, width=0.3\textwidth]{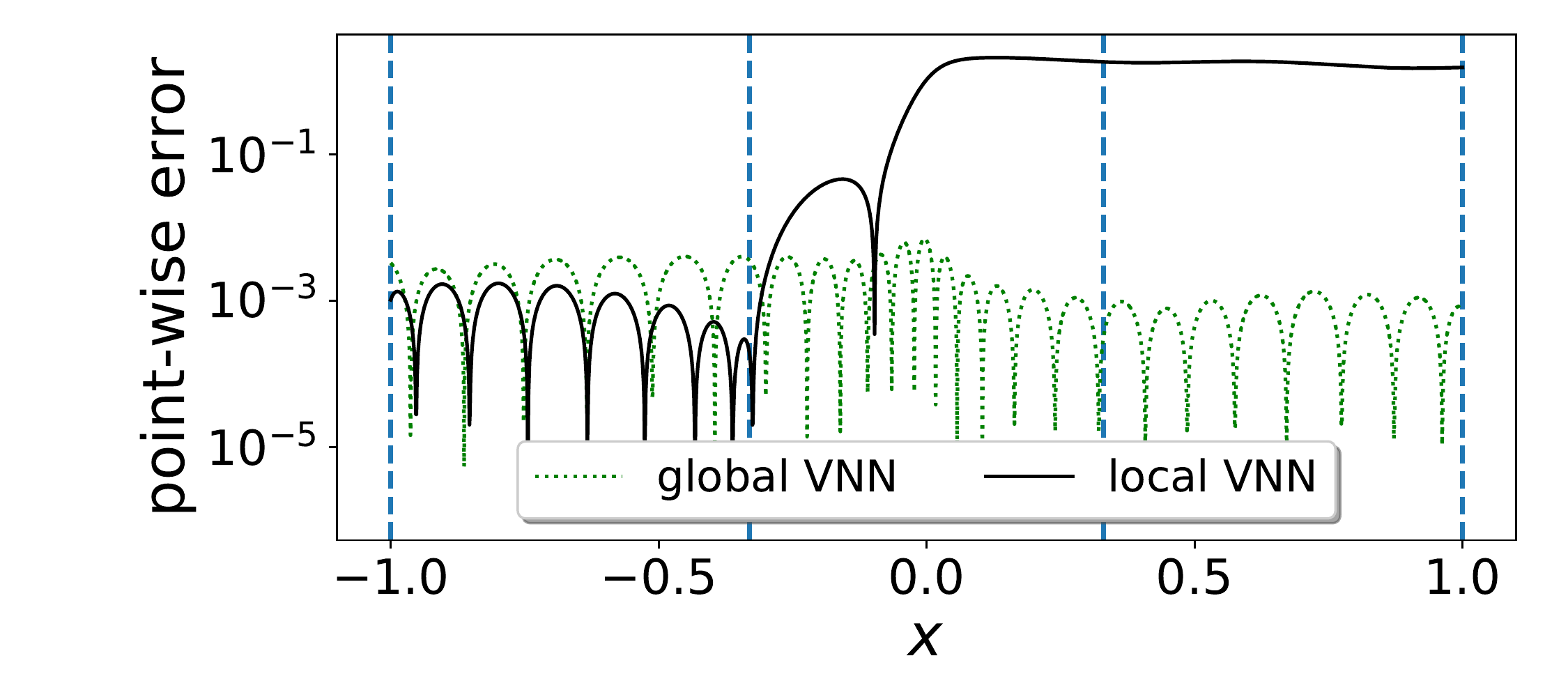}
		\\	[-11 pt]	
		\includegraphics[clip, trim=2.25cm 0cm 0.5cm 0cm, width=0.3\textwidth]{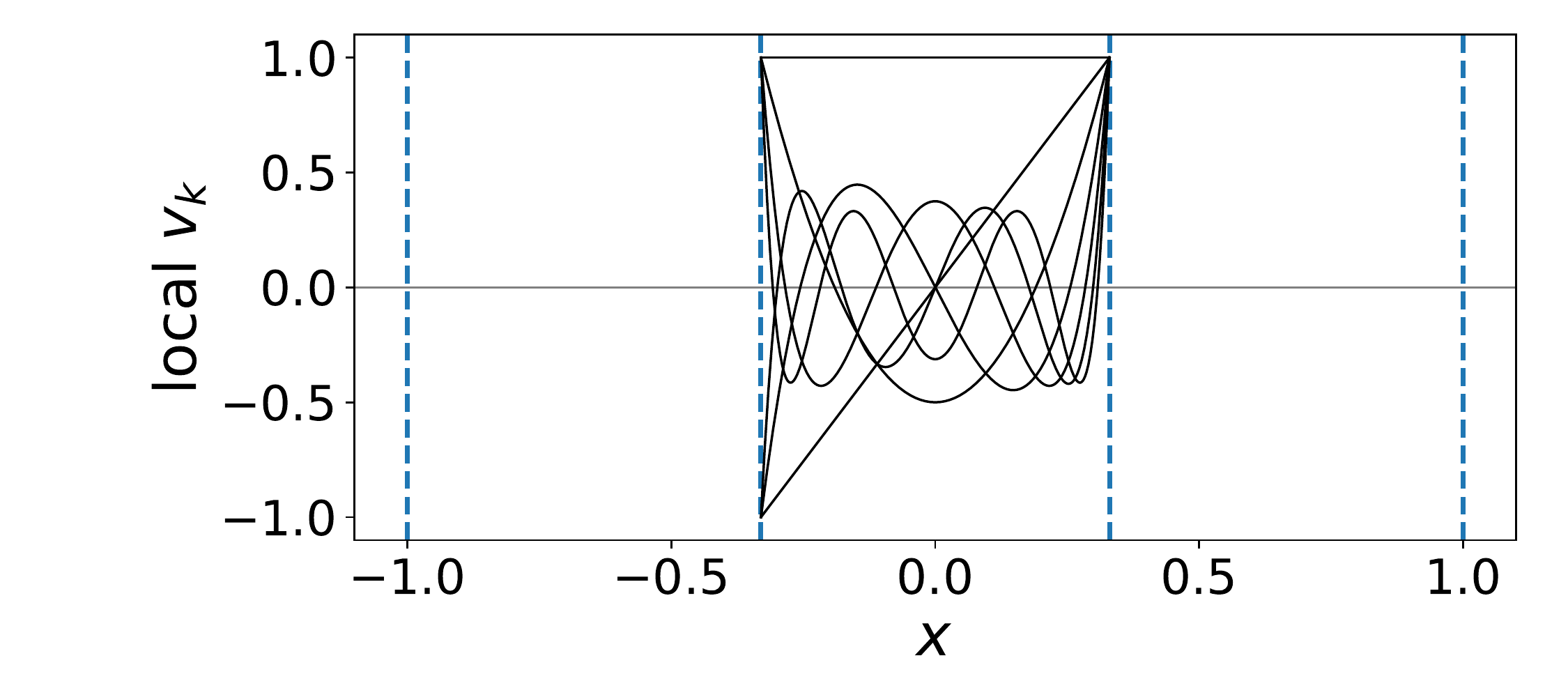}
		&
		\includegraphics[clip, trim=2.3cm 0cm 0.5cm 0cm, width=0.3\textwidth]{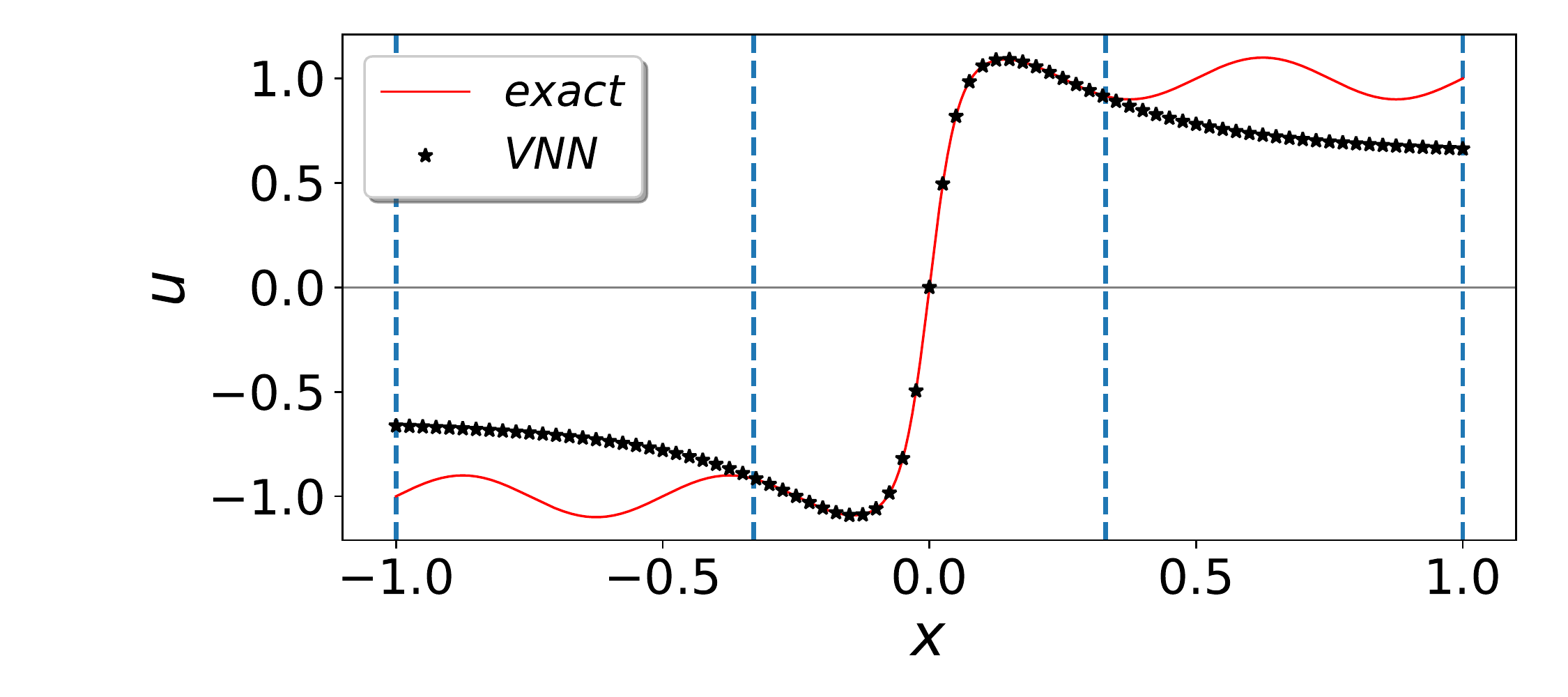}
		&
		\includegraphics[clip, trim=2cm 0cm 0.5cm 0cm, width=0.3\textwidth]{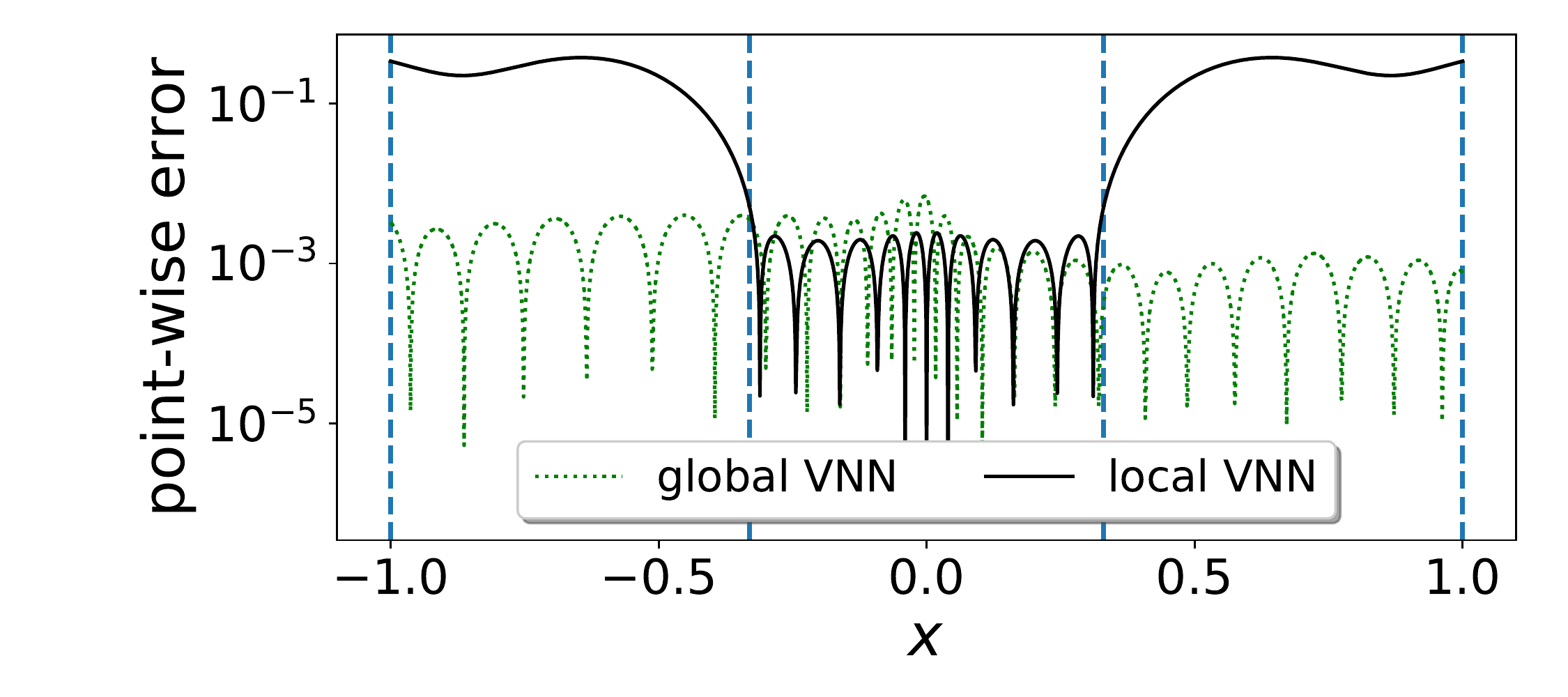}
		\\	[-11 pt]
		\includegraphics[clip, trim=2.25cm 0cm 0.5cm 0cm, width=0.3\textwidth]{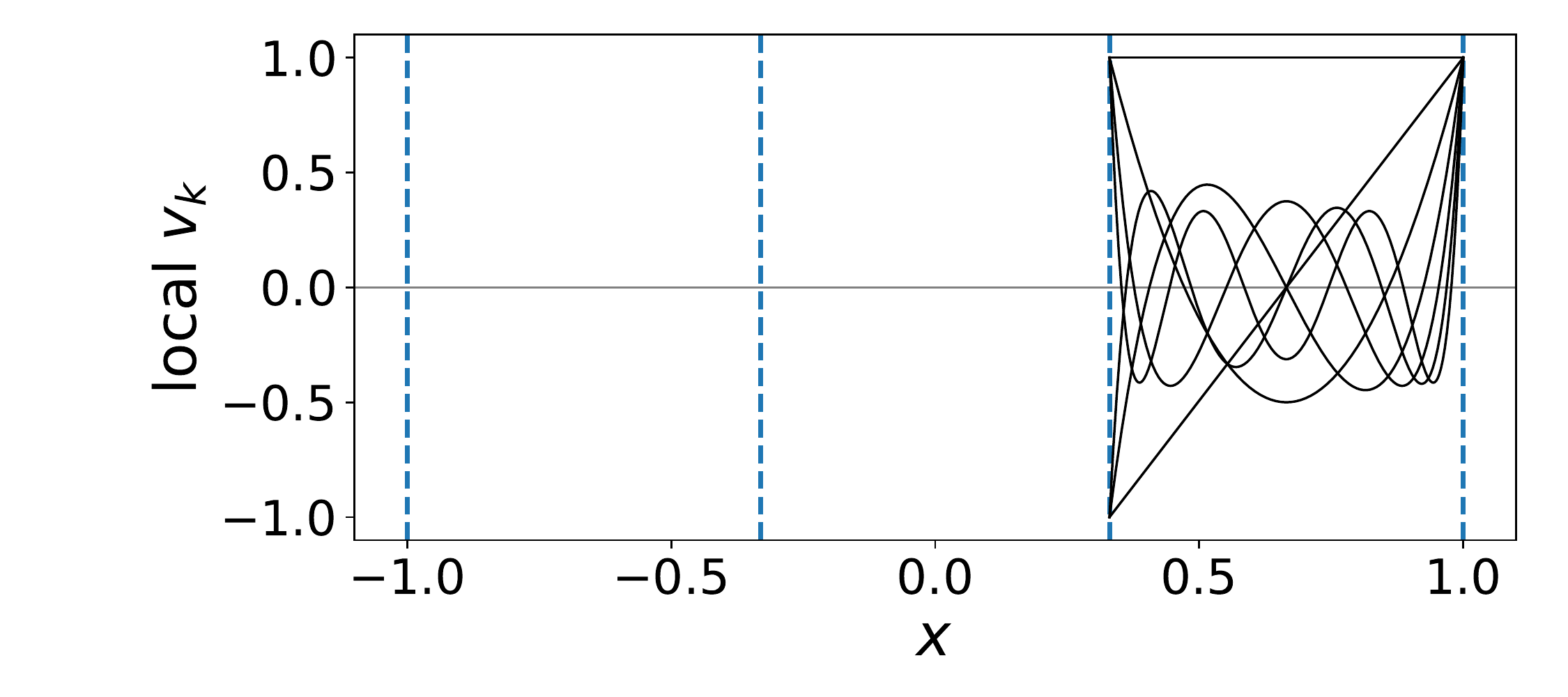}
		&
		\includegraphics[clip, trim=2.3cm 0cm 0.5cm 0cm, width=0.3\textwidth]{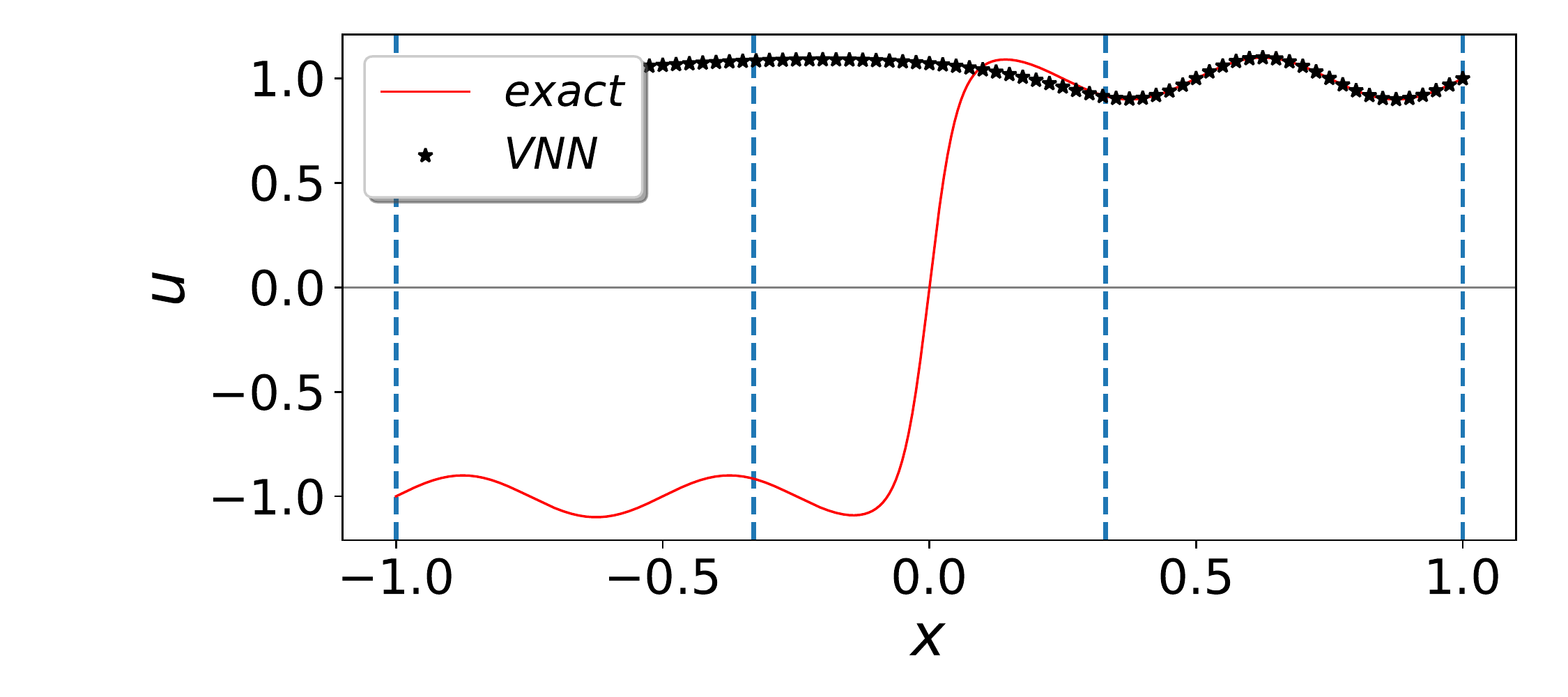}
		&
		\includegraphics[clip, trim=2cm 0cm 0.5cm 0cm, width=0.3\textwidth]{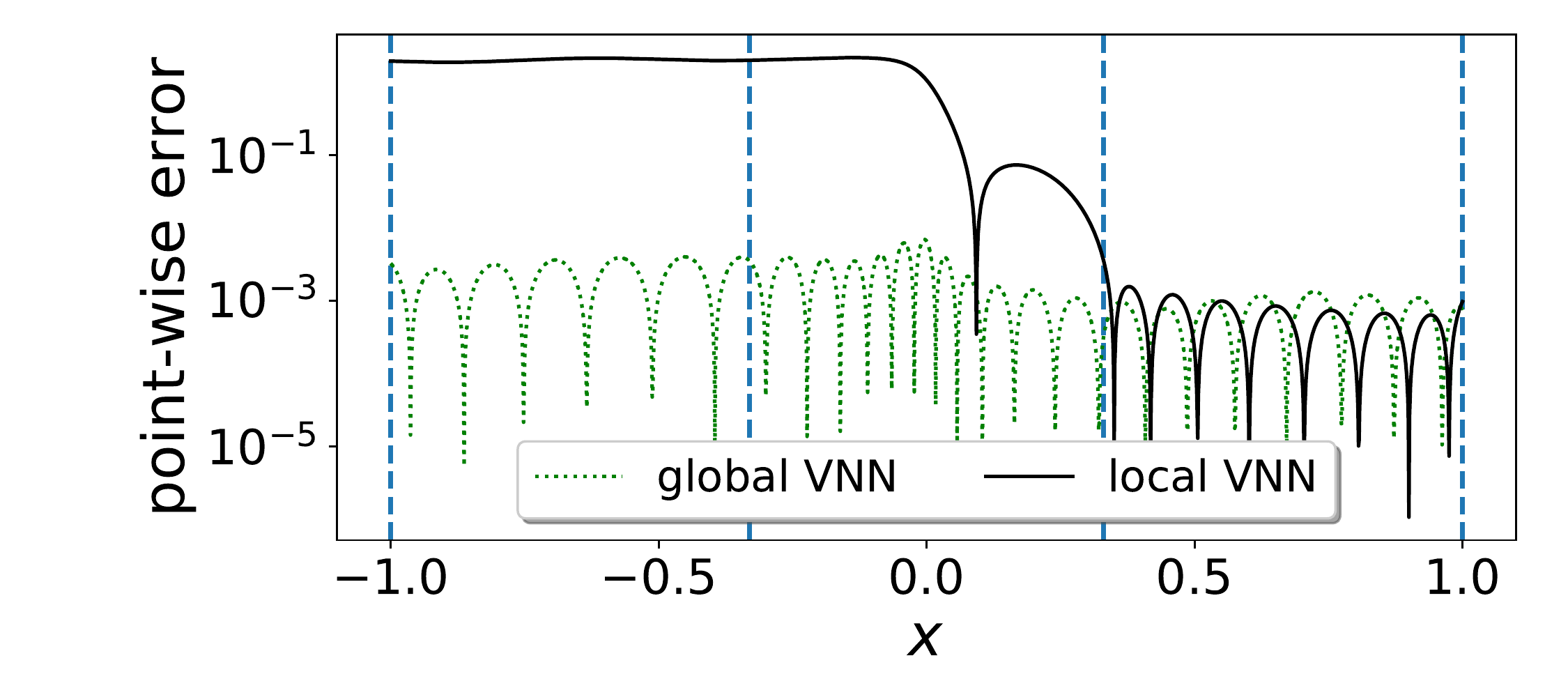}
		\\	[-11 pt]
	\end{tabular}
	\vspace{-0.05 in}
	\caption{\scriptsize \label{Fig: VNN cont fcn approx} VNN continuous function approximation. (\textbf{A}) global VNN: (left) the test functions, defined over the whole computational domain; (middle) the exact function and VNN approximation; (right) point-wise error. (\textbf{B}) local VNN: (left column) the test functions, defined locally over the individual sub-domains; (middle column) the exact function and VNN approximation; (right column) point-wise error. The dashed blue lines are the sub-domain boundaries. The VNN parameters are $\lbrace \ell = 4, \mathcal{N} = 20 , K = 60 , Q = 80 \rbrace$. We use Adam optimizer with learning rate $10^{-3}$.  }
\end{figure}

\begin{exm}[Continuous Function Approximation]
	\label{Ex: VNN ConFcn}
	We consider a smooth target function of the form $$u^{exact} = 0.1 \, \sin(4 \pi x) + \tanh (20 x),$$ which is defined over the domain $x \in \Omega = [-1,1]$. We use VNN to approximate the target function, using the global and local test functions. The results are shown in Fig. \ref{Fig: VNN cont fcn approx}.
	
\end{exm}

%

\noindent The considered function is smooth, continuous, and thus, can be approximated accurately using VNN with $N_{el}=1$. Using $\ell = 4$, $\mathcal{N} = 20$, $K = 60$, and $Q = 80$, we obtain $L^{\infty}$ error of $O(10^{-3})$. By dividing the domain into three equally spaced sub-domains, we define the test functions over each sub-domain to locally approximate the target function. The key point here in local VNN is to focus the learning process by zooming into the sub-domain, where we are more interested to approximate accurately. In this setting, the network parameters are specifically optimized such that the network solely captures the function within that sub-domain. The local VNN results in a slightly more accurate approximation in each sub-domain, compared to global VNN. We show later that this setting also extends the approximation beyond the local sub-domain.

\begin{figure}[!ht]
	\center
	\begin{tabular}{l  l  l}
		\multicolumn{3}{c}{global (single element) VNN with global test functions} \\  [-1 pt] 
		\includegraphics[clip, trim=1cm 0cm 0.5cm 0cm, width=0.3\linewidth]{globallearning_tanh_LegTest_L4_N20_NT60_NQ80_testfcns.pdf}
		&
		\includegraphics[clip, trim=1cm 0cm 0.5cm 0cm, width=0.3\linewidth]{globallearning_tanh_LegTest_L4_N20_NT60_NQ80_fcnplot.pdf}
		&
		\includegraphics[clip, trim=2cm 0cm 0.5cm 0cm, width=0.3\linewidth]{globallearning_tanh_LegTest_L4_N20_NT60_NQ80_pnterr.pdf}
		\\ [-6 pt]
		\multicolumn{3}{c}{
			\includegraphics[clip, trim=1cm 0cm 0.5cm 0cm, width=0.3\linewidth]{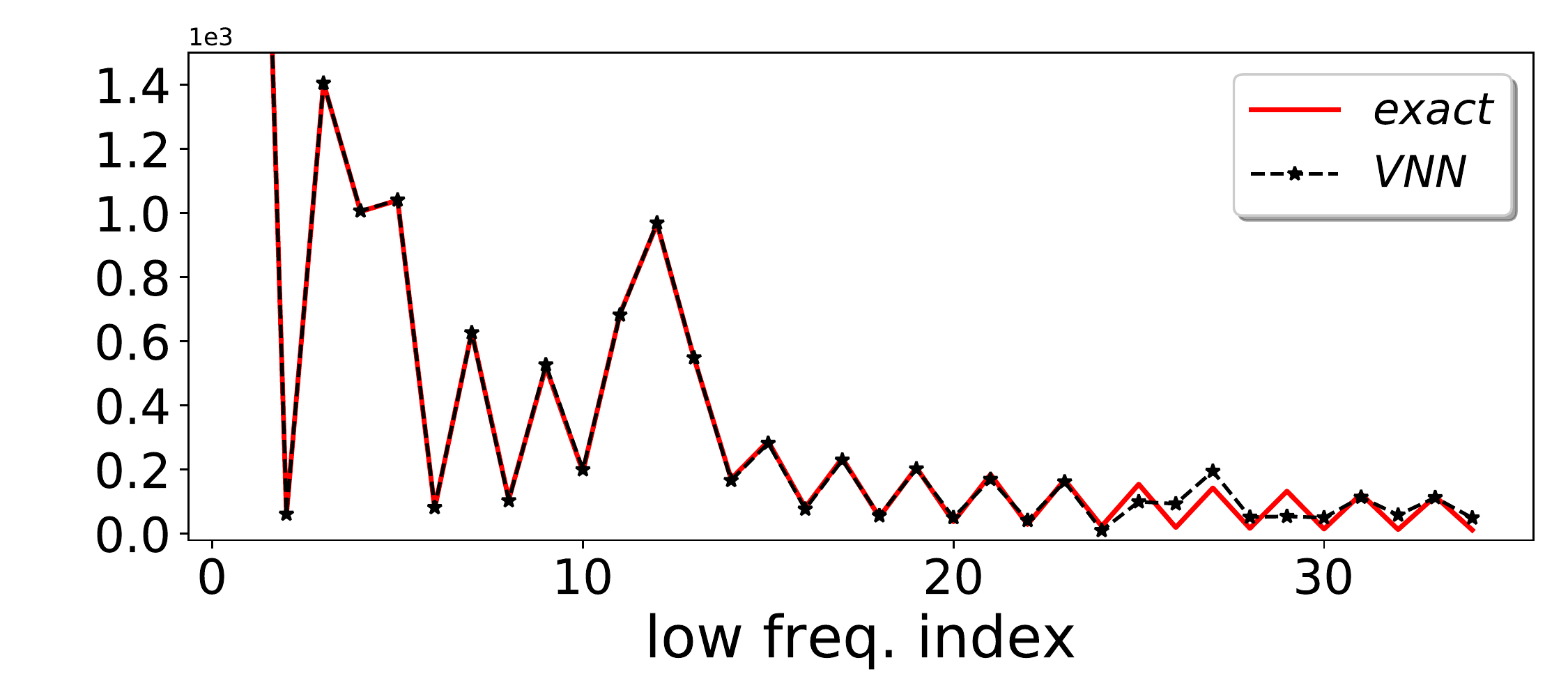}
			\quad
			\includegraphics[clip, trim=1cm 0cm 0.5cm 0cm, width=0.3\linewidth]{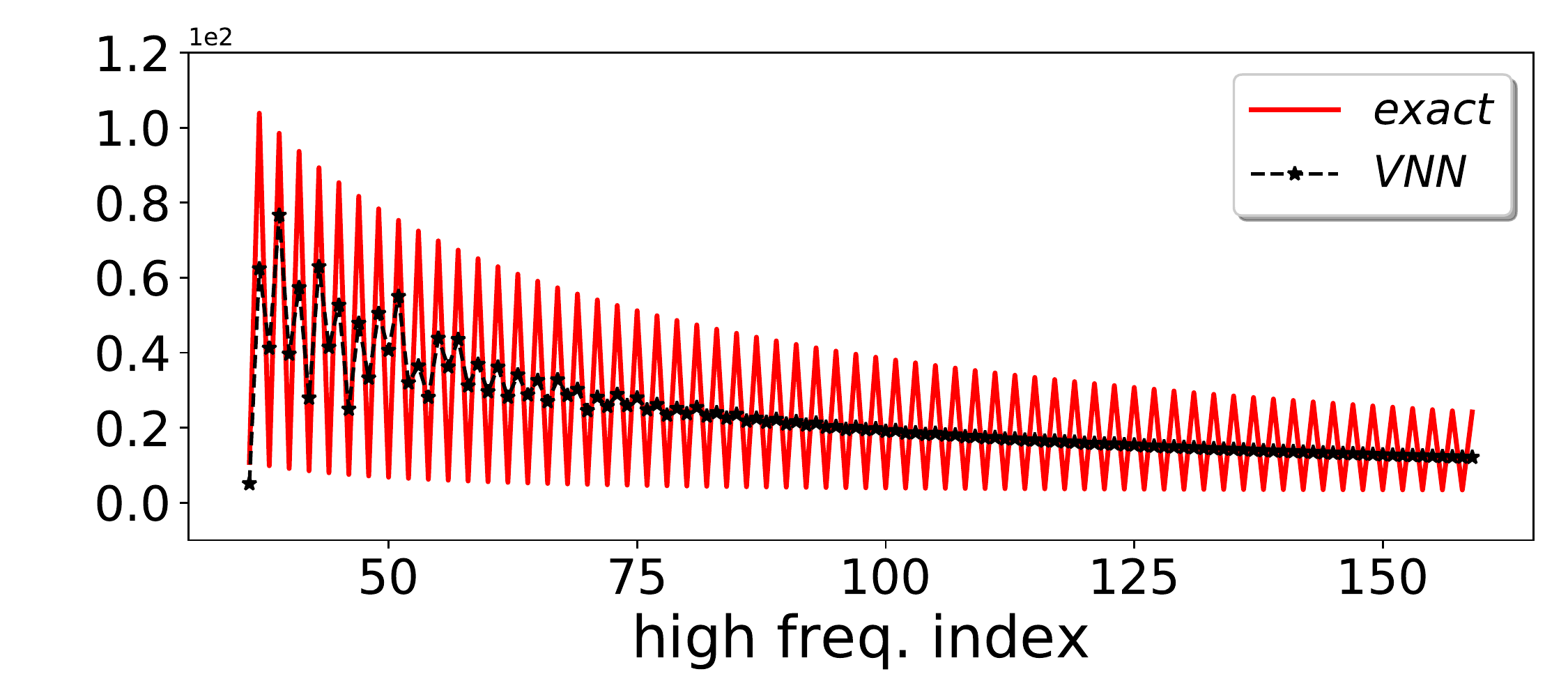}
		}
	\end{tabular}
	\vspace{-0.15 in}
	\caption{\scriptsize \label{Fig: VNN dis cont fcn approx} VNN discontinuous function approximation: global (single element) VNN. Top row from left: test functions defined over the whole computational domain, the exact function and VNN prediction, and point-wise error. Bottom row: the target function and prediction in low and high frequency domain. The VNN parameters are $\lbrace \ell = 4, \mathcal{N} = 20 , K = 60 , Q = 80 \rbrace$. We use Adam optimizer with learning rate $10^{-3}$.  }
\end{figure}

\begin{figure}[!ht]
	\center
	\begin{tabular}{l  l  l}
		\multicolumn{3}{c}{local (elemental) VNN with local test functions} 
		\\  [-1 pt] 
		\includegraphics[clip, trim=1cm 0cm 0.5cm 0cm, width=0.3\textwidth]{locallearning_tanh_LegTest_L4_N20_NT60_NQ80_testfcns_elem1.pdf}
		&
		\includegraphics[clip, trim=1cm 0cm 0.5cm 0cm, width=0.3\textwidth]{locallearning_tanh_LegTest_L4_N20_NT60_NQ80_testfcns_elem2.pdf}
		&
		\includegraphics[clip, trim=1cm 0cm 0.5cm 0cm, width=0.3\textwidth]{locallearning_tanh_LegTest_L4_N20_NT60_NQ80_testfcns_elem3.pdf}
		\\	[-6 pt]
		\includegraphics[clip, trim=1cm 0cm 0.5cm 0cm, width=0.3\textwidth]{locallearning_tanh_LegTest_L4_N20_NT60_NQ80_fcnplot_elem1.pdf}
		&
		\includegraphics[clip, trim=1cm 0cm 0.5cm 0cm, width=0.3\textwidth]{locallearning_tanh_LegTest_L4_N20_NT60_NQ80_fcnplot_elem2.pdf}
		&
		\includegraphics[clip, trim=1cm 0cm 0.5cm 0cm, width=0.3\textwidth]{locallearning_tanh_LegTest_L4_N20_NT60_NQ80_fcnplot_elem3.pdf}
		\\	[-7 pt]	
		\includegraphics[clip, trim=0cm 0cm 2.5cm 0cm, width=0.3\textwidth]{locallearning_tanh_LegTest_L4_N20_NT60_NQ80_pnterr_compare_elem1.pdf}
		&
		\includegraphics[clip, trim=0cm 0cm 2.5cm 0cm, width=0.3\textwidth]{locallearning_tanh_LegTest_L4_N20_NT60_NQ80_pnterr_compare_elem2.pdf}
		&
		\includegraphics[clip, trim=0cm 0cm 2.5cm 0cm, width=0.3\textwidth]{locallearning_tanh_LegTest_L4_N20_NT60_NQ80_pnterr_compare_elem3.pdf}
		\\	[-3 pt]			
		\includegraphics[clip, trim=1cm 0cm 0.5cm 0cm, width=0.3\textwidth]{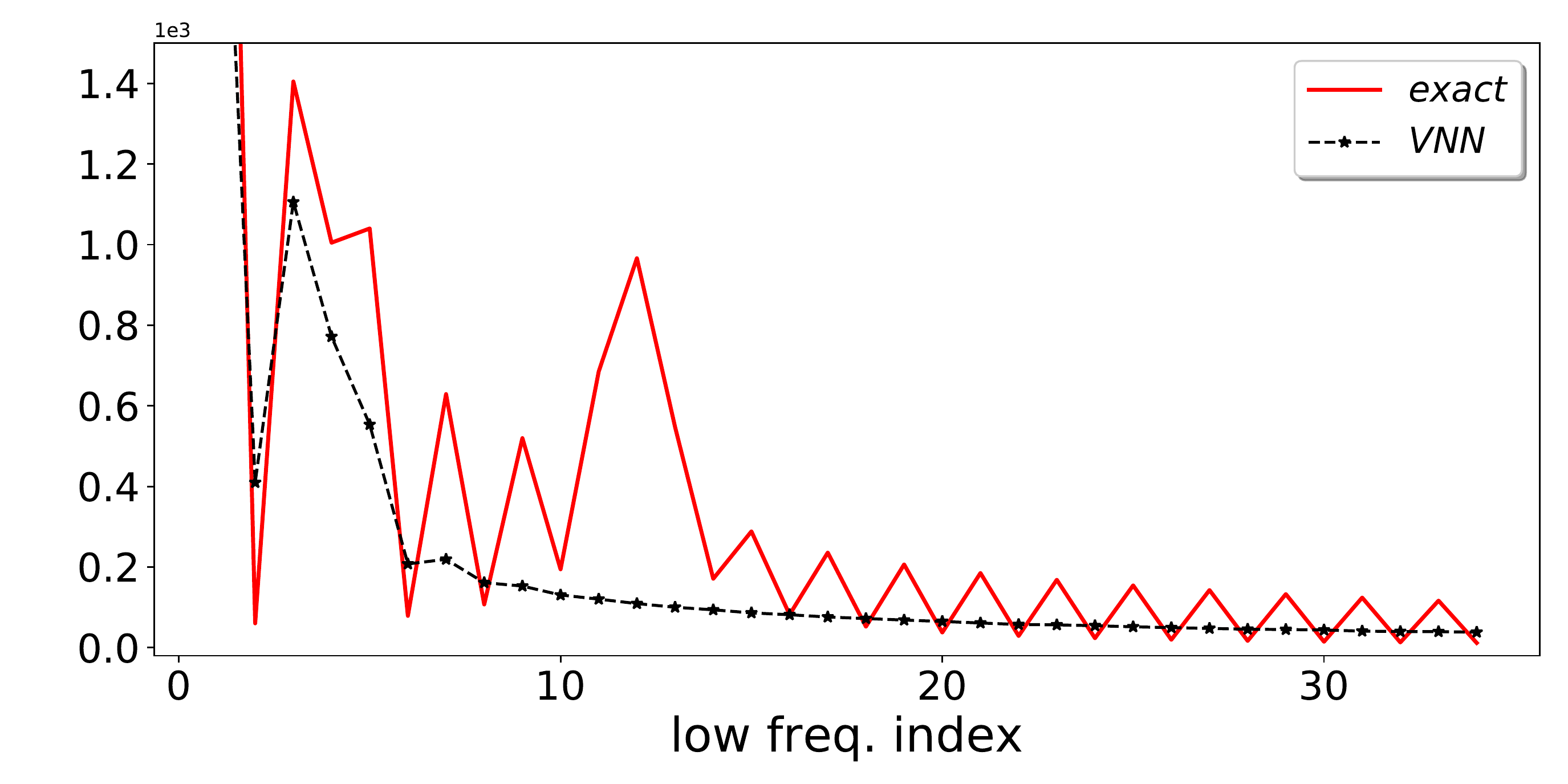}
		&
		\includegraphics[clip, trim=1cm 0cm 0.5cm 0cm, width=0.3\textwidth]{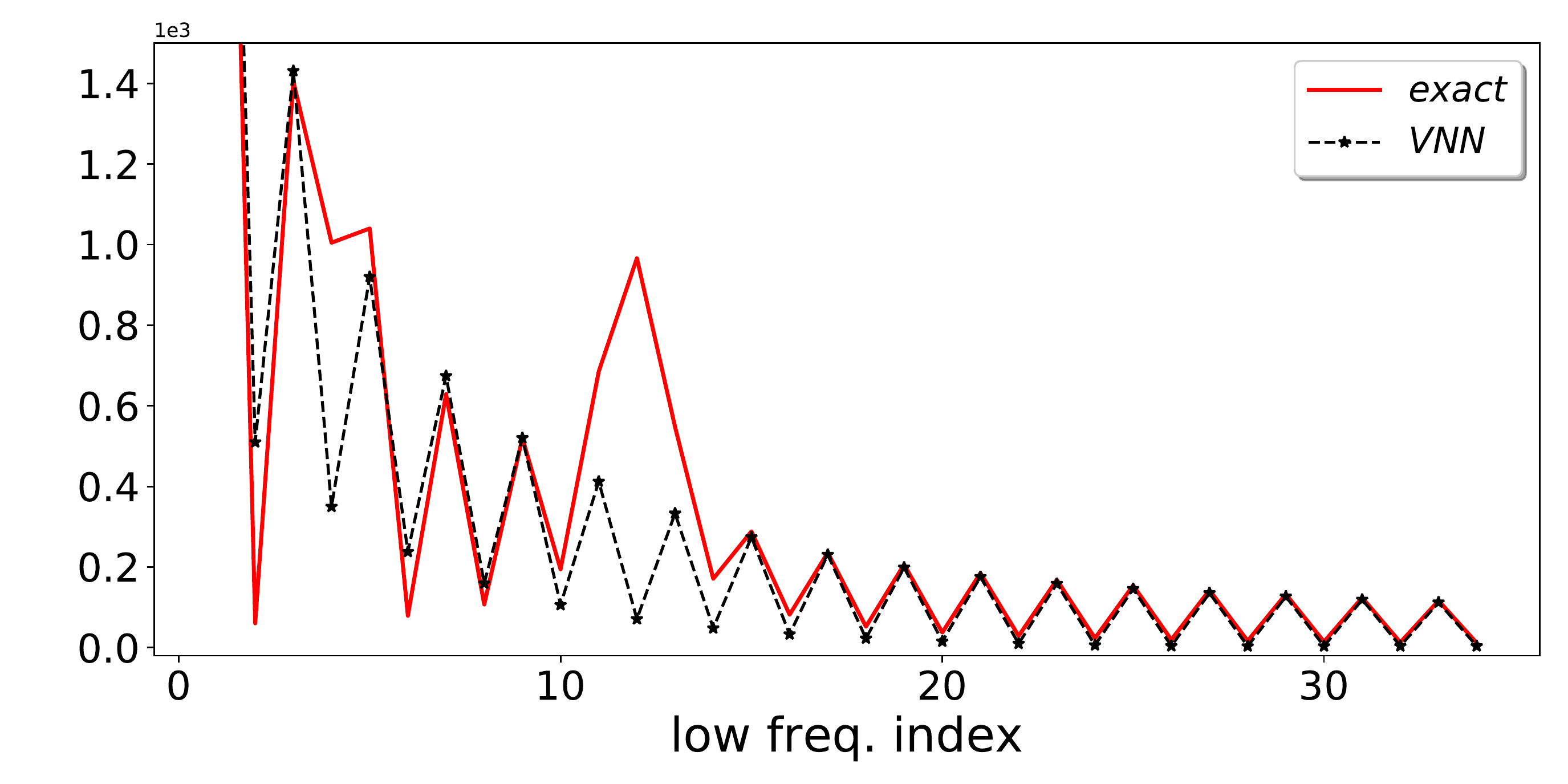}
		&
		\includegraphics[clip, trim=1cm 0cm 0.5cm 0cm, width=0.3\textwidth]{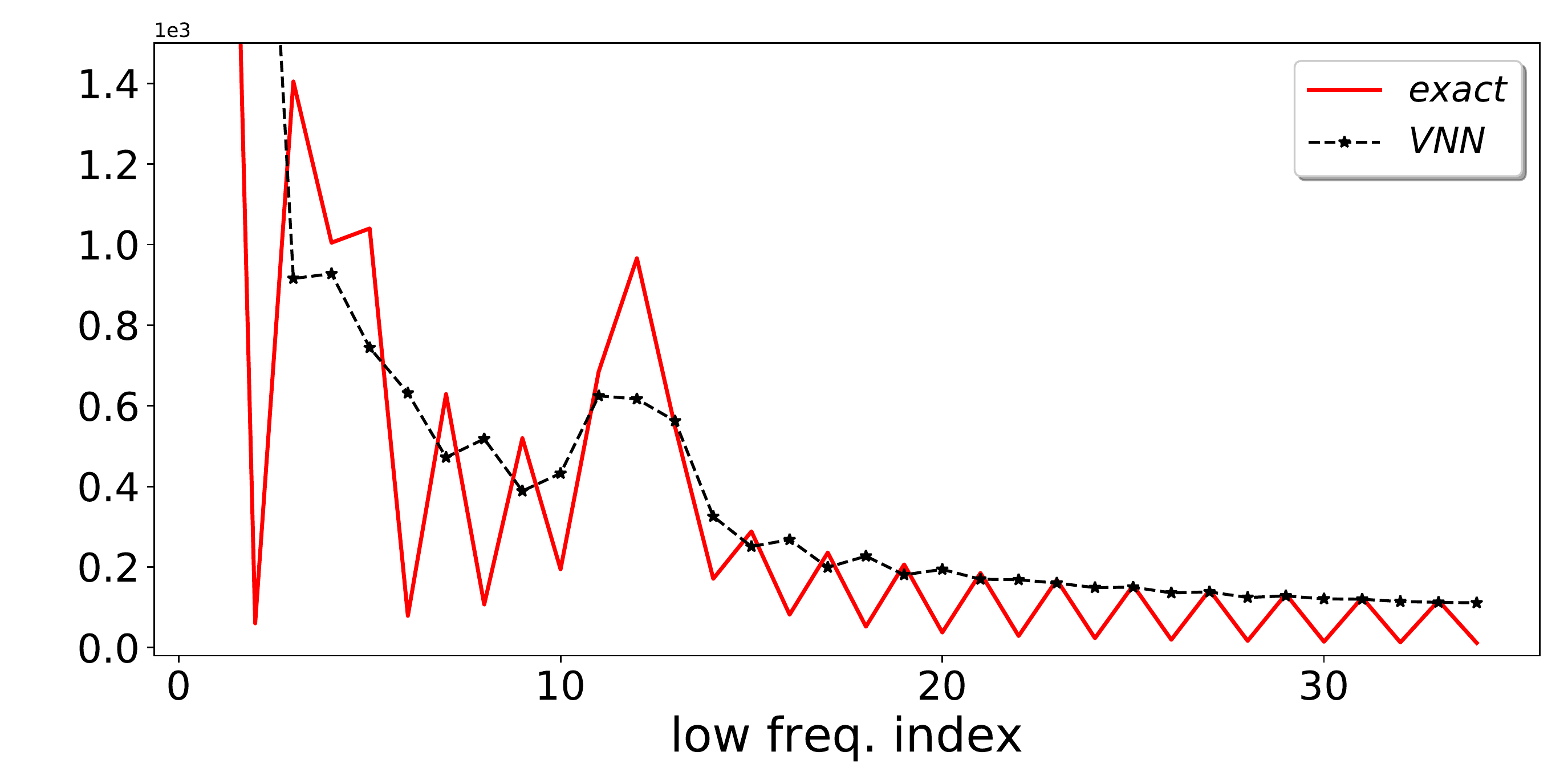}
		\\	[-4 pt]	
		\includegraphics[clip, trim=1cm 0cm 0.5cm 0cm, width=0.3\textwidth]{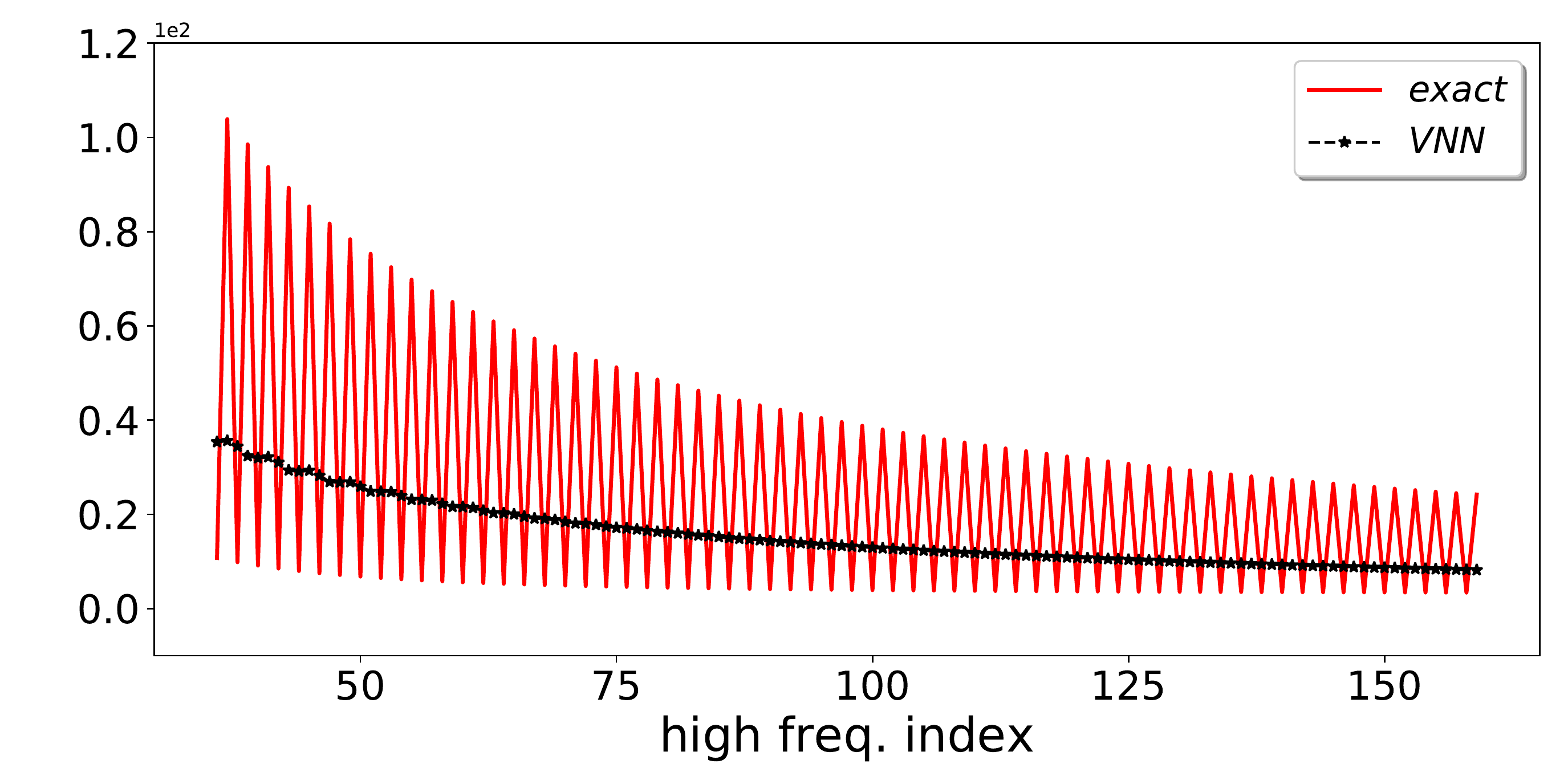}
		&
		\includegraphics[clip, trim=1cm 0cm 0.5cm 0cm, width=0.3\textwidth]{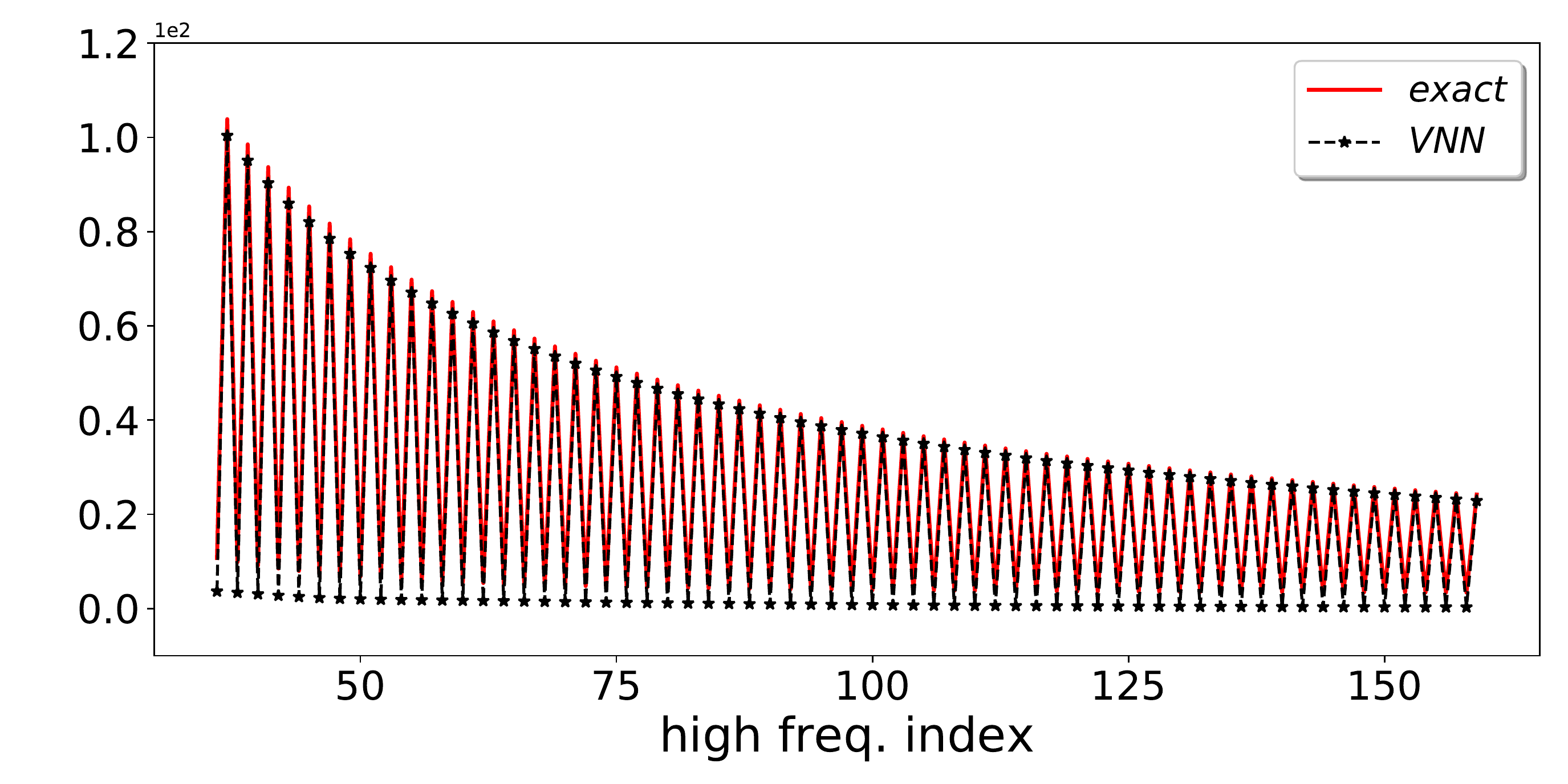}
		&
		\includegraphics[clip, trim=1cm 0cm 0.5cm 0cm, width=0.3\textwidth]{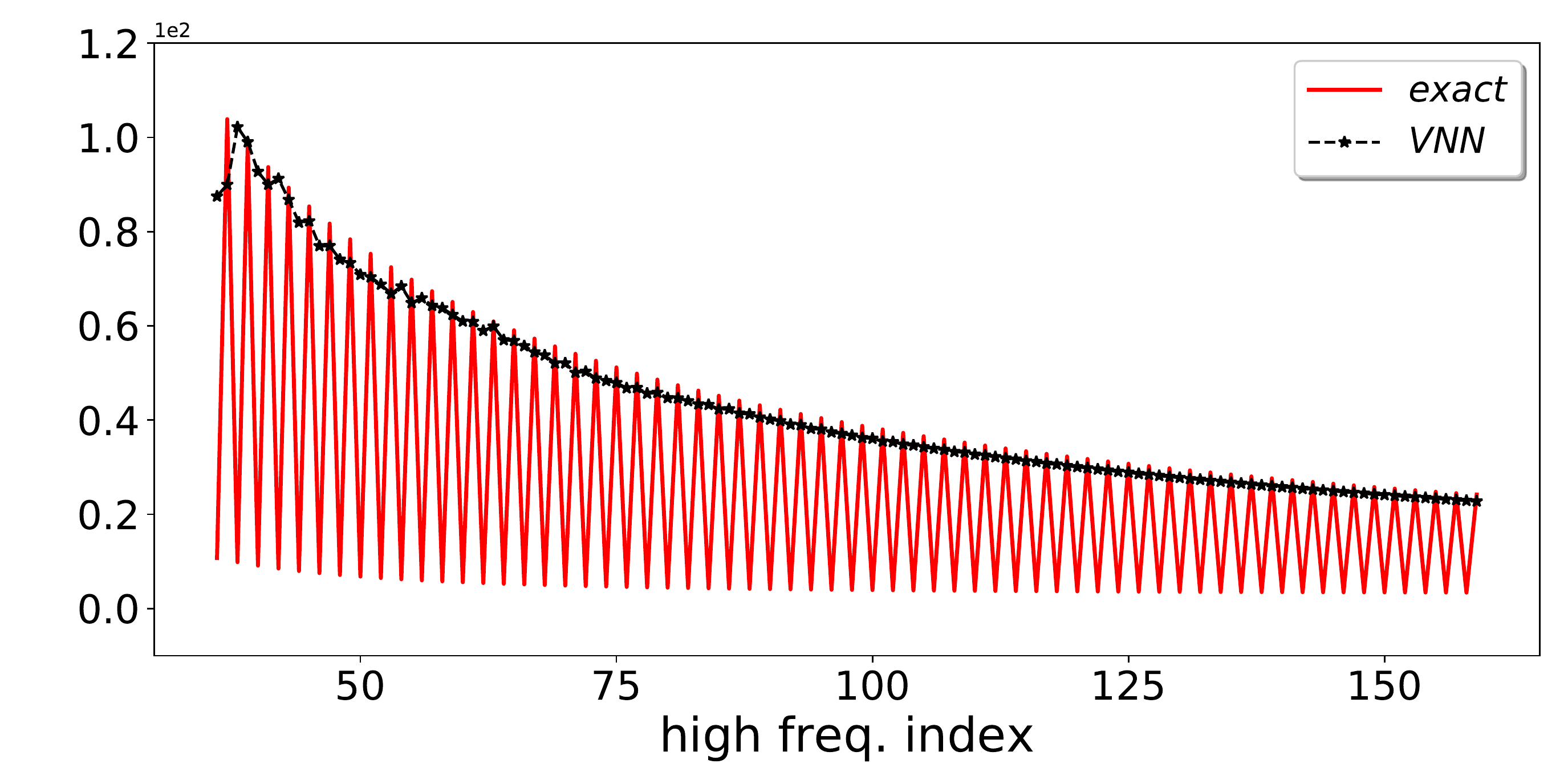}
	\end{tabular}
	\vspace{-0.15 in}
	\caption{\scriptsize \label{Fig: VNN dis cont fcn approx 2} VNN discontinuous function approximation: local (elemental) VNN. The dashed blue line are the sub-domain boundaries. The row-wise captions are; First: locally defined test functions over each sub-domain. Second: 
	the exact function and VNN prediction. Third: point-wise error. Fourth and fifth: low and high frequency indexes of the exact function and VNN prediction.  
	The VNN parameters are $\ell = 4$, $\mathcal{N} = 20 $, $K = 60$ local test functions, $Q = 80 $ quadrature points in each sub-domain, and Adam optimizer with learning rate $10^{-3}$.  }
\end{figure}

\begin{figure}[!ht]
	\center
	\begin{tabular}{l  l  l}
		\multicolumn{3}{c}{multi elemental VNN with local test functions} \\  [-1 pt] 
		\includegraphics[clip, trim=0cm 0cm 0.5cm 0cm, width=0.3\linewidth]{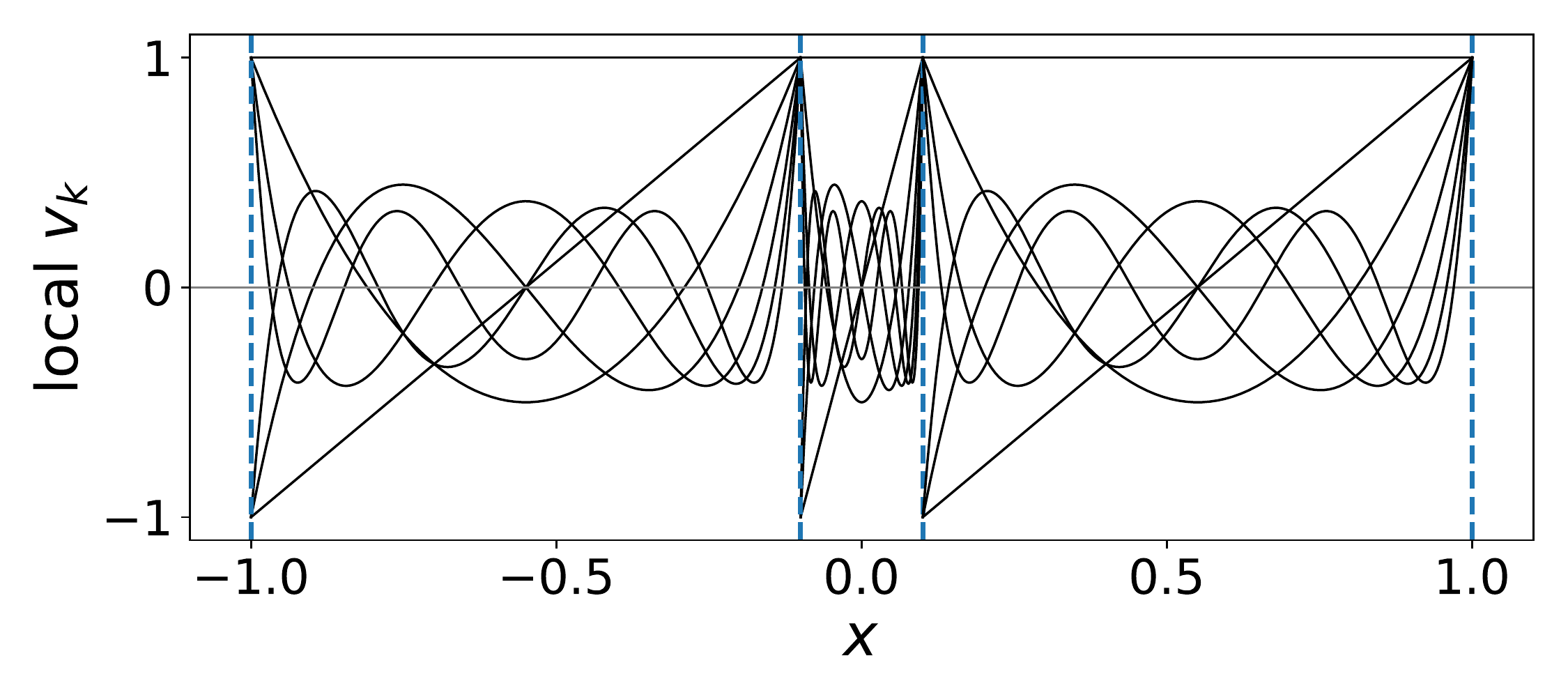}
		&
		\includegraphics[clip, trim=1cm 0cm 0.5cm 0cm, width=0.3\linewidth]{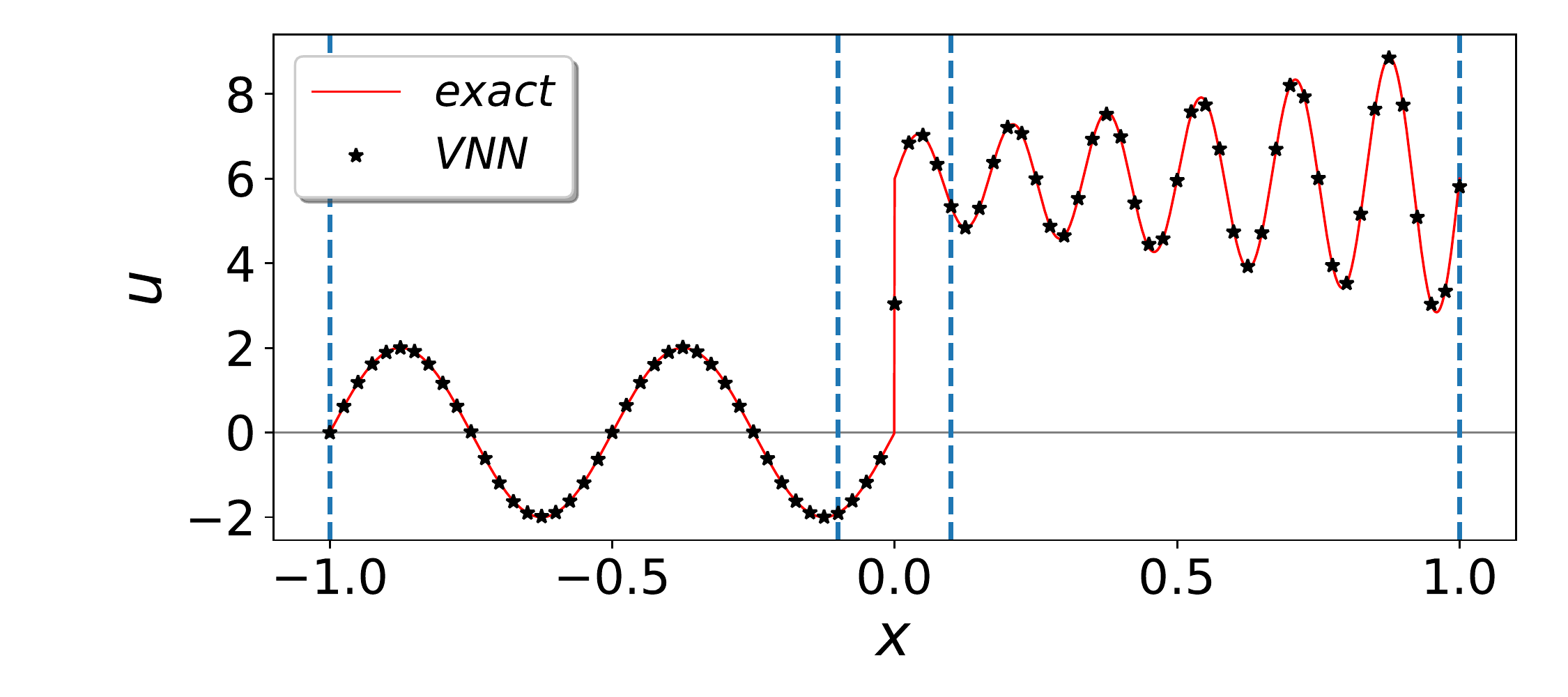}
		&
		\includegraphics[clip, trim=2cm 0cm 0.5cm 0cm, width=0.3\linewidth]{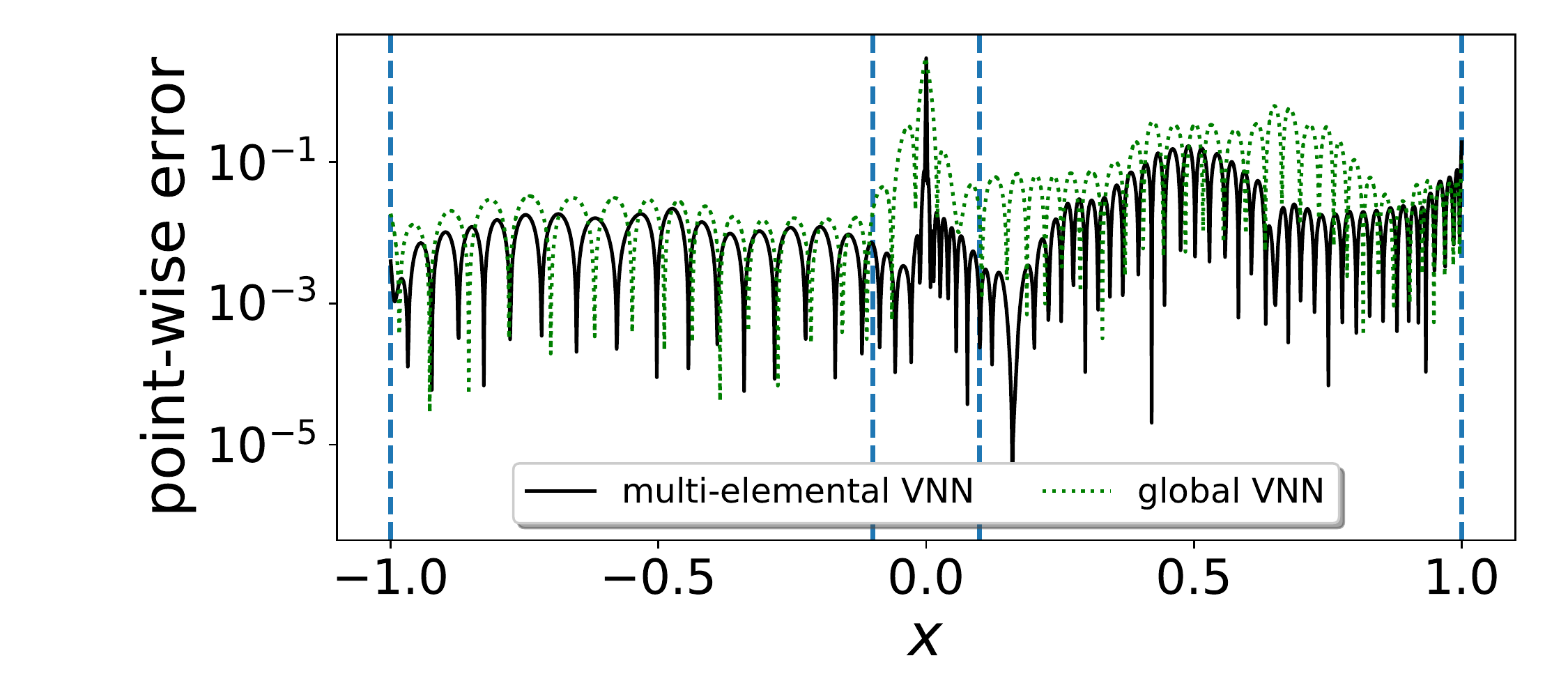}
		\\ [-6 pt]
		\multicolumn{3}{c}{
			\includegraphics[clip, trim=0cm 0cm 0cm 0cm, width=0.3\linewidth]{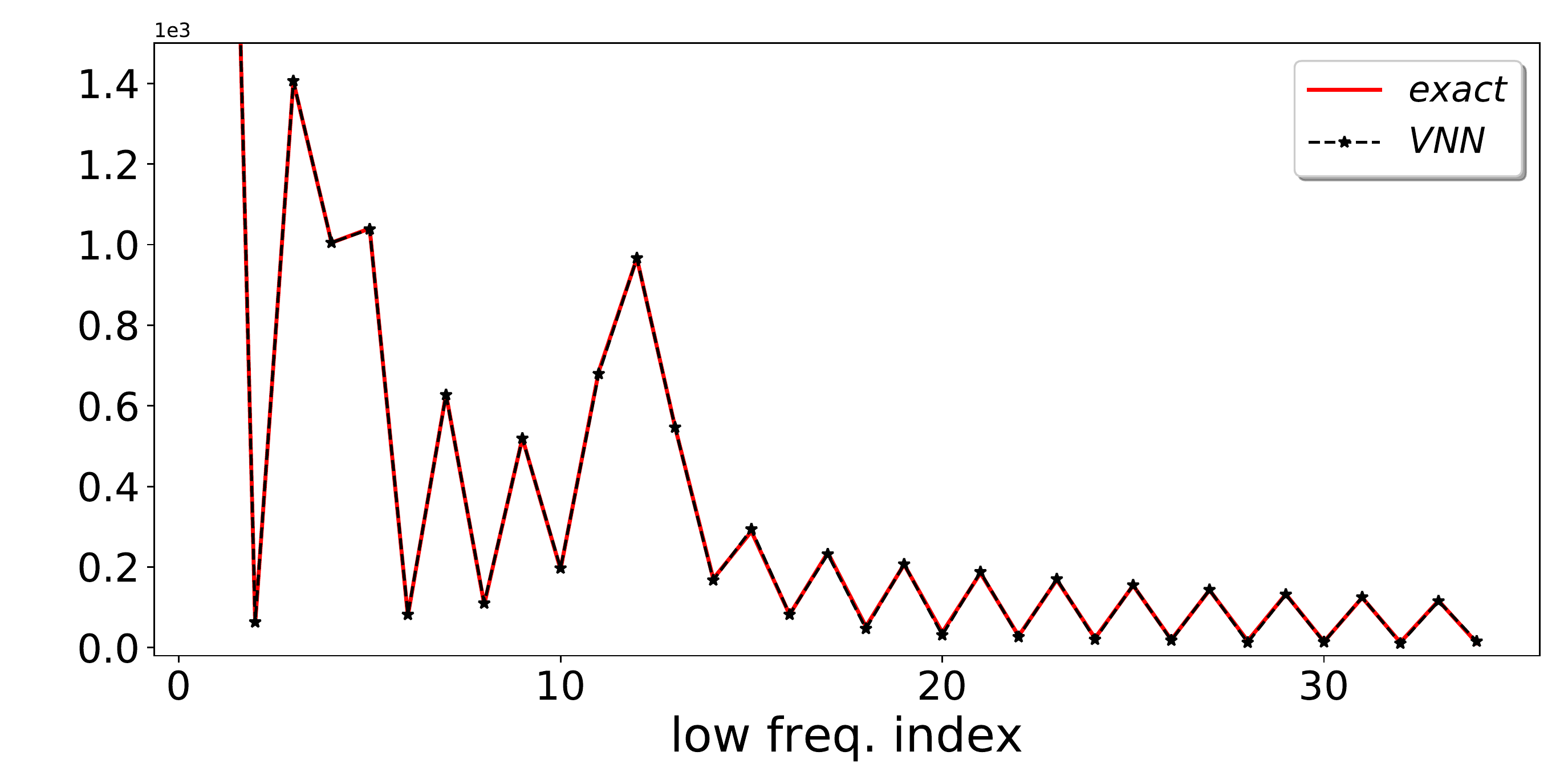}
			\quad
			\includegraphics[clip, trim=0cm 0cm 0cm 0cm, width=0.3\linewidth]{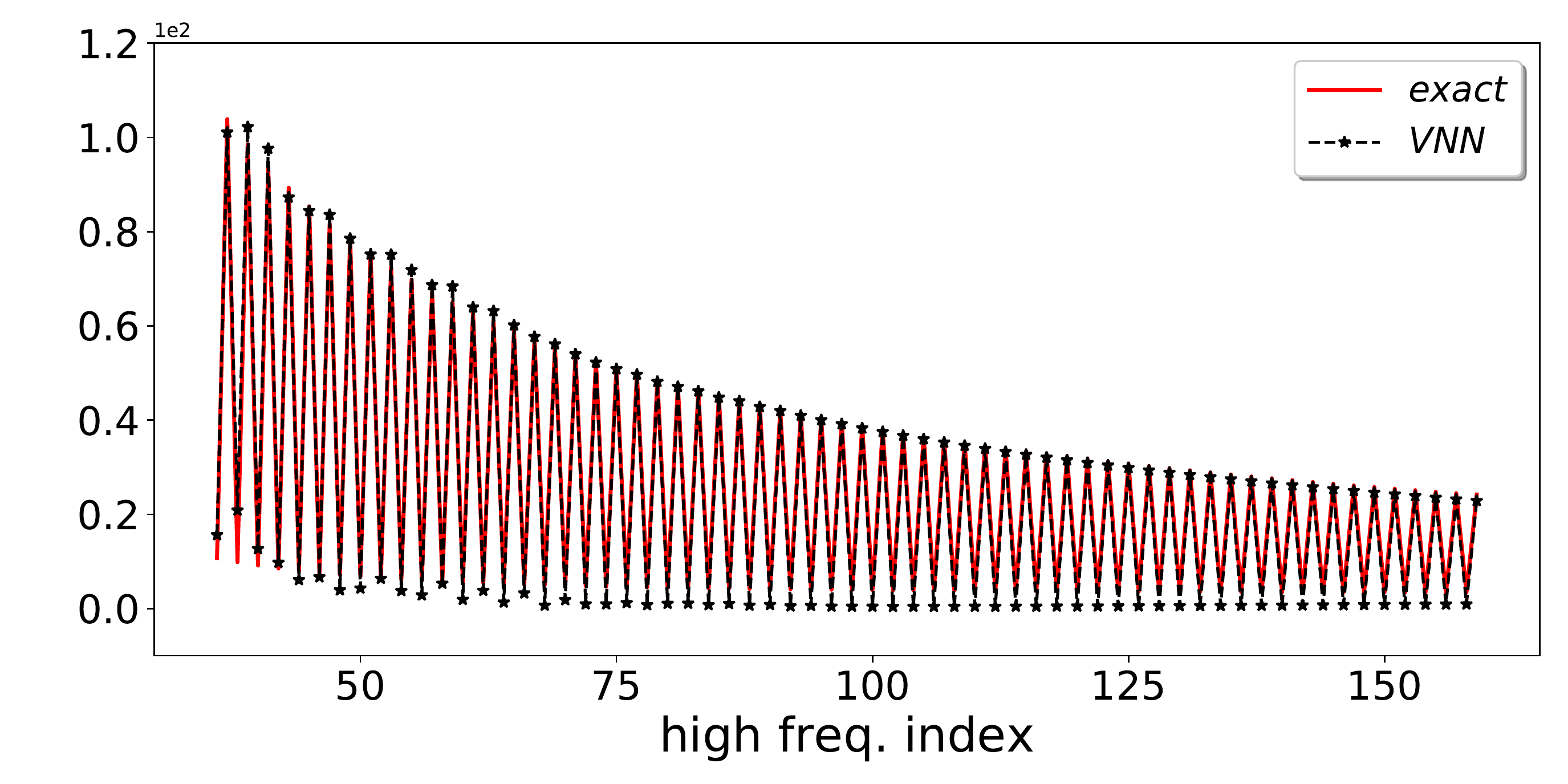}
		}
	\end{tabular}
 	\vspace{-0.15 in}
	\caption{\scriptsize \label{Fig: VNN dis cont fcn approx 3} VNN discontinuous function approximation: multi-elemental VNN. Top left: the domain is divided into three sub-domains and the test functions are defined locally over each sub-domain. The dashed blue line are the sub-domains boundaries. Top middle: the exact function and VNN prediction. Top right: the point-wise error. Bottom row: the low and high frequency indexes of exact function and VNN prediction. The VNN parameters are $\ell = 4$, $\mathcal{N} = 20 $, $K = 60$ local test functions, $Q = 80 $ quadrature points in each sub-domain, and Adam optimizer with learning rate $10^{-3}$.  }
\end{figure}

\begin{exm}[Discontinuous Function Approximation]
	We consider a piecewise continuous target function of the form $$u^{exact} = \begin{cases}
	2 \sin(4\pi x) & x \in [-1,0), \\
	6+ \text{e}^{1.2 x} \sin(12 \pi x) & x \in (0, 1],
	\end{cases},$$ which is defined over the domain $x \in \Omega = [-1,1]$ and has a jump of magnitude 6 at $x = 0$. We use VNN to approximate the target function, using the global and local test functions. The results are shown in Figs. \ref{Fig: VNN dis cont fcn approx}, \ref{Fig: VNN dis cont fcn approx 2}, and \ref{Fig: VNN dis cont fcn approx 3}.
\end{exm}

%

%

%

\noindent  The exact solution is comprised of sinusoidal waves with frequencies $4 \pi$ and $12 \pi$, and a discontinuity. It is interesting to compare the approximations in the Fourier domain, where the two sinusoidal waves are represented by low frequency index and the discontinuity is represented by high frequency index. Figure \ref{Fig: VNN dis cont fcn approx} shows the results of global (single element) VNN with global test functions in approximating the discontinuous function, where we see that the $L^{\infty}$ error is of order $O(10^{-1})$, happening close to the discontinuity. In the Fourier domain, the network learns the low frequency index of the target function, however, fails to capture the high frequency index. In Fig. \ref{Fig: VNN dis cont fcn approx 2}, we show the results of local (elemental) VNN with local test functions, in which by defining a relatively small sub-domain close to the discontinuity, we make sure that the network can capture the high frequency index very accurately. In Fig. \ref{Fig: VNN dis cont fcn approx 3}, we show the results of multi-elemental VNN, where we see that the network can capture the target function accurately in the Fourier domain. We note that in general a DNN regresses a function by first learning the discontinuity and then the low to high frequencies. The multi-elemental setting, however, can optimally change this learning pattern by considering different domain decompositions. 

\begin{figure}[!ht]
	\center
	\begin{tabular}{c  c c}
		\multicolumn{3}{c}{{\scriptsize \quad\quad exact solution}} \\  [-3 pt] 
		
		\multicolumn{3}{c}{
			\includegraphics[clip, trim=0cm 0cm 0.5cm 1.5cm, width=0.4\linewidth]{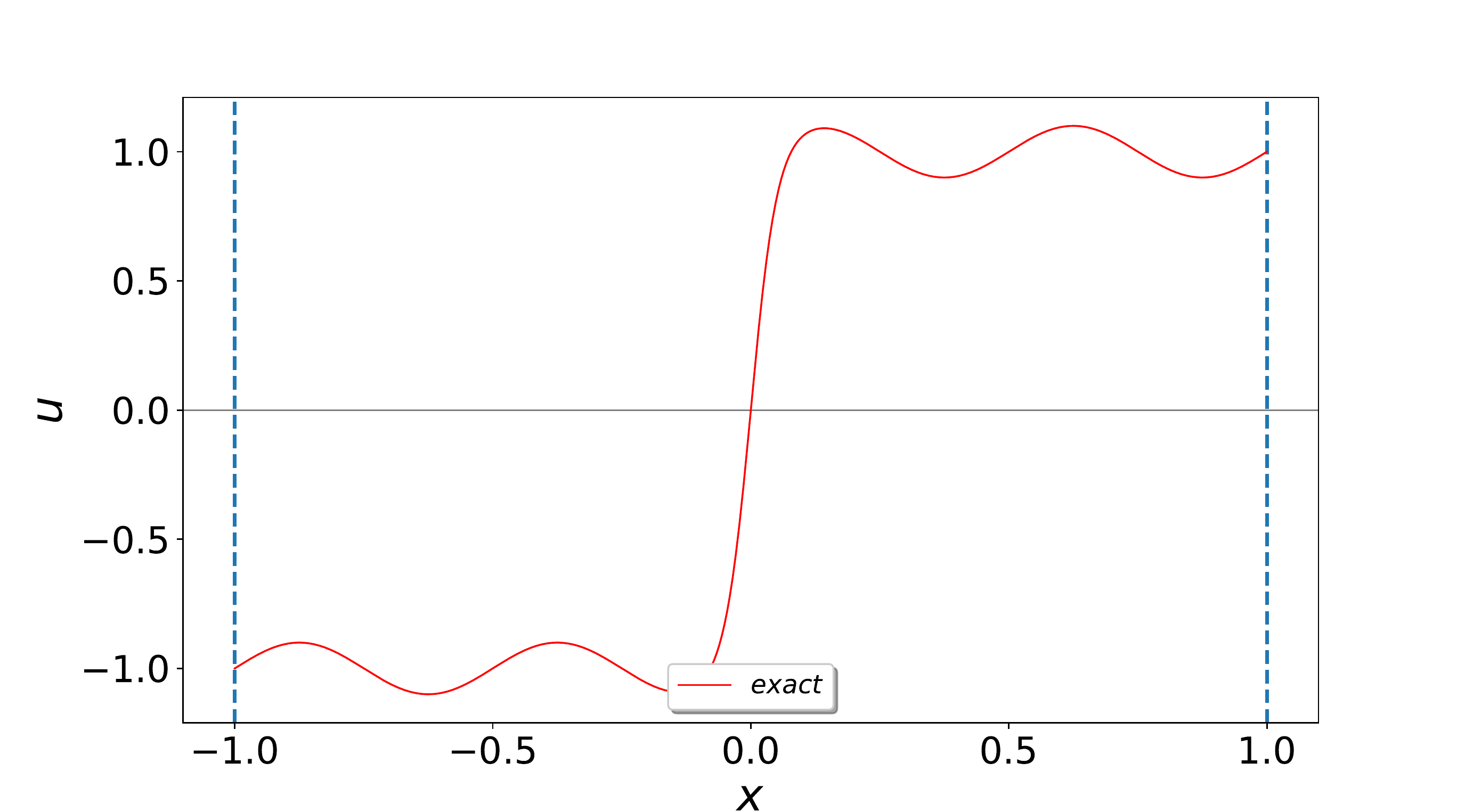}
		}
		\\
		{\scriptsize \quad\quad VNN: sine activation function} 
		& {\scriptsize \quad\quad VNN: tanh activation function} 
		& {\scriptsize \quad\quad VNN: relu activation function}\\  [-3 pt] 
		\\ [-12 pt]
		\includegraphics[clip, trim=0cm 0cm 2.75cm 1.5cm, width=0.3\linewidth]{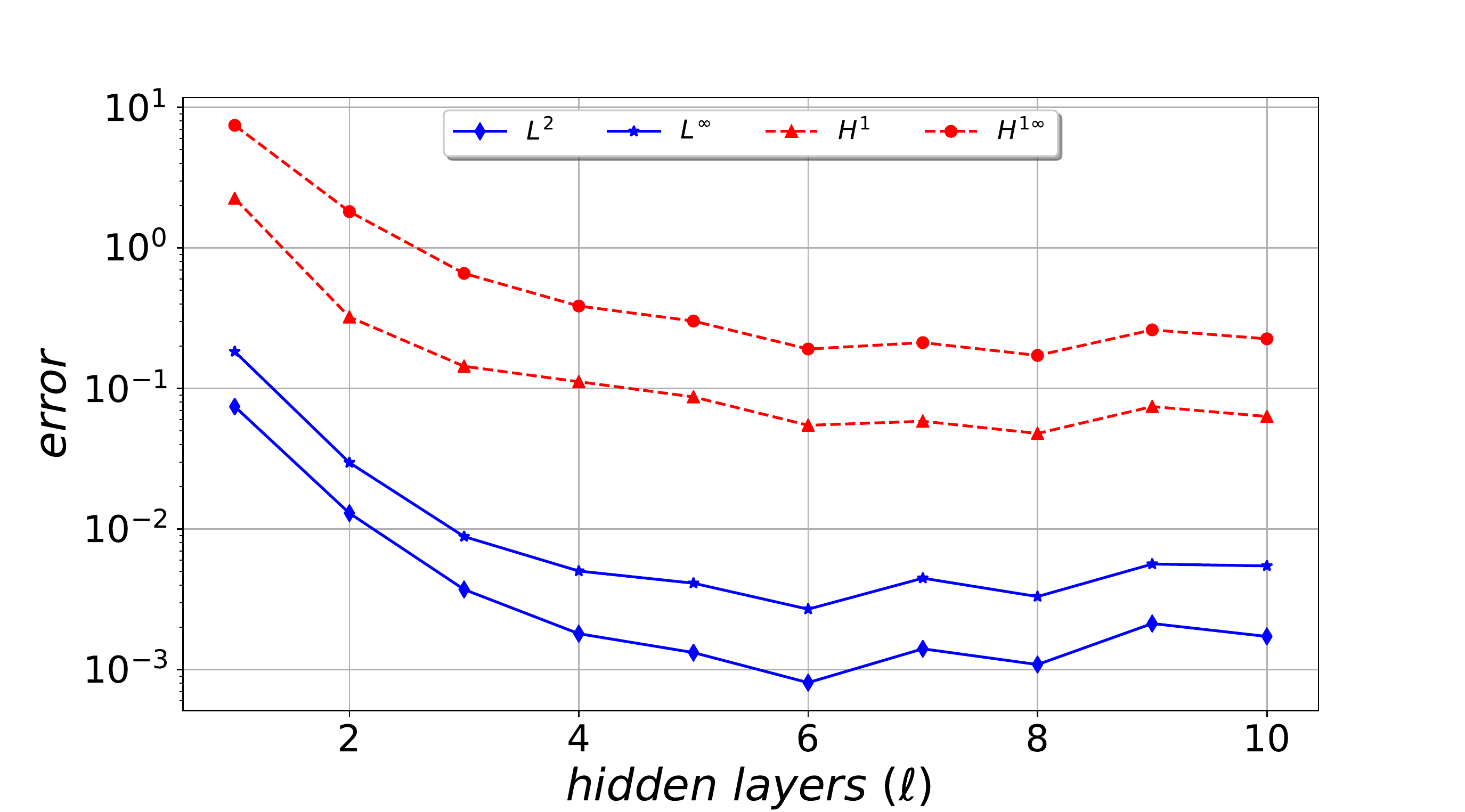}
		&
		\includegraphics[clip, trim=0cm 0cm 2.75cm 1.5cm, width=0.3\linewidth]{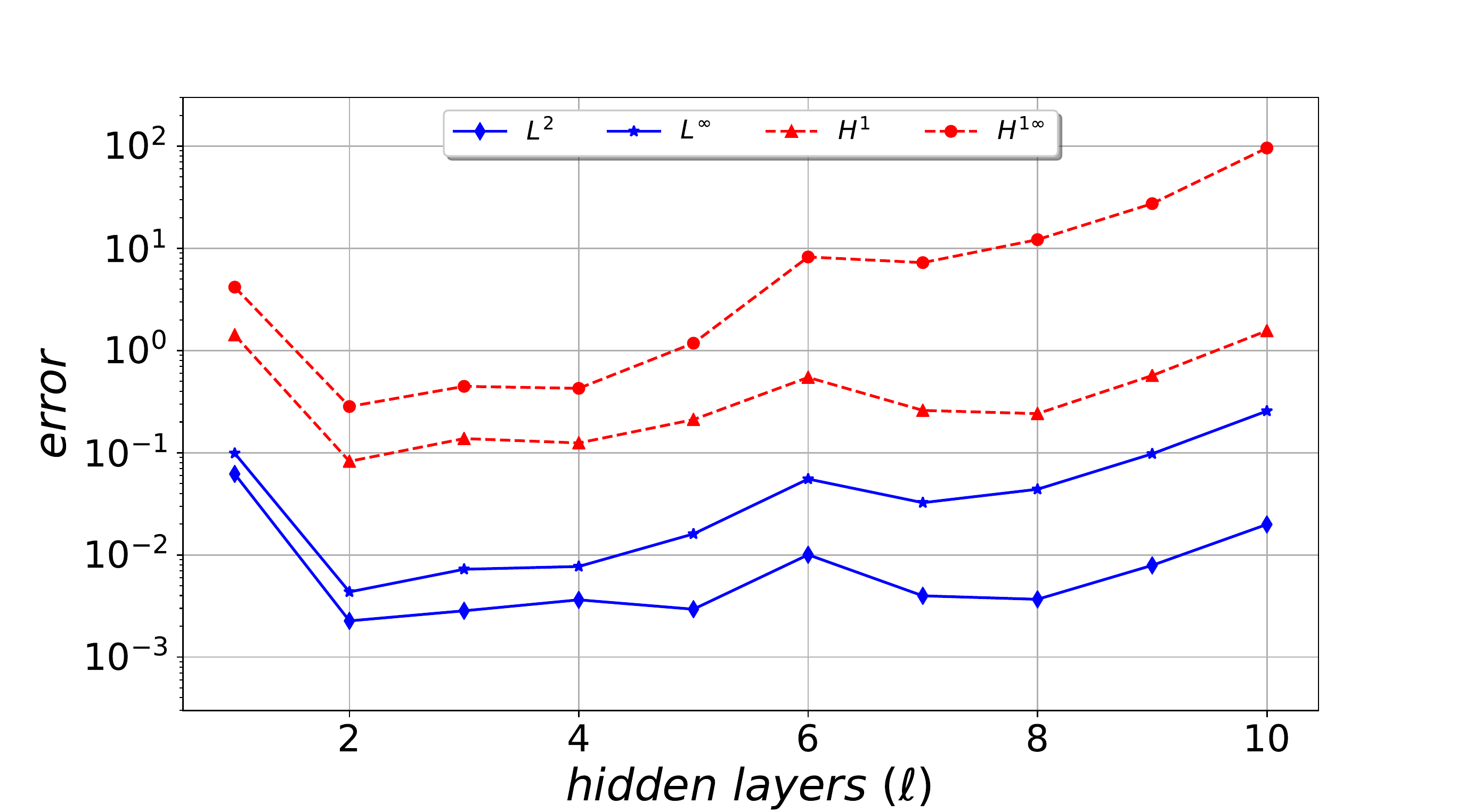}
		&
		\includegraphics[clip, trim=0cm 0cm 2.75cm 1.5cm, width=0.3\linewidth]{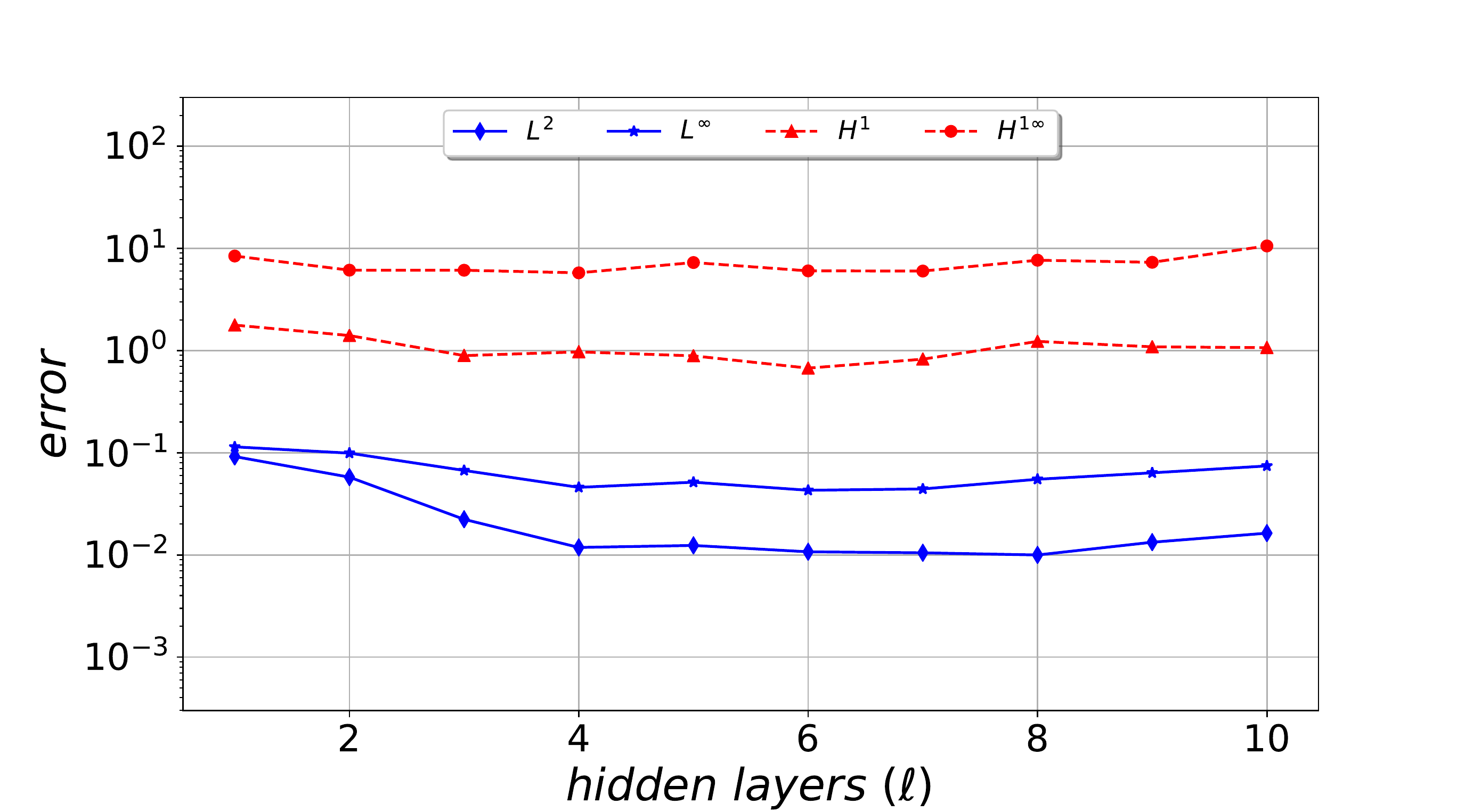}
	\end{tabular}
	\vspace{-0.15 in}
	\caption{\scriptsize \label{Fig: VNN err convergance cont fcn} Global (single element) VNN prediction of the exact function in \ref{Ex: VNN ConFcn} (Top) and error convergence with network depth (bottom) for sine, tanh, and ReLu activation functions. VNN is a fully connected network with parameters $\mathcal{N} = 20$, $K = 60 $, $Q = 80 $, and Adam optimizer with learning rate $10^{-3}$. The results are averaged over 8 different network initialization. }
\end{figure}

\begin{exm}[Convergence By Depth]
	We consider the target function given in example \ref{Ex: VNN ConFcn}. We study the approximation error convergence of global (single element) VNN for different activation functions and network depth. The results are shown in Fig. \ref{Fig: VNN err convergance cont fcn}.
\end{exm}

%

\noindent  The compositional structure of DNNs is responsible for their high expressivity and thus a (relatively) deeper network is assumed to provide a more accurate regression approximation. We see in Fig. \ref{Fig: VNN err convergance cont fcn} that by increasing the depth of network in VNN formulation while keeping the width constant, the error drops with different rates for various activation functions. The error saturates after certain depth, which is mainly because the network parameters cannot be further optimized more accurately. 

We recall that the loss function in VNN formulation is based on the projection of discrepancy of network output and the target function onto polynomial function space, where a successful minimization of loss function leads to convergence in that space. We observe, however, that the error in $H^1$-norm also drops and, therefore, in addition to accurate approximation of the target function, the network also learns the first derivative of the target function (but less accurately). The accuracy and convergence rate strongly depend on the choice of activation function as we observe that ReLu is not successful in learning the derivative of target function compared to sine and tanh activations. This is an important feature of local VNN, as the network can further capture the target function beyond the local support by following the trend of its first derivative at the boundary of a sub-domain, while the loss function is only obtained over the local sub-domain; we further discuss this feature in the following.

\begin{figure}[!ht]
	\center
	\includegraphics[clip, trim=0cm 0cm 0.5cm 0cm, width=0.46\linewidth]{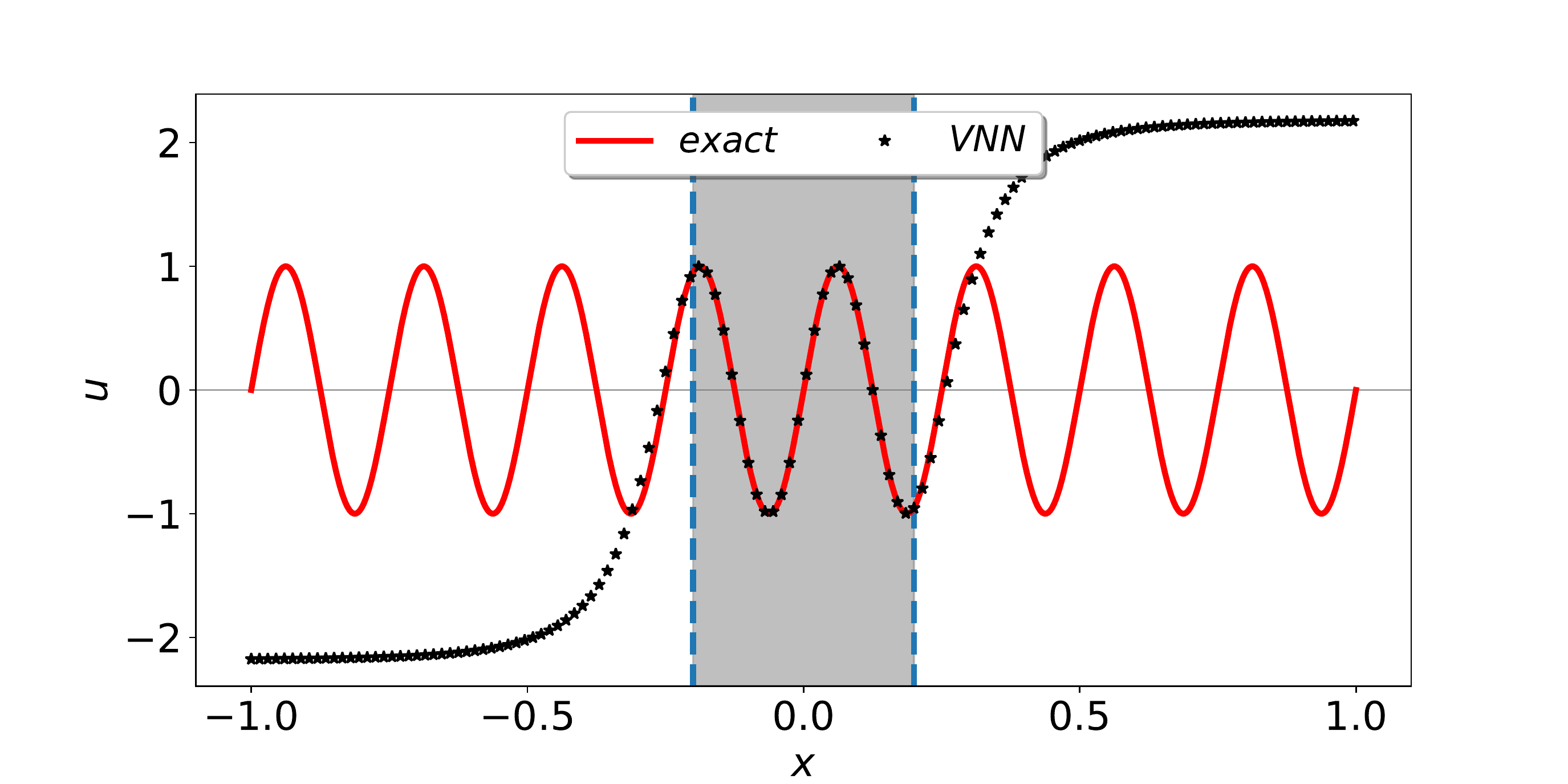}
	\includegraphics[clip, trim=0cm 0cm 0.5cm 0cm, width=0.46\linewidth]{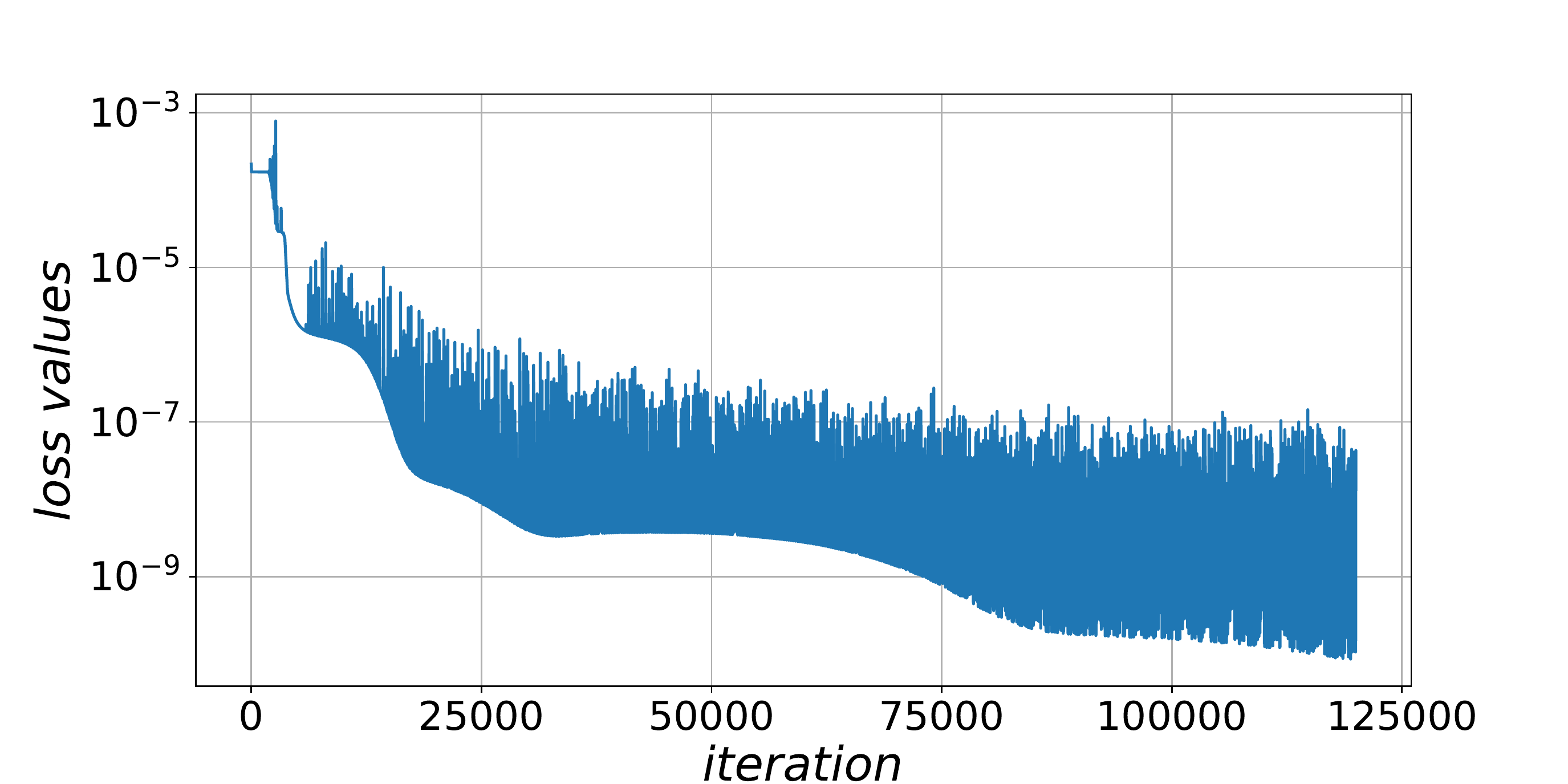}
	\\
	\includegraphics[clip, trim=0cm 0cm 0.5cm 0cm, width=0.46\linewidth]{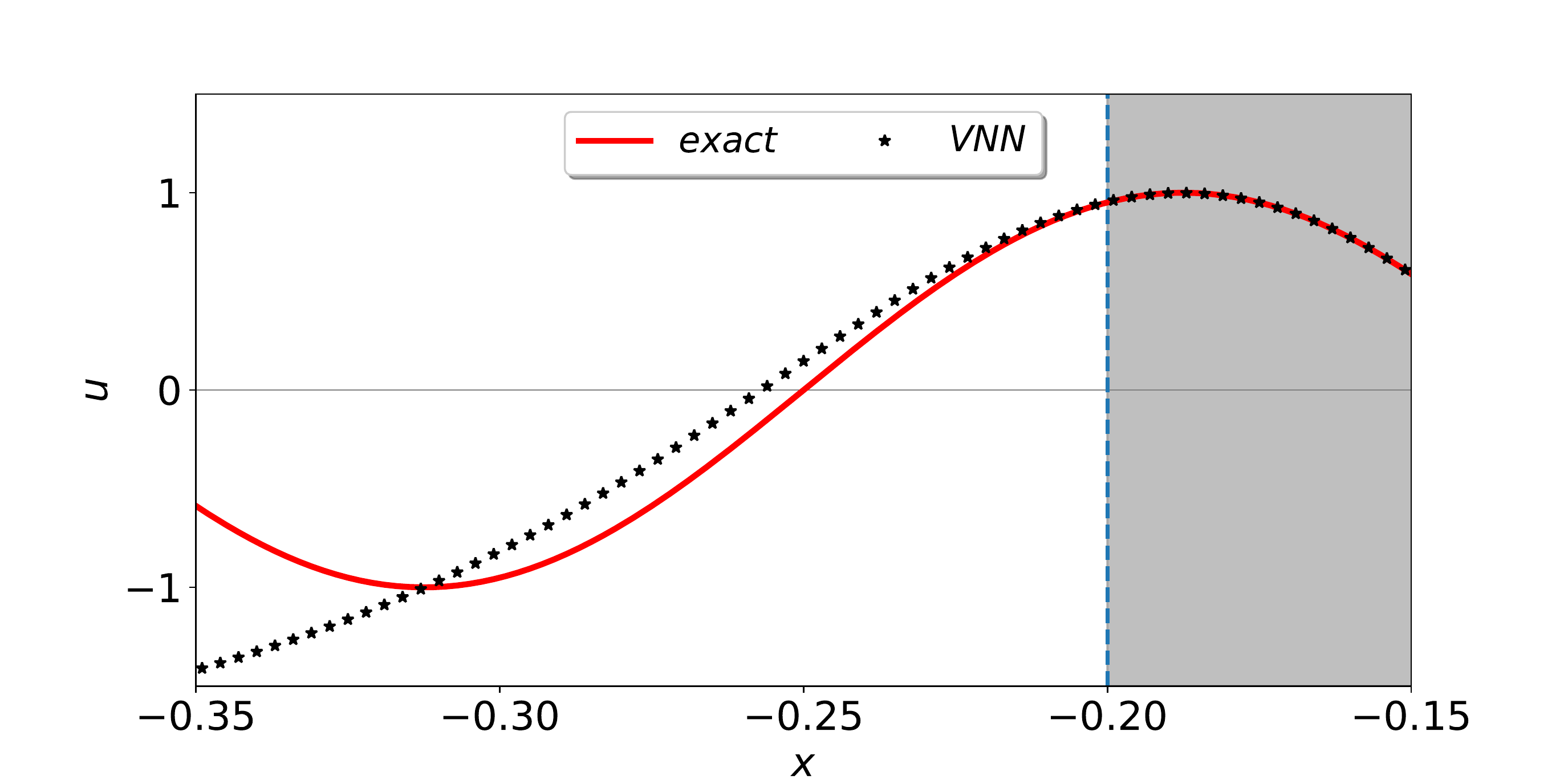}
	\includegraphics[clip, trim=0cm 0cm 0.5cm 0cm, width=0.46\linewidth]{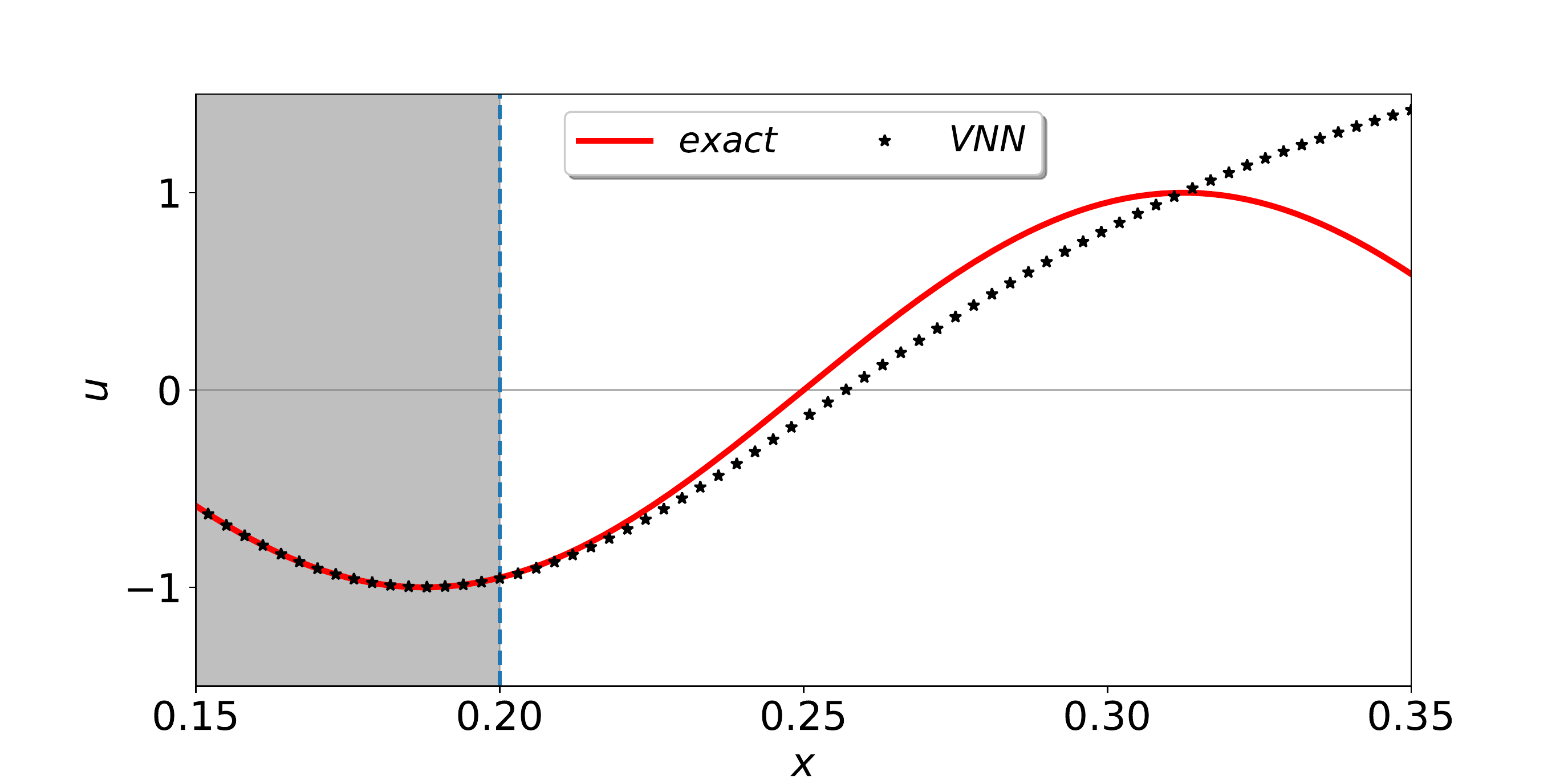}
	\vspace{-0.15 in}
	\caption{\scriptsize \label{Fig: VNN out element} VNN function approximation: learning out of the local element. The top left panel shows the exact function and VNN prediction. The shaded area is the sub-domain over which the test functions are locally defined and the network is trained. The bottom row shows the zoomed-in frame of the left and right boundaries of the sub-domain. The VNN parameters are $\ell = 4$, $\mathcal{N} = 20$, $K = 60$, $Q = 80$, and Adam optimizer with learning rate $10^{-3}$.  }
\end{figure}

\textbf{Learning Out-of-The-Box.} In the previous examples of local VNN, where we only train the network within a single sub-domain, we observe that the network can capture the target function with less accuracy slightly outside of that sub-domain. In fact, in addition to learning the target function, the network learns the derivative(s) of the target function within that sub-domain. The regularity of target function, localization of VNN, and structure of the network are important for more accurately predicting outside of the sub-domain. Figure \ref{Fig: VNN out element} shows an example, where the target function is $u^{exact} = \sin(8 \pi x)$. We define the local test functions over a symmetric sub-domain $[-0.2, 0.2]$; see the dashed blue line. We observe that after locally learning the function within this sub-domain, the network follows the same trend at the sub-domain boundaries and therefore extrapolates outside the sub-domain. A zoomed-in plot of the left and right boundaries of the sub-domain is shown in Fig. \ref{Fig: VNN out element}.
%

%
\section{One-Dimensional Poisson's Equation}
\label{Sec: VPINN Poisson examples}
%
Here, we discuss in detail the derivation of our proposed formulation \textit{hp}-VPINN for the one-dimensional problem. Let $u(x): \Omega \rightarrow \mathbb{R}$, where $\Omega = [-1,1]$. We consider the Poisson's equation given as
\begin{align}
\label{Eq: 1-d Poisson}
-& \frac{d^2 u(x)}{d x^2} = f(x), 
\\ 
\label{Eq: 1-d BVP BC}
& u(-1) = g, \quad u(1) = h, 
\end{align}
where $h$ and $g$ are constants and we assume the force term $f(x)$ is available at some quadrature points. Let the approximate solution be $u(x) \approx \tilde u(x) = u_{NN}(x)$, then the strong-form residual \eqref{Eq: residue strong form} becomes  
\begin{align}
\label{Eq: 1-d BVP residue}
r(x) &= -\frac{d^2 u_{NN}(x)}{d x^2} - f(x), \quad  x\in(-1,1) ,
\\ \nonumber
r_b(x) &=  u_{NN}(x) - u(x), \quad\quad\quad x = \pm 1.
\end{align}
We divide the domain $\Omega = [-1,1]$ into non-overlapping elements $\Omega_e = [x_{e-1},x_e]$ by defining a domain decomposition grid as $\lbrace -1 = x_0, x_1, \cdots, x_{N_{el}} = 1   \rbrace$. We choose a set of localized nonoverlapping test functions $v_k(x)$, given in \eqref{Eq: subdomain} with the nonvanishing function be the high-order polynomials. The variational residual then becomes
\begin{align}
\label{Eq: 1-d Poisson var residue element}
\mathcal{R}_k  & = \sum_{e=1}^{N_{el}} \mathcal{R}^{(e)}_k = \sum_{e=1}^{N_{el}}  \int_{x_{e-1}}^{x_e} 
\left( - \frac{d^2 u_{NN}(x)}{d x^2} - f(x) \right)\,  
v^{(e)}_k(x) \, dx ,
%
\end{align}
where, in each term $\mathcal{R}^{(e)}_k$, $e=1,2,\cdots,N_{el}$, the integral variable $x$ belongs to the sub-domain $ \Omega_e$. We can define the following three variational residual forms by integrating by parts the first term of $\mathcal{R}^{(e)}_k$. Thus, 
\begin{align}
\label{Eq: 1-d Poisson var residue element - 1}
\prescript{(1)}{}{\mathcal{R}}_{k}^{(e)} &= - \int_{x_{e-1}}^{x_e} \frac{d^2 u_{NN}(x)}{d x^2} \,  v^{(e)}_k(x) \, dx -  F_k^{(e)},
\\
\label{Eq: 1-d Poisson var residue element - 2}
\prescript{(2)}{}{\mathcal{R}}_{k}^{(e)} &= \,\,\,\,\,  \int_{x_{e-1}}^{x_e} \frac{d u_{NN}(x)}{d x} \,  \frac{d v^{(e)}_k(x)}{d x}  \, dx   
- \frac{d u_{NN}(x)}{d x} \, v^{(e)}_k(x) \bigg\vert_{x_{e-1}}^{x_{e}}- F_k^{(e)},
\\
\label{Eq: 1-d Poisson var residue element - 3}
\prescript{(3)}{}{\mathcal{R}}_{k}^{(e)} &= - \int_{x_{e-1}}^{x_e} u_{NN}(x)   \frac{d^2 v^{(e)}_k(x)}{d x^2} \, dx  
- \frac{d u_{NN}(x)}{d x} v^{(e)}_k(x) \bigg\vert_{x_{e-1}}^{x_{e}} 
+ u_{NN}(x) \frac{d v^{(e)}_k(x)}{d x} \bigg\vert_{x_{e-1}}^{x_{e}} - F_k^{(e)},
\end{align}
in which $F^{(e)}_k = \int_{x_{e-1}}^{x_e} f(x) \, v^{(e)}_k(x) \, dx $. Because $v^{(e)}_k(x)$ has a compact support over $ \Omega_e$, the first boundary term in \eqref{Eq: 1-d Poisson var residue element - 2} and \eqref{Eq: 1-d Poisson var residue element - 3} vanishes. The corresponding \emph{variational loss function} for each case takes the form
\begin{align}
\label{Eq: loss eVPINN}
&L^{\mathfrak{v}(i)} 
= \sum_{e=1}^{N_{el}} \, \frac{1}{K^{(e)}} \sum_{k = 1}^{K^{(e)}} \Big| \prescript{(i)}{}{\mathcal{R}}_{k}^{(e)}  \Big|^2 
+ \frac{\tau_b}{2} \left( \Big|u_{NN}(-1) - g \Big|^2 + \Big|u_{NN}(1) - h \Big|^2 \right),
\quad i=1,2,3,\quad
\end{align}
where $K^{(e)}$ is the number of test functions in element $e$.
For each element $e$, where $x \in [x_{e-1}, x_e]$, we transform the variational residual into the standard domain $\xi \in [-1,1]$ via a proper affine mapping to compute the integrals.

\subsection{Numerical Results}
%
We examine the performance of VPINN by considering different numerical examples. We construct a fully connected neural network with $\ell = 4$ layers and $\mathcal{N} = 20$ neurons in each layer with sine activation function. We employ up to order 60 Legendre polynomials and perform the integral using 80 Gauss-Lobatto quadrature points and weights (in each element). We write our formulation in Python, and employ Tensorflow to take advantage of its automatic differentiation capability. We also use the extended stochastic gradient descent Adam algorithm \cite{kingma2014adam} to optimize the loss function.

\begin{figure}[!ht]
	\center
	\begin{tabular}{c c c}
		\multicolumn{3}{c}{VPINN} \\ 	
		\hline \\ [-8 pt]
		%
		\includegraphics[clip, trim=0cm 0cm 2.5cm 1.2cm, width=0.3\linewidth]{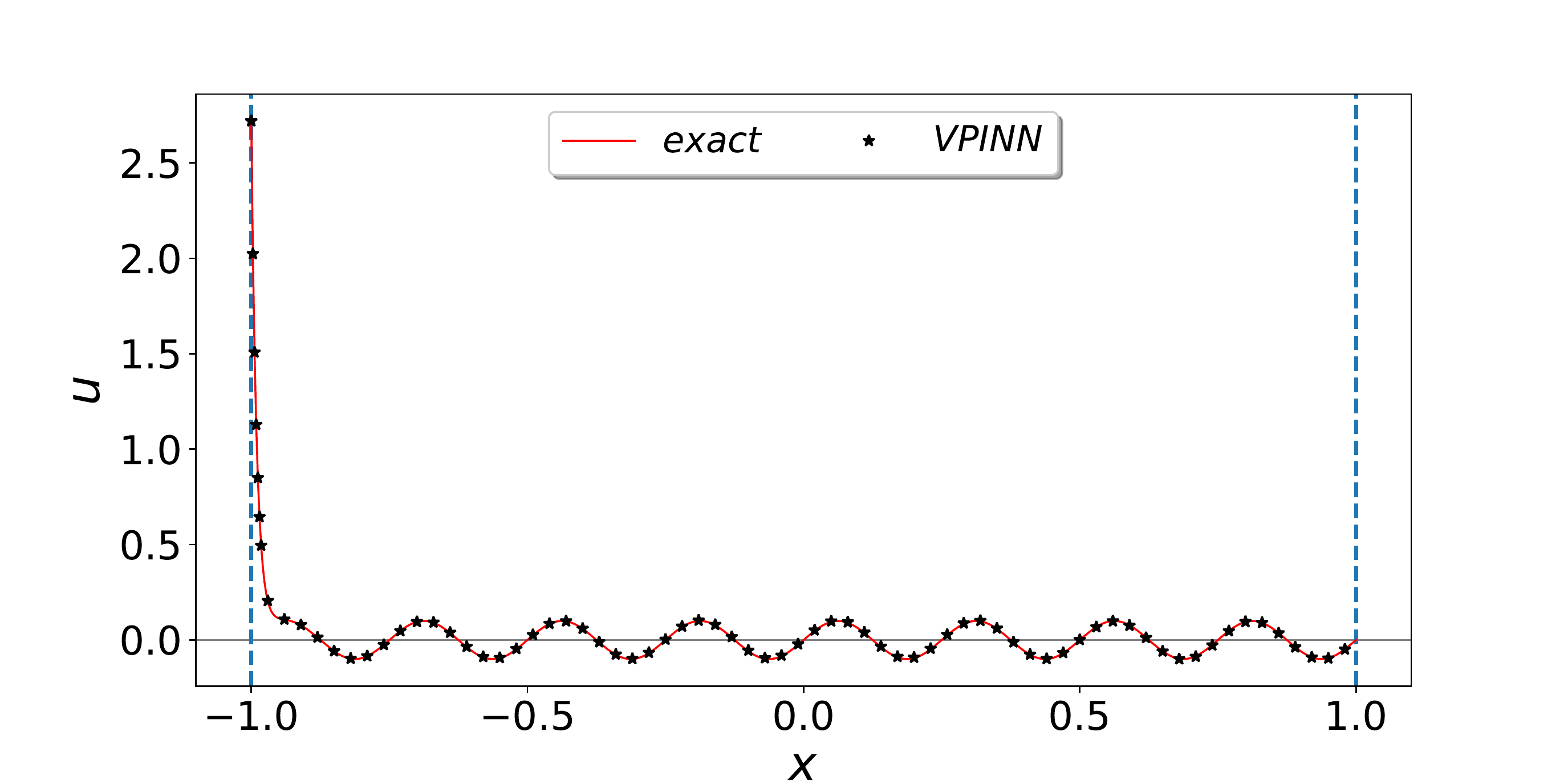}
		&
		\includegraphics[clip, trim=0cm 0cm 2.5cm 1.2cm, width=0.3\linewidth]{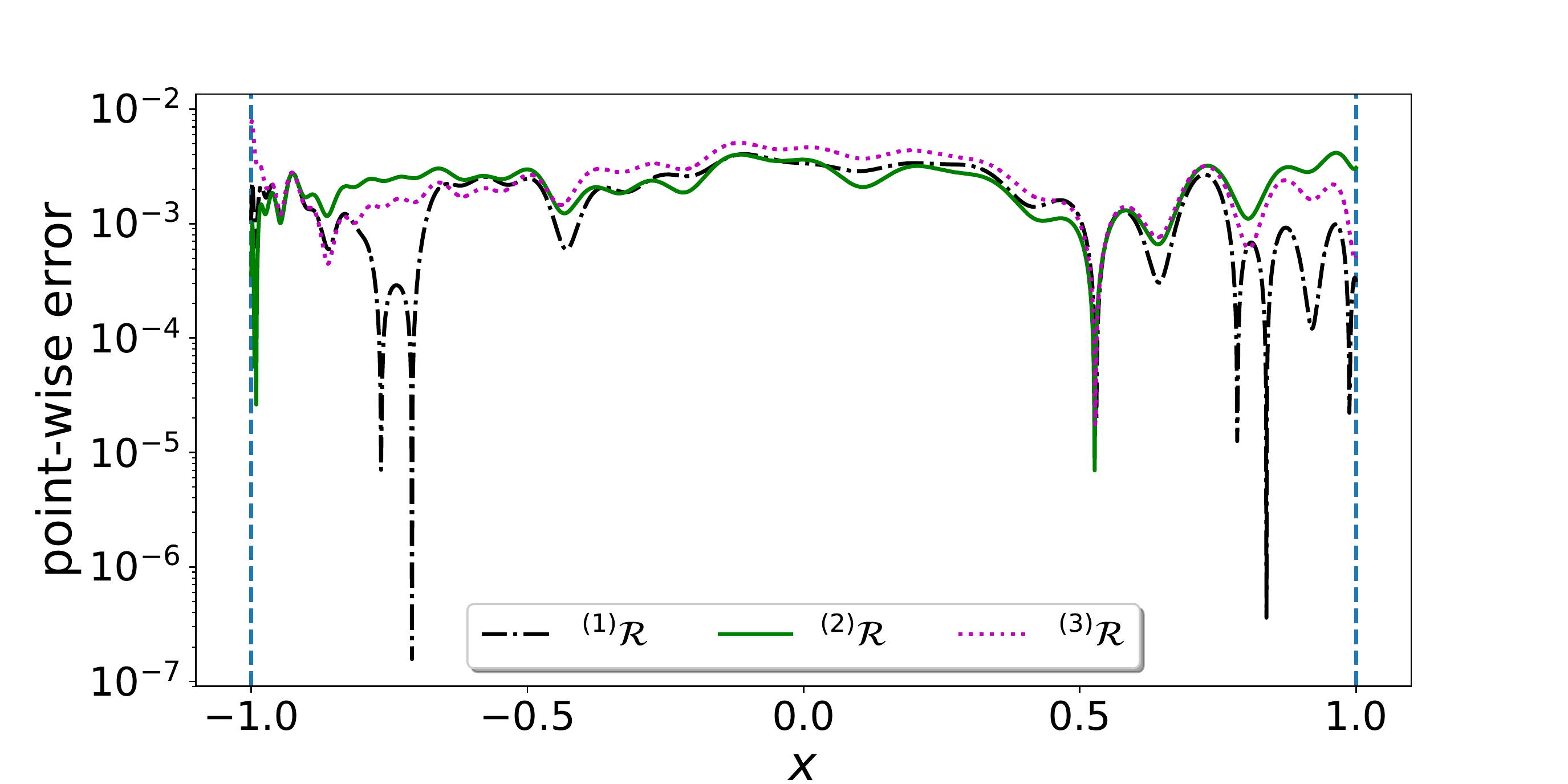}
		&
		\includegraphics[clip, trim=0cm 0cm 2.5cm 1.2cm, width=0.3\linewidth]{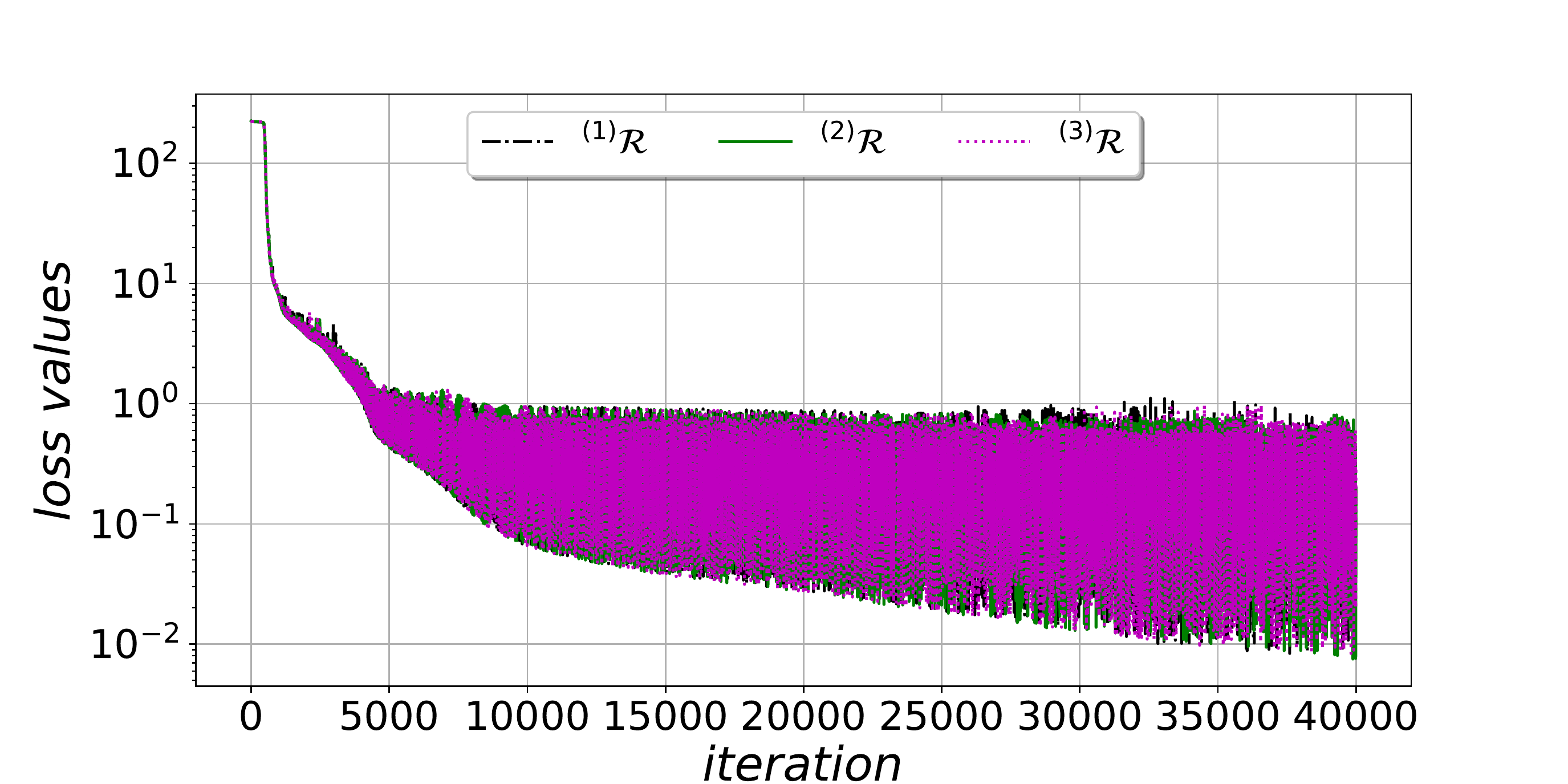}
		\\ [-7 pt]
		%
		\includegraphics[clip, trim=0cm 0cm 2.5cm 1.2cm, width=0.3\linewidth]{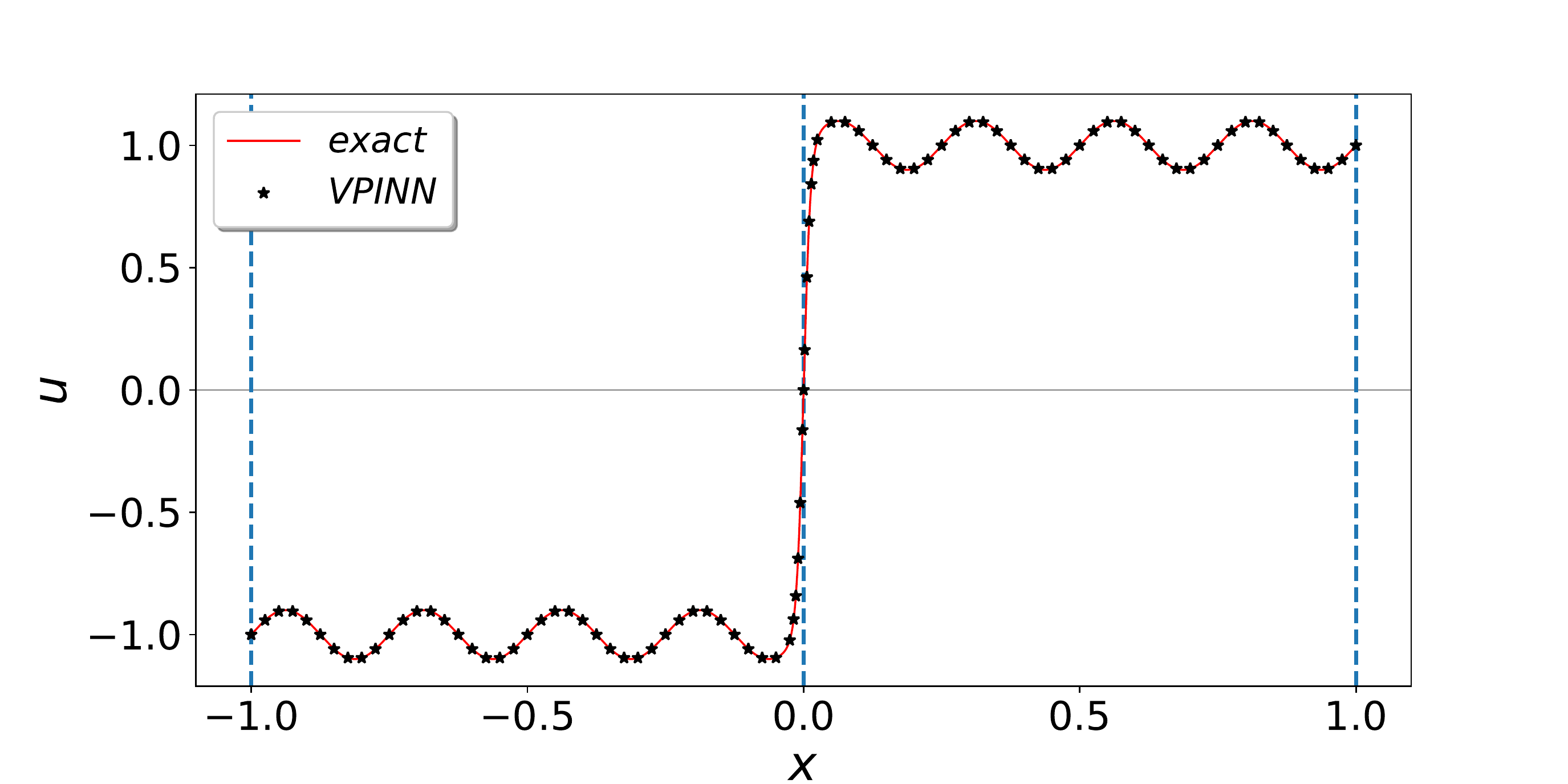}
		&
		\includegraphics[clip, trim=0cm 0cm 2.5cm 1.2cm, width=0.3\linewidth]{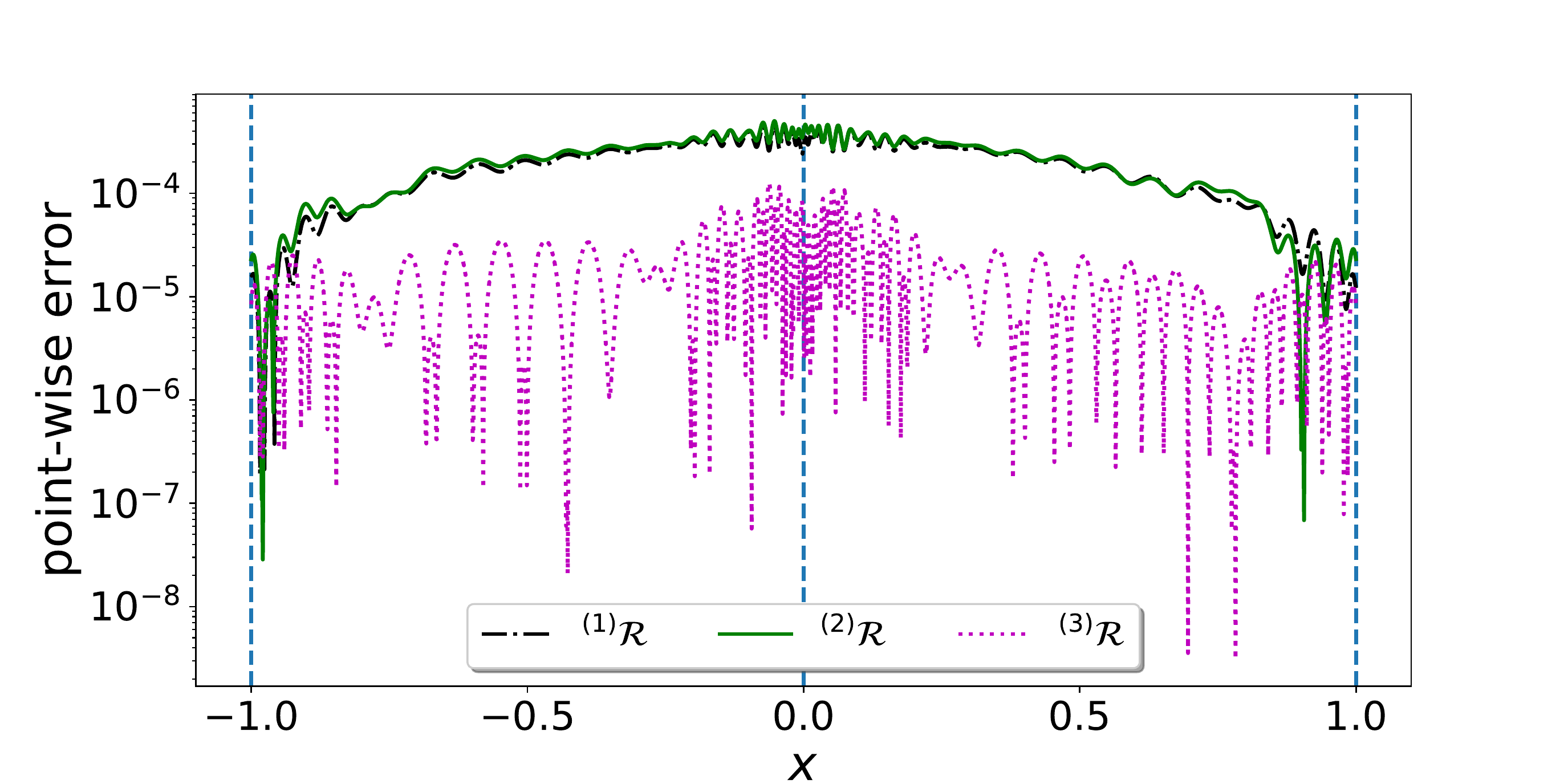}
		&
		\includegraphics[clip, trim=0cm 0cm 2.5cm 1.2cm, width=0.3\linewidth]{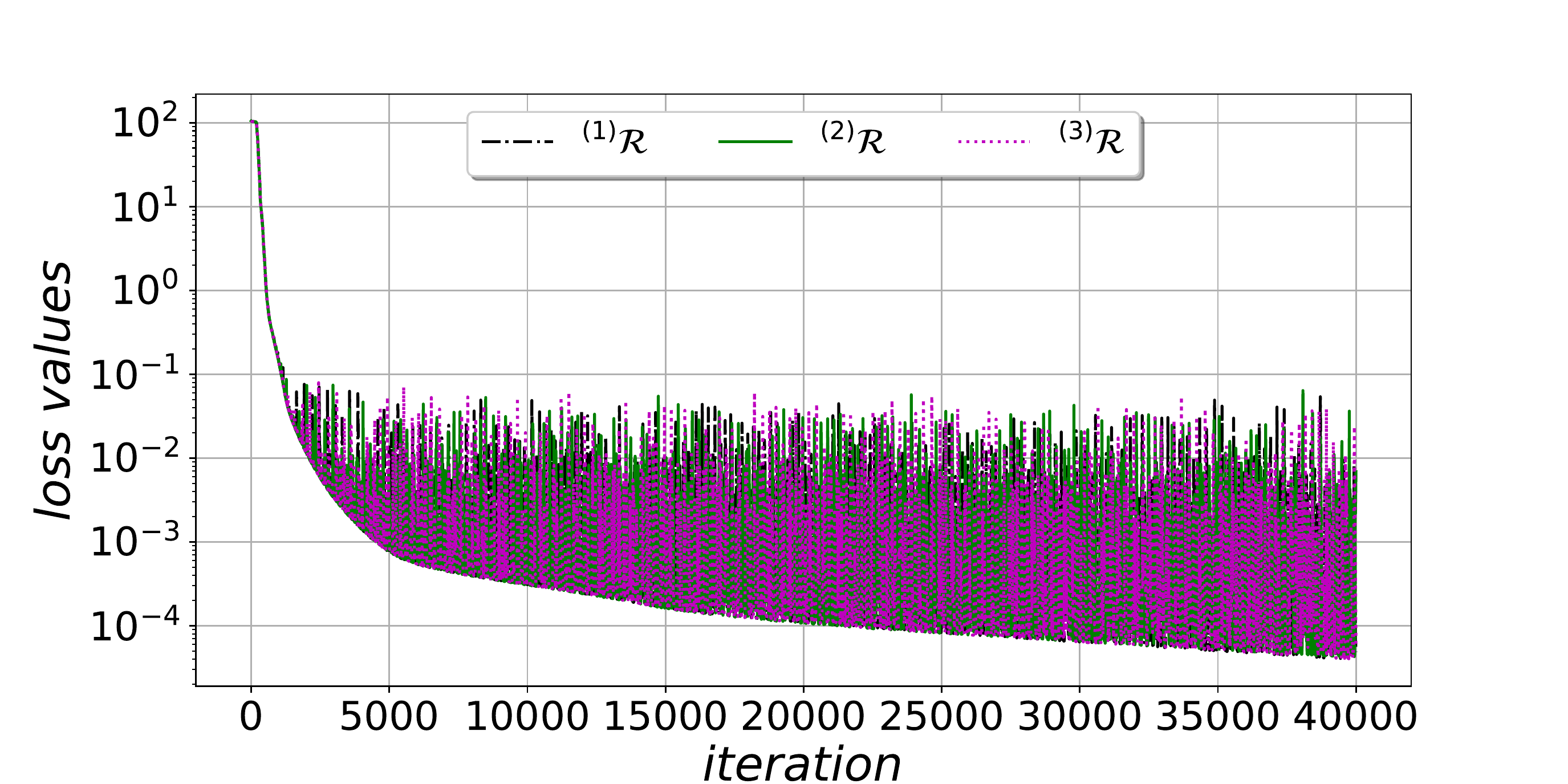}
		\\ [-7 pt]
		%
		\includegraphics[clip, trim=0cm 0cm 2.5cm 1.2cm, width=0.3\linewidth]{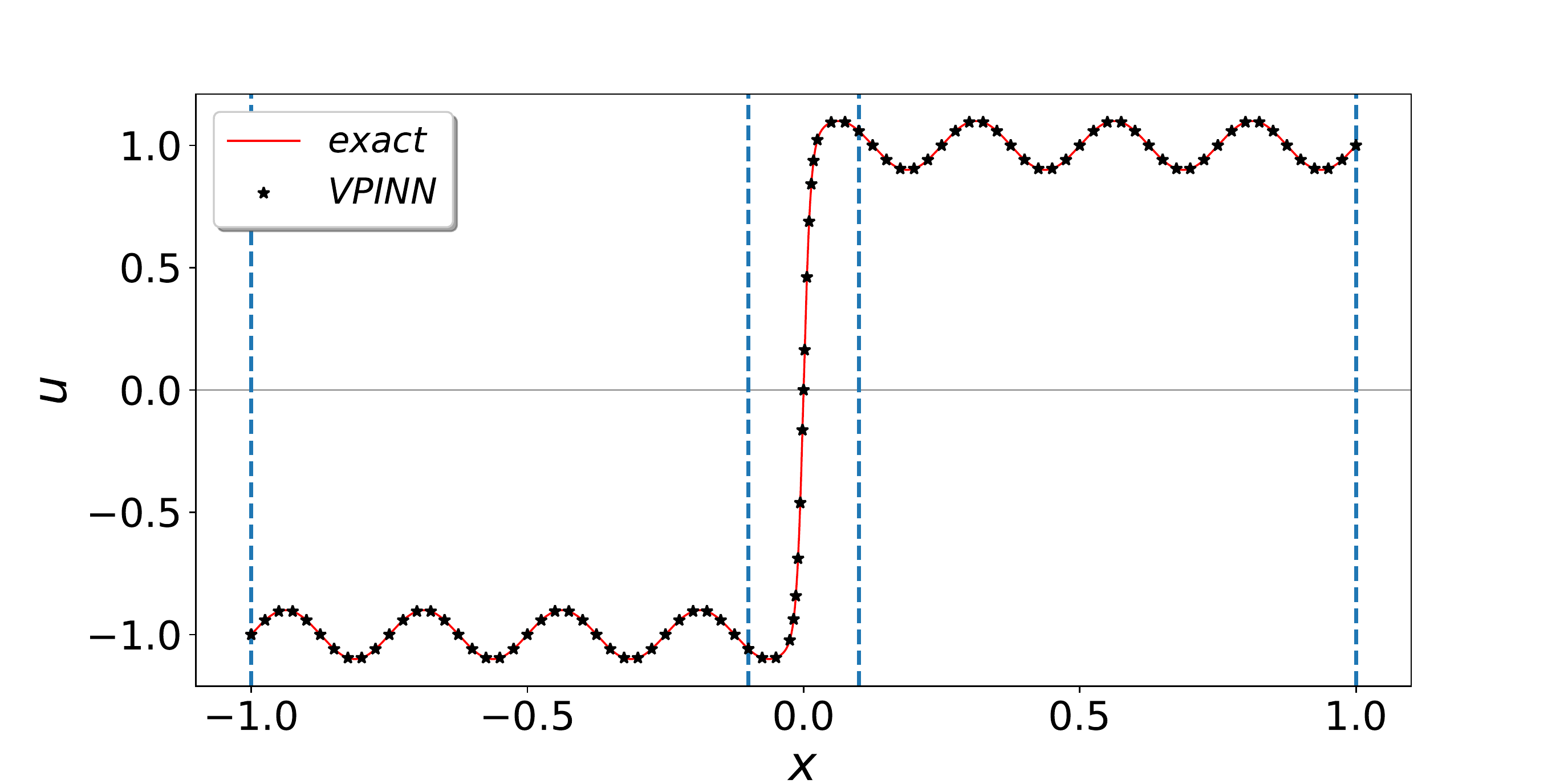}
		&
		\includegraphics[clip, trim=0cm 0cm 2.5cm 1.2cm, width=0.3\linewidth]{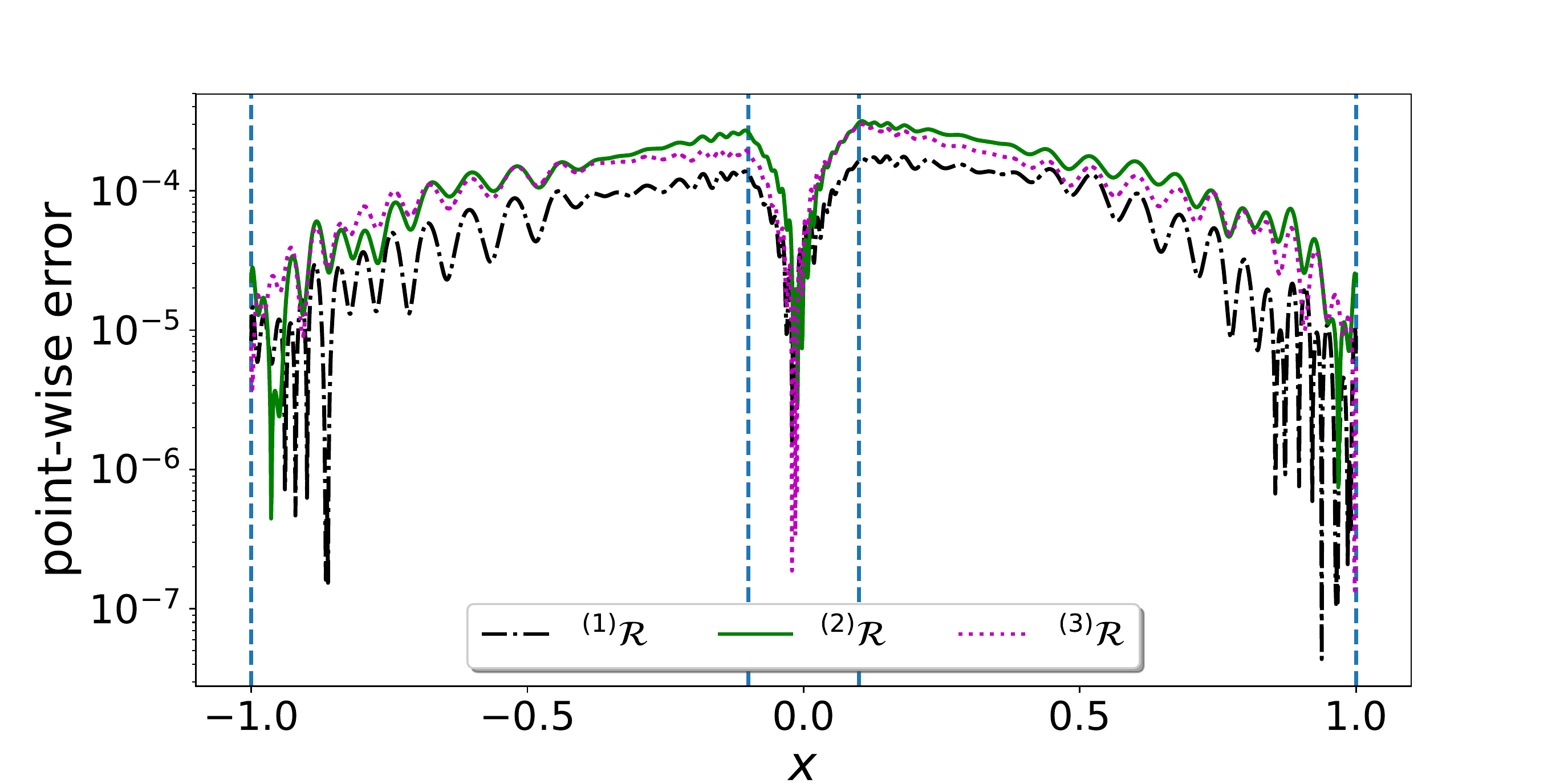}
		&
		\includegraphics[clip, trim=0cm 0cm 2.5cm 1.2cm, width=0.3\linewidth]{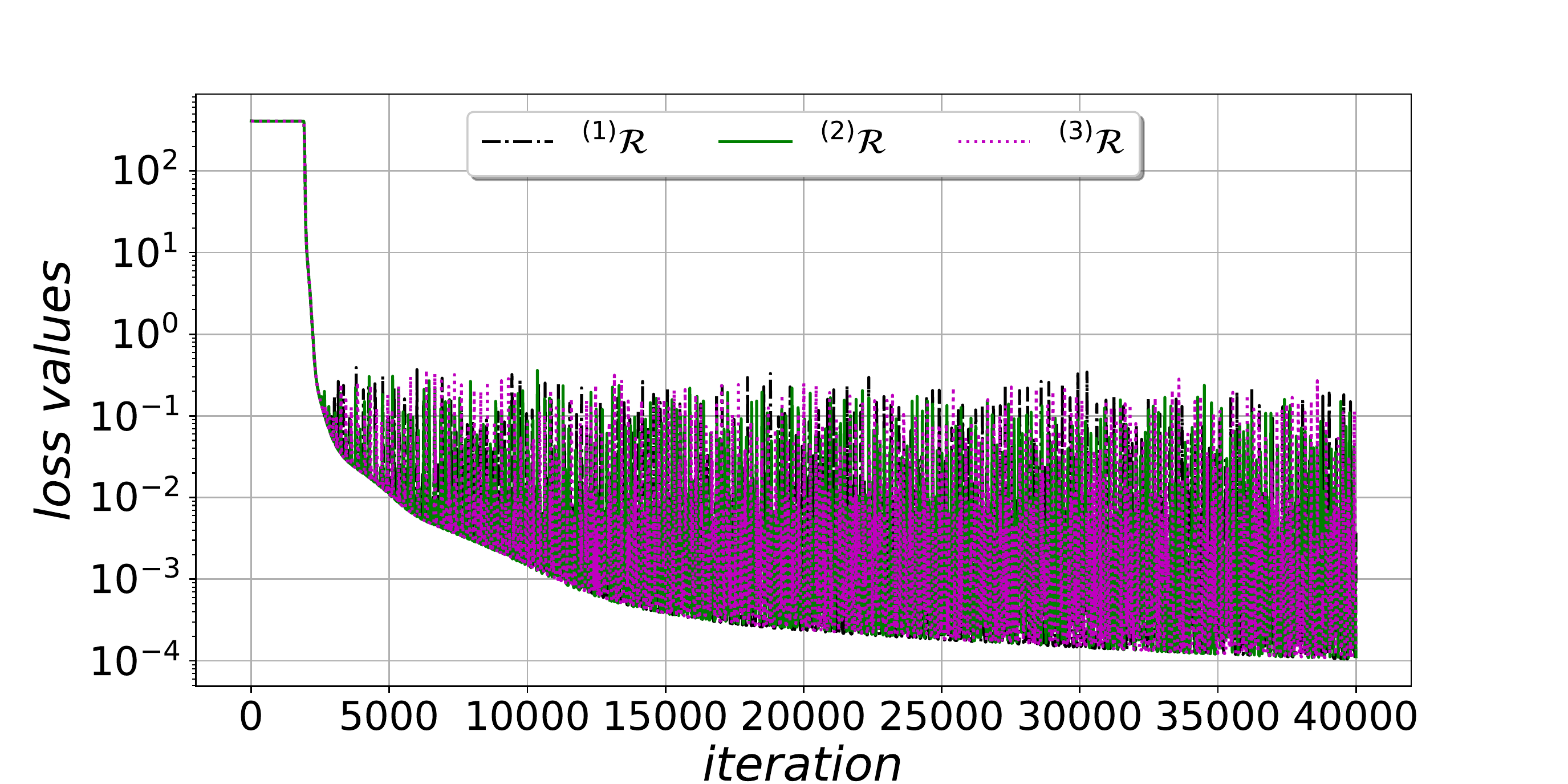}
		\\
		\multicolumn{3}{c}{PINN} \\  
		\hline \\ [-8 pt]
		%
		\includegraphics[clip, trim=0cm 0cm 2.5cm 1.2cm, width=0.3\linewidth]{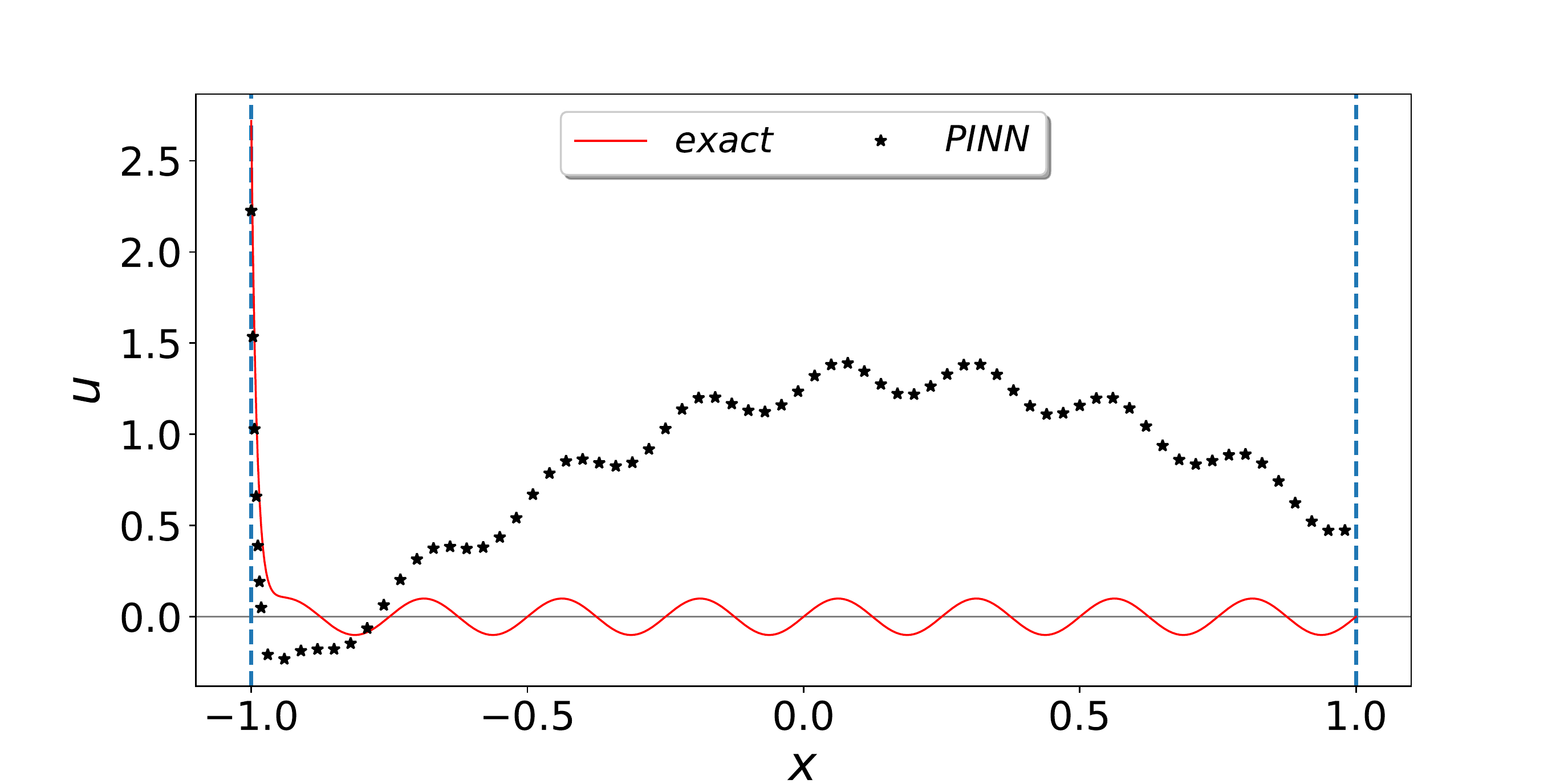}
		&
		\includegraphics[clip, trim=0cm 0cm 2.5cm 1.2cm, width=0.3\linewidth]{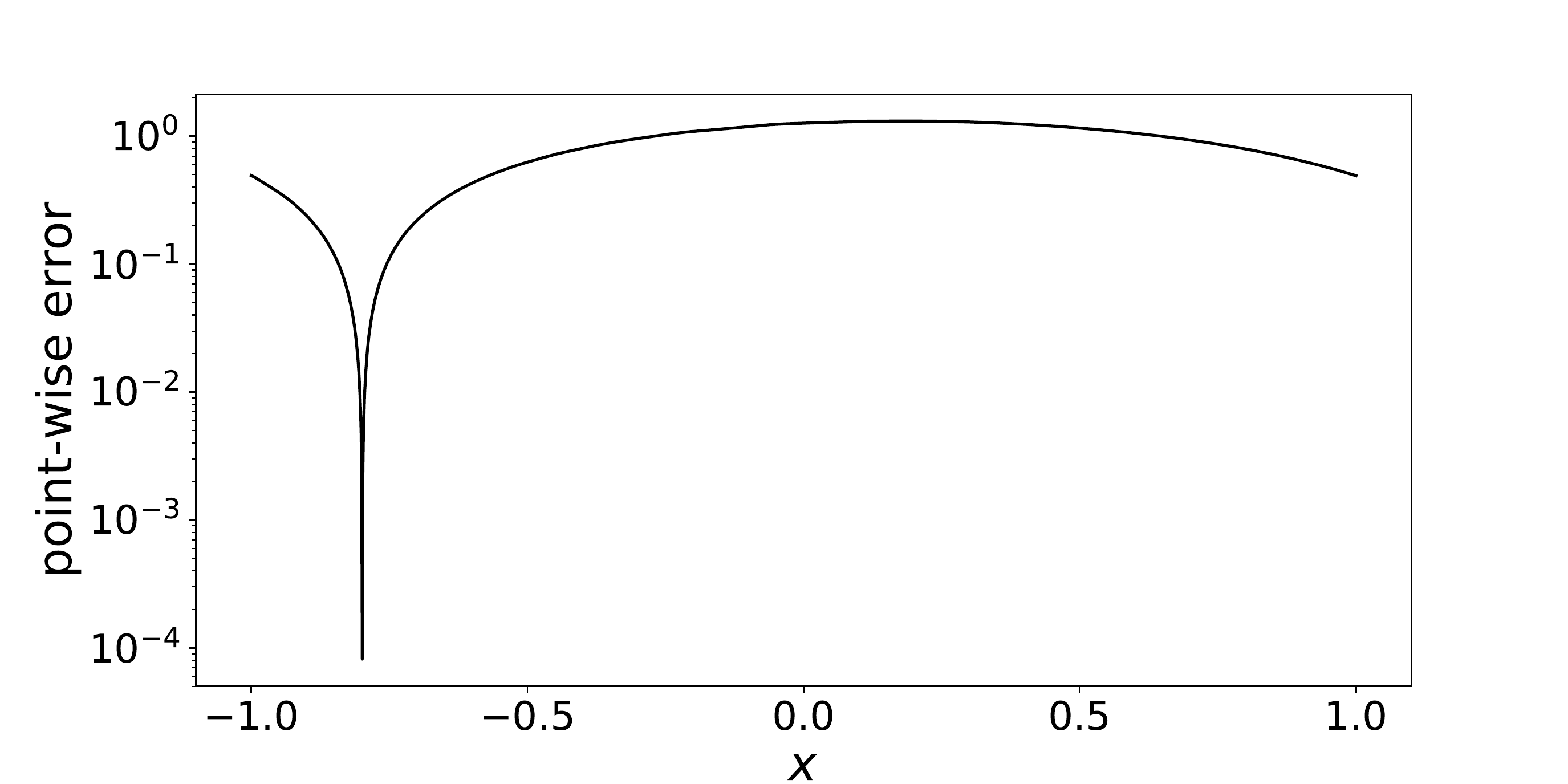}
		&
		\includegraphics[clip, trim=0cm 0cm 2.5cm 1.2cm, width=0.3\linewidth]{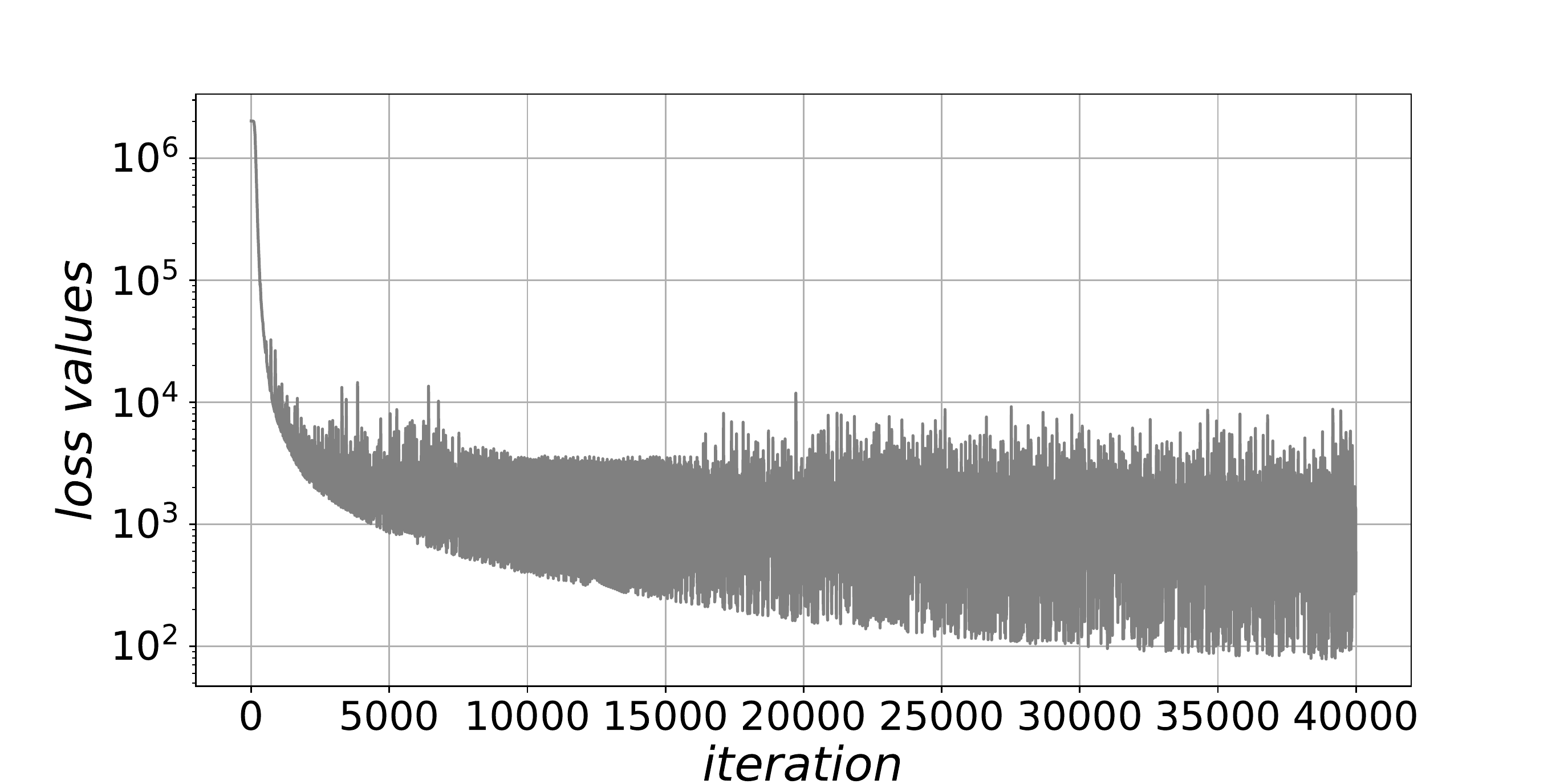}
	\end{tabular}
	\vspace{-0.15 in}
	\caption{\scriptsize \label{Fig: Poisson steep vPINN} One-dimensional Poisson's equation with steep solution \eqref{Eq: exact solution steep}: comparison of VPINN and PINN. Top panel shows the \textit{h}-refinement of VPINN with single element (first row), two elements (second row), and three elements (third row). Column-wise captions: (left) the exact solution and VPINN prediction, (middle) point-wise error, (right) loss function versus training iterations. The VPINN parameters are $\lbrace \ell = 4, \mathcal{N} = 20, \tau_ b =1 \rbrace$ and $\lbrace K = 60 , Q = 80 \rbrace$ in each element. The PINN parameters are $\lbrace \ell = 4, \mathcal{N} = 20, N_r = 500, \tau_ b =10 \rbrace$. The networks are fully connected with sine activation function and we use Adam optimizer with learning rate $10^{-3}$.}
\end{figure}

\begin{figure}[!ht]
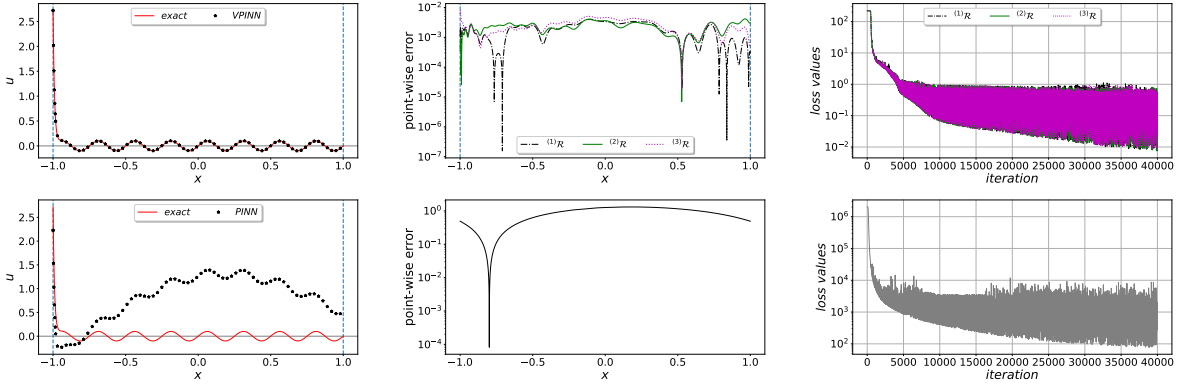

	\center
	\begin{tabular}{c c c}
		\includegraphics[clip, trim=0cm 0cm 2.5cm 1.2cm, width=0.3\linewidth]{VPINN_Poisson_testLegendre_Actsin_L4N20_NT60_NQ80_NE1_fcnplot.pdf}
		&
		\includegraphics[clip, trim=0cm 0cm 2.5cm 1.2cm, width=0.3\linewidth]{VPINN_Poisson_testLegendre_Actsin_L4N20_NT60_NQ80_NE1_pnterr_compareVF.pdf}
		&
		\includegraphics[clip, trim=0cm 0cm 2.5cm 1.2cm, width=0.3\linewidth]{VPINN_Poisson_testLegendre_Actsin_L4N20_NT60_NQ80_NE1_loss_compare.pdf}
		\\
		\includegraphics[clip, trim=0cm 0cm 2.5cm 1.2cm, width=0.3\linewidth]{PINN_Poisson_testLegendre_Actsin_L4N20_NF500_fcnplot.pdf}
		&
		\includegraphics[clip, trim=0cm 0cm 2.5cm 1.2cm, width=0.3\linewidth]{PINN_Poisson_testLegendre_Actsin_L4N20_NF500_pnterr.pdf}
		&
		\includegraphics[clip, trim=0cm 0cm 2.5cm 1.2cm, width=0.3\linewidth]{PINN_Poisson_testLegendre_Actsin_L4N20_NF500_loss.pdf}
	\end{tabular}
	\vspace{-0.15 in}
	\caption{\scriptsize \label{Fig: Poisson BL vPINN} One-dimensional Poisson's equation with boundary layer solution \eqref{Eq: exact solution BL-2}: comparison of VPINN (top row) and PINN (bottom row). Column-wise captions: (left) the exact solution and VPINN prediction, (middle) point-wise error, (right) loss function versus training iterations. The VPINN parameters are $\lbrace \ell = 4, \mathcal{N} = 20, K = 60 , Q = 80, \tau_ b =1 \rbrace$. The PINN parameters are $\lbrace \ell = 4, \mathcal{N} = 20, N_r = 500, \tau_ b =10 \rbrace$. The networks are fully connected with sine activation function, and we use Adam optimizer with learning rate $10^{-3}$.}
\end{figure}

\begin{exm}
	We solve the problem \eqref{Eq: 1-d Poisson}-\eqref{Eq: 1-d BVP BC} with exact solutions of the form
	\begin{align}
	\label{Eq: exact solution steep}
	\text{steep solution:  }&  u^{exact}(x) =  0.1 \sin(8 \pi x) + \tanh(80 x),
	\\
	\label{Eq: exact solution BL-2}
	\text{boundary layer solution:  }&  u^{exact}(x) = 0.1 \sin(5 \pi x) + e^{\frac{0.01-(x+1)}{0.01}}.
	\end{align}
	In each case, we obtain the force term by substituting the exact solution in \eqref{Eq: 1-d Poisson}. The results are shown in Figs. \ref{Fig: Poisson steep vPINN} and \ref{Fig: Poisson BL vPINN}.
	
\end{exm}

%

%
%
Figure \ref{Fig: Poisson steep vPINN} shows the VPINN and PINN approximation to the Poisson's equation with steep solution \eqref{Eq: exact solution steep}. In VPINN, we see that the point-wise error is oscillatory, which is expected due to the modal nature of test functions. Compared with PINN results, the error is orders of magnitude less, yet it does not oscillate in PINN. Similar behavior is observed in the other example of boundary layer exact solution, shown in Fig. \ref{Fig: Poisson BL vPINN}. It should be noted that for PINN to accurately capture a sharp change in the solution, we need to provide a larger number of residual points especially closer to the location of sharp change. We also note that in the steep and boundary layer cases, the force term becomes very large, leading to a large loss value initially, which may sometimes results in an optimization failure. Unlike VPINN, we need to give a higher weight to the boundary term in the loss function in PINN to make sure that the network learns the boundary correctly. 

\begin{figure}[!ht]
	\center
	\includegraphics[clip, trim=0cm 0cm 2.5cm 1.2cm, width=0.47\linewidth]{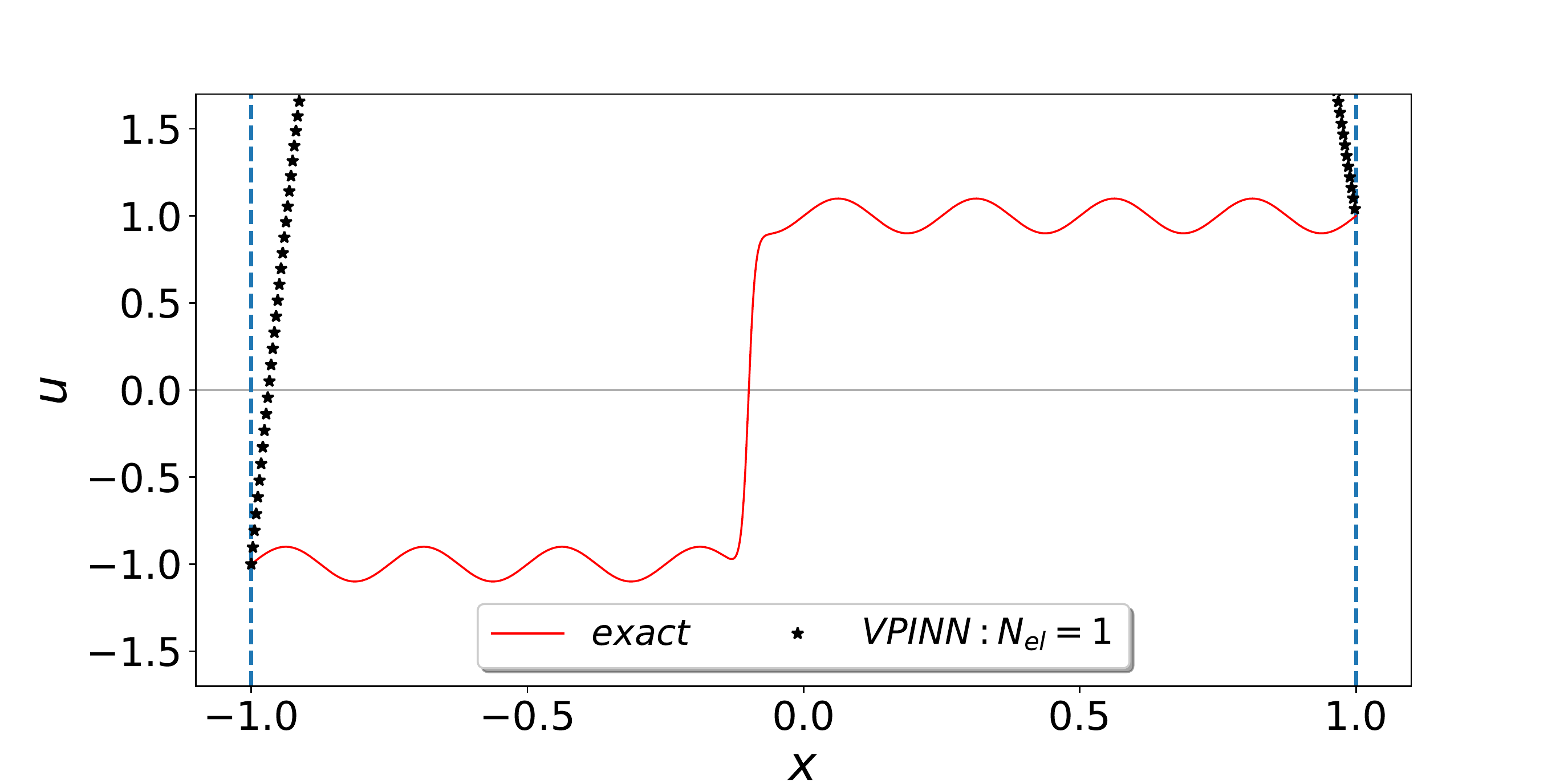}
	\includegraphics[clip, trim=0cm 0cm 2.5cm 1.2cm, width=0.47\linewidth]{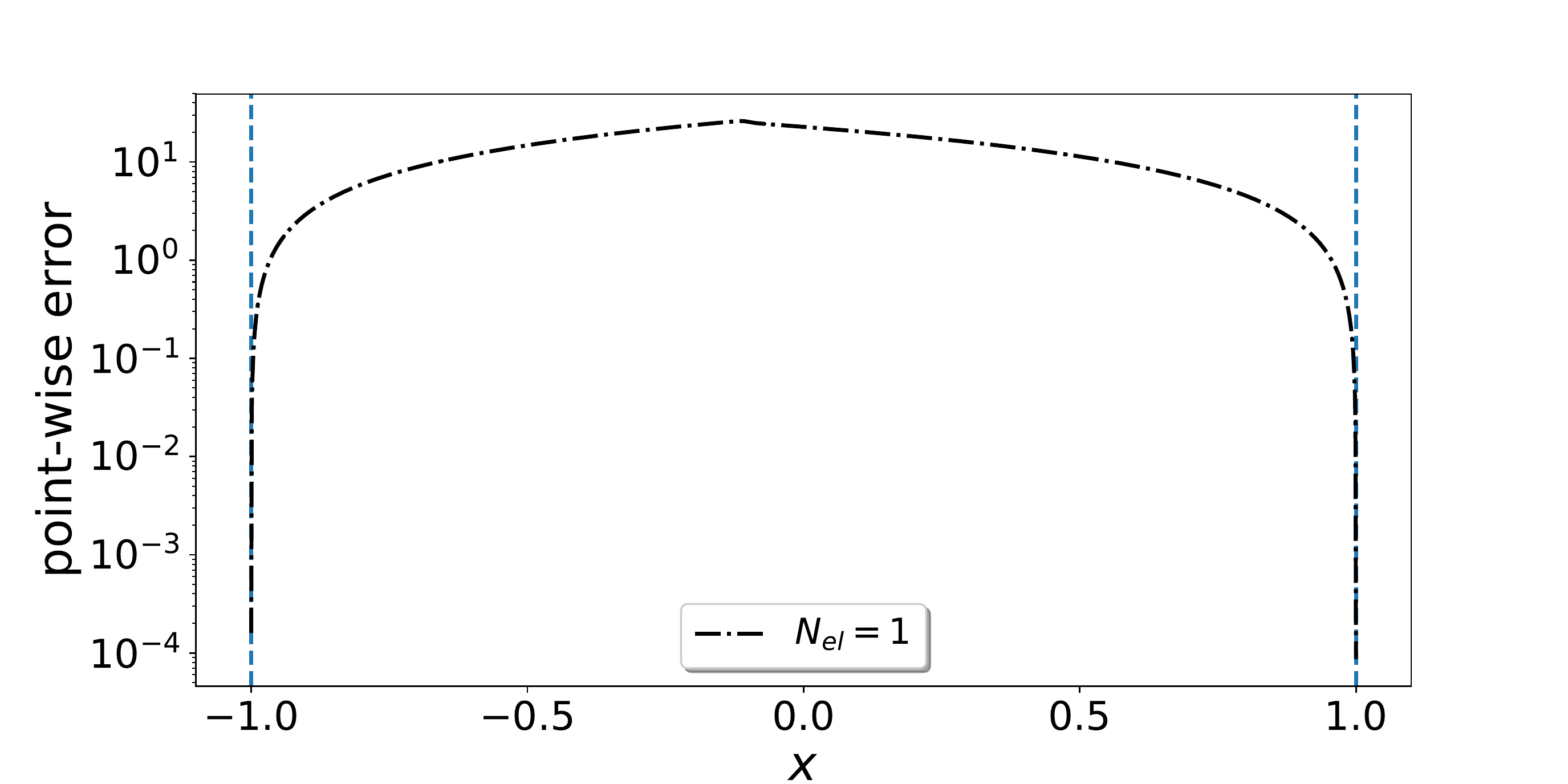}
	\\ 
	\includegraphics[clip, trim=0cm 0cm 2.5cm 1.2cm, width=0.47\linewidth]{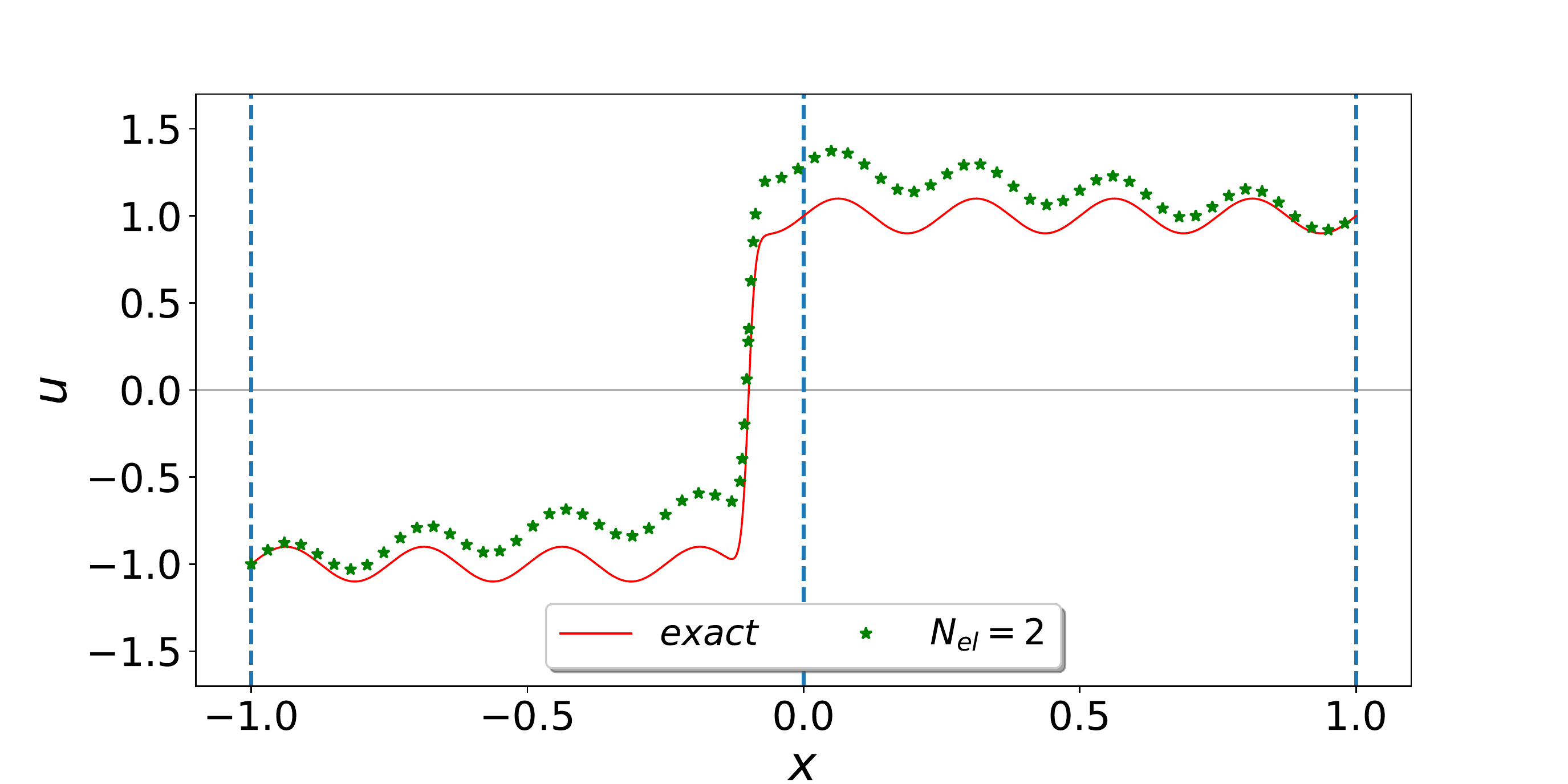}
	\includegraphics[clip, trim=0cm 0cm 2.5cm 1.2cm, width=0.47\linewidth]{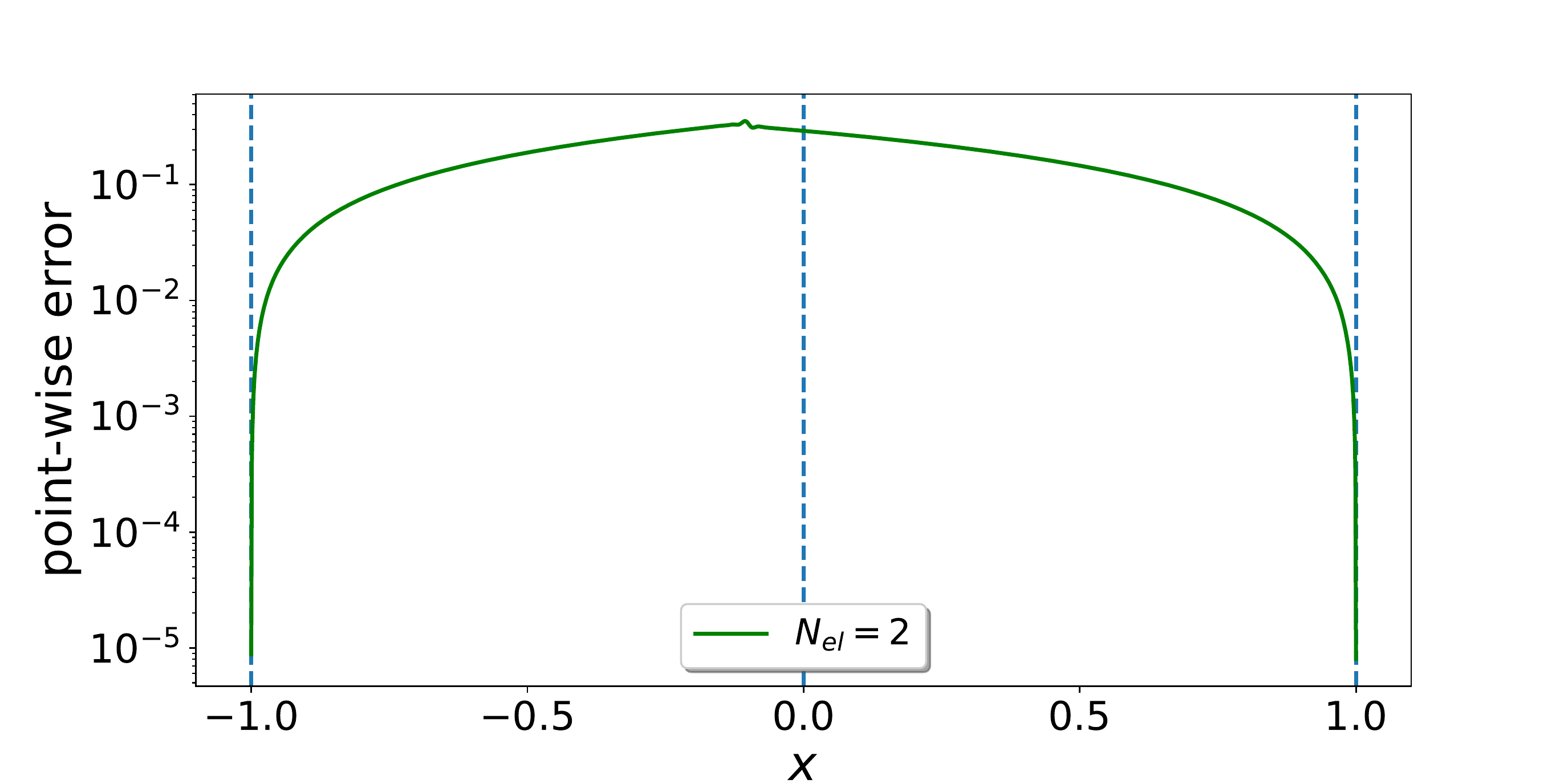}	
	\\ 
	\includegraphics[clip, trim=0cm 0cm 2.5cm 1.2cm, width=0.47\linewidth]{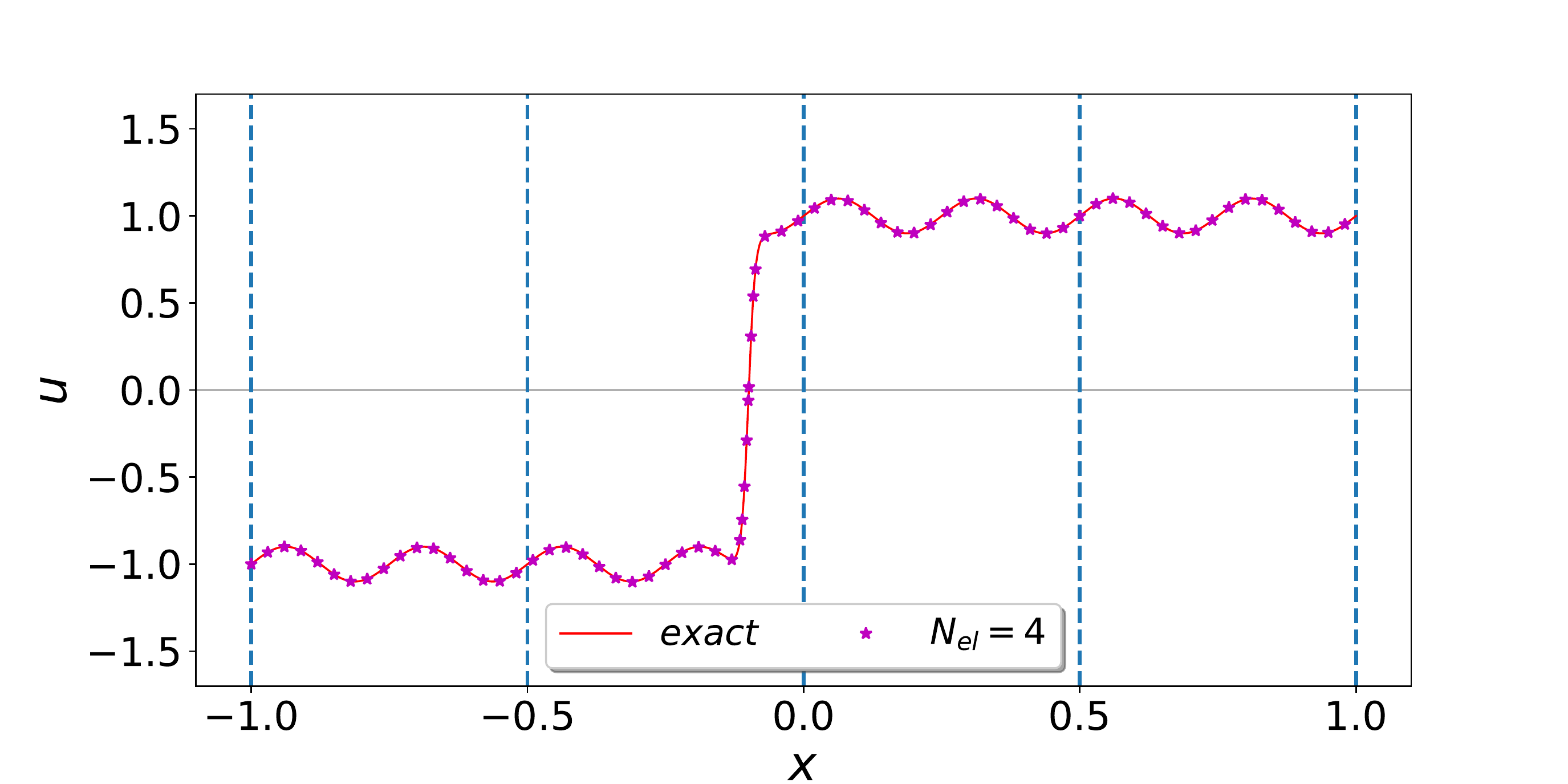}
	\includegraphics[clip, trim=0cm 0cm 2.5cm 1.2cm, width=0.47\linewidth]{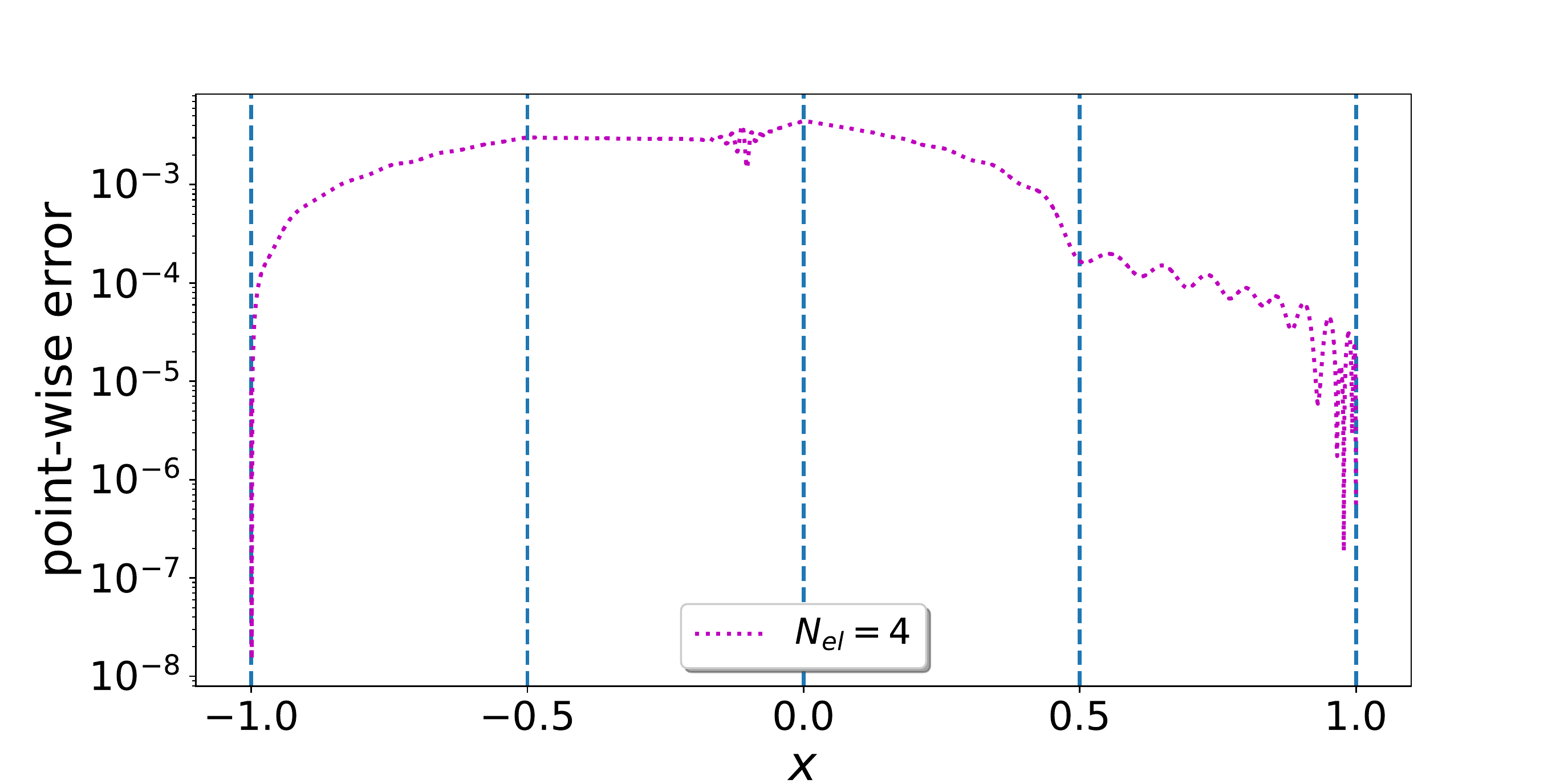}
	\vspace{-0.15 in}
	\caption{\scriptsize \label{Fig: Poisson assym steep evPINN} One-dimensional Poisson's equation with asymmetric steep solution \eqref{Eq: exact solution assym steep}. The location of sharp change is not known a priori and is predicted by increasing number of elements in VPINN.  Left column: the exact solution and VPINN prediction. Right column: point-wise error. The VPINN is based on the $\prescript{(1)}{}{\mathcal{R}}$ formulation and has the parameters $\lbrace \ell = 4, \mathcal{N} = 20, K = 60 , Q = 80, \tau_ b =1 \rbrace$. The network is fully connected with sine activation function, and we use Adam optimizer with learning rate $10^{-3}$.}
\end{figure}

\begin{exm}
	We solve the problem \eqref{Eq: 1-d Poisson}-\eqref{Eq: 1-d BVP BC} with asymmetric steep solution of the form
	\begin{align}
	\label{Eq: exact solution assym steep}
	\text{asymmetric steep solution:  }  u^{exact}(x) =  0.1 \sin(8 \pi x) + \tanh(80 (x + 0.1)),
	\end{align}
	where the sharp change happens slightly off the origin. We assume that the location of sharp change is not known a priori and is obtained by successive domain decompositions into larger number of sub-domains; the results are shown in Fig. \ref{Fig: Poisson assym steep evPINN}. 
	
\end{exm}

%

It is interesting to note that since the solution is asymmetric while the test functions are symmetric, the single-element VPINN does a pretty inaccurate approximation compared to the symmetric steep function \eqref{Eq: exact solution steep}. However, as we increase the number of elements, the network eventually captures the solution.

%
\section{Two-Dimensional Poisson's Equation}
\label{Sec: VPINN Poisson examples 2d}
%
Here, we discuss in detail the derivation of our proposed formulation \textit{hp}-VPINN for the two-dimensional problem. Let $u(x,y): \Omega \rightarrow \mathbb{R}$, where $\Omega = [-1,1] \times [-1,1]$. We consider the two-dimensional Poisson's equation
\begin{align}
\label{Eq: 2-d Poisson}
\nabla^2 u(x,y) = f(x,y), 
\end{align}
subject to Dirichlet boundary conditions $h(x,y)$, and we assume the force term $f(x,y)$ is available at some quadrature points. Let the approximate solution be $u(x,y) \approx \tilde u(x,y) = u_{NN}(x,y)$, then the strong-form residual \eqref{Eq: residue strong form} becomes  
\begin{align}
\label{Eq: residue strong form 2d poisson}
r(x, y) 
&= \nabla^2 u_{NN}(x, y) - f(x, y), \quad \forall \{x, y\} \in \{x^i_r, y^i_r \}_{i=1}^{N_r} \in \Omega,
\\ \nonumber
r_b(x, y) 
&= u_{NN}(x, y) - h(x,y) , \quad\quad \forall \{x, y\} \in \{x^i_b, y^i_b \}_{i=1}^{N_b} \in \partial \Omega.
\end{align}
We construct a discrete finite dimensional test space by choosing finite set of size $K_1$ and $K_2$ of admissible test functions in $x$ and $y$, respectively, and using the tensor product rule as 
\begin{align}
\label{Eq: test space 2d poisson}
\widetilde V = \text{span} \big\lbrace v_{k_1 k_2 } (x, y) 
& = \phi_{k_1}(x)\phi_{k_2}(y), \,\,
k_m  =1, 2, \cdots, K_m, \,\, m=1,2 \big\rbrace.
\end{align} 
The variational residual then reads as 
\begin{align}
\label{Eq: 2-d Poisson var residue}
\mathcal{R}_{k_1 k_2}  = \left( \nabla^2 u_{NN}(x, y) - f(x,y) \, , \, v_{k_1 k_2}(x,y) \right)_{\Omega} .
%
\end{align}

\noindent We define grids in $x$ and $y$ as $\lbrace -1 = x_0, x_1, \cdots, x_{N_{el_x}} = 1   \rbrace$ and $\lbrace -1 = y_0, y_1, \cdots, y_{N_{el_y}} = 1   \rbrace$, respectively, divide the domain $\Omega $ into structured sub-domains by constructing non-overlapping elements $\Omega_{e_x e_y} = [x_{e_x-1},x_{e_x}] \times [y_{e_y-1}, y_{e_y}], \,\, e_x = 1,2,\cdots, N_{el_x}, \,\, e_y = 1,2,\cdots, N_{el_y}$. Therefore, the variational residual becomes
\begin{align}
\label{Eq: 2-d Poisson var residue element}
\mathcal{R}_{k_1 k_2}  & = \sum_{e_x=1}^{N_{el_x}} \sum_{e_y=1}^{N_{el_y}} \mathcal{R}^{(e_x e_y)}_{k_1 k_2} ,
\\ \nonumber
\mathcal{R}^{(e_x e_y)}_{k_1 k_2} & =   \int_{x_{e_x-1}}^{x_{e_x}} \int_{y_{e_y-1}}^{y_{e_y}} 
\left(\nabla^2 u_{NN}(x, y) - f(x, y)\right)\,  \phi^{(e_x)}_{k_1}(x) \,  \phi^{(e_y)}_{k_2}(y) 
\, dx \, dy,
\end{align}
where $k_1 = 1,2,\cdots, K_1$ and $k_2 = 1,2,\cdots, K_2$. We note that we can employ different number of test functions in each element, however, for simplicity in the derivation of formulation, we assume that we have similar number of test functions in all elements. We also note that similar to the one-dimensional case, the local test functions $v^{(e)}_{k_1 k_2}(x, y)$ have the compact support over $ \Omega_e$. Therefore, for all elements, we have 
\begin{align}
\label{Eq: test boundary}
&\phi^{(e_x)}_{k_1}(x_{e_x-1}) = \phi^{(e_x)}_{k_1}(x_{e_x}) = 0, \quad k_1 = 1,2,\cdots, K_1,
\\ \nonumber
&\phi^{(e_y)}_{k_2}(y_{e_y-1}) = \phi^{(e_y)}_{k_2}(y_{e_y}) = 0, \quad k_2 = 1,2,\cdots, K_2.
\end{align}
By integrating by parts in the first term of $\mathcal{R}^{(e_x e_y)}_{k_1 k_2}$, we can define the following variational residual forms, in which the term $F_{k_1 k_2}^{(e_x e_y)}$ is associated with the integral of the force term $f(x,y)$. 
\begin{align}
\label{Eq: 2-d Poisson var residue element - 1}
\prescript{(1)}{}{\mathcal{R}}_{k_1 k_2}^{(e_x e_y)} = 
\int_{x_{e_x-1}}^{x_{e_x}} \int_{y_{e_y-1}}^{y_{e_y}} 
\left( \frac{\partial^2 u_{NN}}{\partial x^2} + \frac{\partial^2 u_{NN}}{\partial y^2}\right) \,  \phi^{(e_x)}_{k_1}(x) \,  \phi^{(e_y)}_{k_2}(y) 
\, dx \, dy - F_{k_1 k_2}^{(e_x e_y)}.&&
\end{align}
\begin{align}
\label{Eq: 2-d Poisson var residue element - 2}
&\prescript{(2)}{}{\mathcal{R}}_{k_1 k_2}^{(e_x e_y)} 
= 
- \int_{x_{e_x-1}}^{x_{e_x}}  \int_{y_{e_y-1}}^{y_{e_y}} 
\left( 
\frac{\partial u_{NN}}{\partial x} \,  \frac{d \phi^{(e_x)}_{k_1}(x)}{dx}  \, \phi^{(e_y)}_{k_2}(y) 
+
\frac{\partial u_{NN}}{\partial y} \,  \phi^{(e_x)}_{k_1}(x) \, \frac{d \phi^{(e_y)}_{k_2}(y)}{dy} 
\right)
\, dx \, dy
- F_{k_1 k_2}^{(e_x e_y)}.
\end{align}
\begin{align}
\label{Eq: 2-d Poisson var residue element - 3}
\prescript{(3)}{}{\mathcal{R}}_{k_1 k_2}^{(e_x e_y)} 
&=
\int_{x_{e_x-1}}^{x_{e_x}}  \int_{y_{e_y-1}}^{y_{e_y}} 
\left( 
u_{NN}(x,y) \,  \frac{d^2 \phi^{(e_x)}_{k_1}(x)}{dx^2}  \, \phi^{(e_y)}_{k_2}(y) 
+
u_{NN}(x,y) \,  \phi^{(e_x)}_{k_1}(x) \, \frac{d^2 \phi^{(e_y)}_{k_2}(y)}{dy^2} 
\right)
\, dx \, dy
\\ \nonumber
& -
\int_{y_{e_y-1}}^{y_{e_y}} 
\left( 
\frac{\partial u_{NN}}{\partial x} \,  \frac{d \phi^{(e_x)}_{k_1}(x)}{dx}  \bigg|_{x_{e_x-1}}^{x_{e_x}}  
\right)
\phi^{(e_y)}_{k_2}(y) \, dy
+ 
\int_{x_{e_x-1}}^{x_{e_x}} 
\left( 
\frac{\partial u_{NN}}{\partial y} \,  \frac{d \phi^{(e_y)}_{k_2}(y)}{dy}  \bigg|_{y_{e_y-1}}^{y_{e_y}}  
\right)
\phi^{(e_x)}_{k_1}(x) \, dx
\\ \nonumber
& - F_{k_1 k_2}^{(e_x e_y)}
.
\end{align}
We reduce the order of tensor ${\mathcal{R}}_{k_1 k_2}^{(e_x e_y)}$ to one by stacking its entries into vector of size $K_1 K_2$. Subsequently, we define the \emph{variational loss function} as
\begin{align}
\label{Eq: loss var elemental 2-d poisson}
&
L^{\mathfrak{v}(i)} = 
\sum_{e_x=1}^{N_{el_x}} \, \sum_{e_y=1}^{N_{el_y}} \, 
\frac{1}{K_1 K_2} \sum_{k = 1}^{K_1 K_2} \Big| \prescript{(i)}{}{\mathcal{R}}_{k}^{(e_x e_y)} \Big|^2
+ \tau_b \, \frac{1}{N_b} \sum_{i = 1}^{N_b} \Big|r_b(x^i_{b}, y^i_b) \Big|^2,
\quad i=1,2,3,
\end{align}
where the term $\prescript{(i)}{}{\mathcal{R}}_{k}^{(e_x e_y)}$ is the $k$-th entry of the corresponding reduced tensor associated with element $e_x e_y$, and $r_b$ has the same form as in \eqref{Eq: residue strong form}. We note that the integrals in the variational residuals can be mapped into standard element $\{\xi,\eta \} \in [-1,1] \times [-1,1]$ via proper affine mapping.

%
\subsection{Numerical results}
%
We examine the performance of VPINN by considering different numerical examples. We construct a fully connected neural network with different depth/width/activation functions. We employ Legendre polynomials in each direction $x$ and $y$, and perform the integral in each element by employing the proper number of Gauss quadrature points using tensor product rule. We write our formulation in Python, and employ Tensorflow to take advantage of its automatic differentiation capability. We also use the extended stochastic gradient descent Adam algorithm \cite{kingma2014adam} to optimize the loss function.

\begin{figure}[!ht]
	\center
	\begin{tabular}{c c c}
		\multicolumn{3}{c}{{\scriptsize  PINN}} \\  [-3 pt] 
		\hline \\ [-8 pt]
		
        \multicolumn{1}{l}{\, \textbf{A} \quad\quad\quad {\scriptsize exact solution}}
        &\multicolumn{1}{l}{\, \textbf{B} \quad\quad\quad {\scriptsize predicted solution}}
        &\multicolumn{1}{l}{\, \textbf{C} \quad\quad\quad {\scriptsize point-wise error}}\\ [-3 pt]		

		\includegraphics[clip, trim=0cm 2cm 0cm 3.5cm, width=0.3\linewidth]{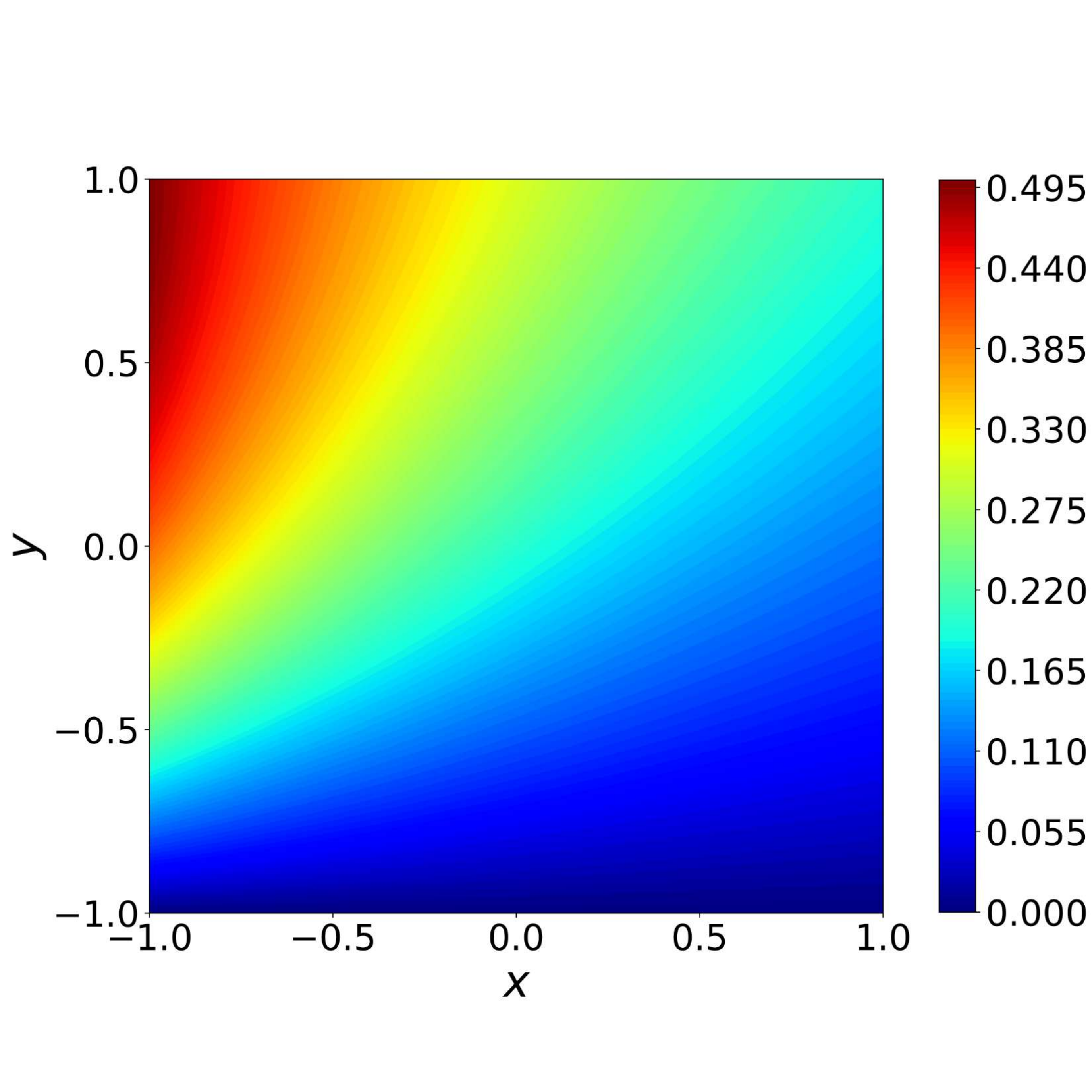}
		&
		\includegraphics[clip, trim=0cm 2cm 0cm 4.25cm, width=0.3\linewidth]{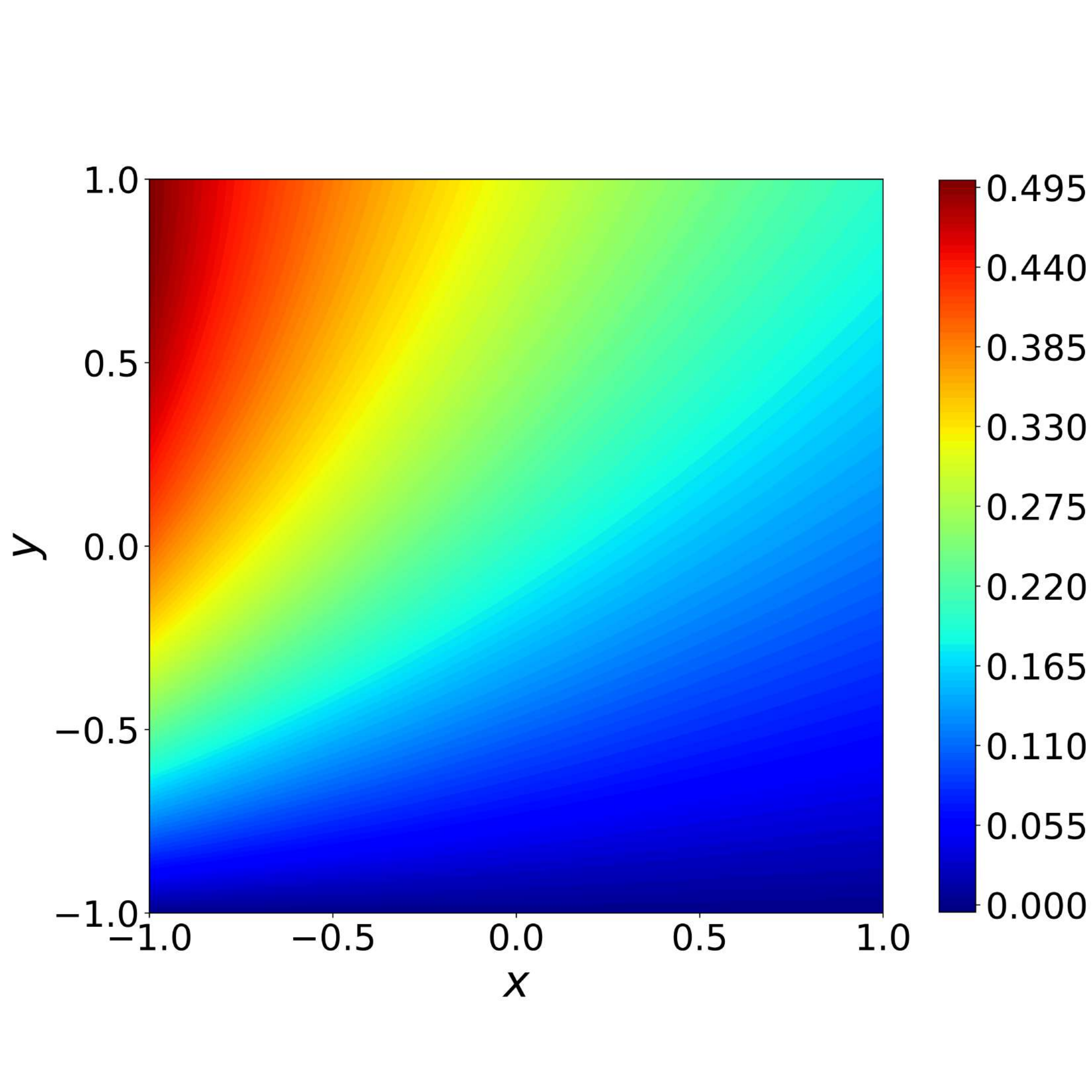}
		&
		\includegraphics[clip, trim=0cm 2cm 0cm 4.25cm, width=0.3\linewidth]{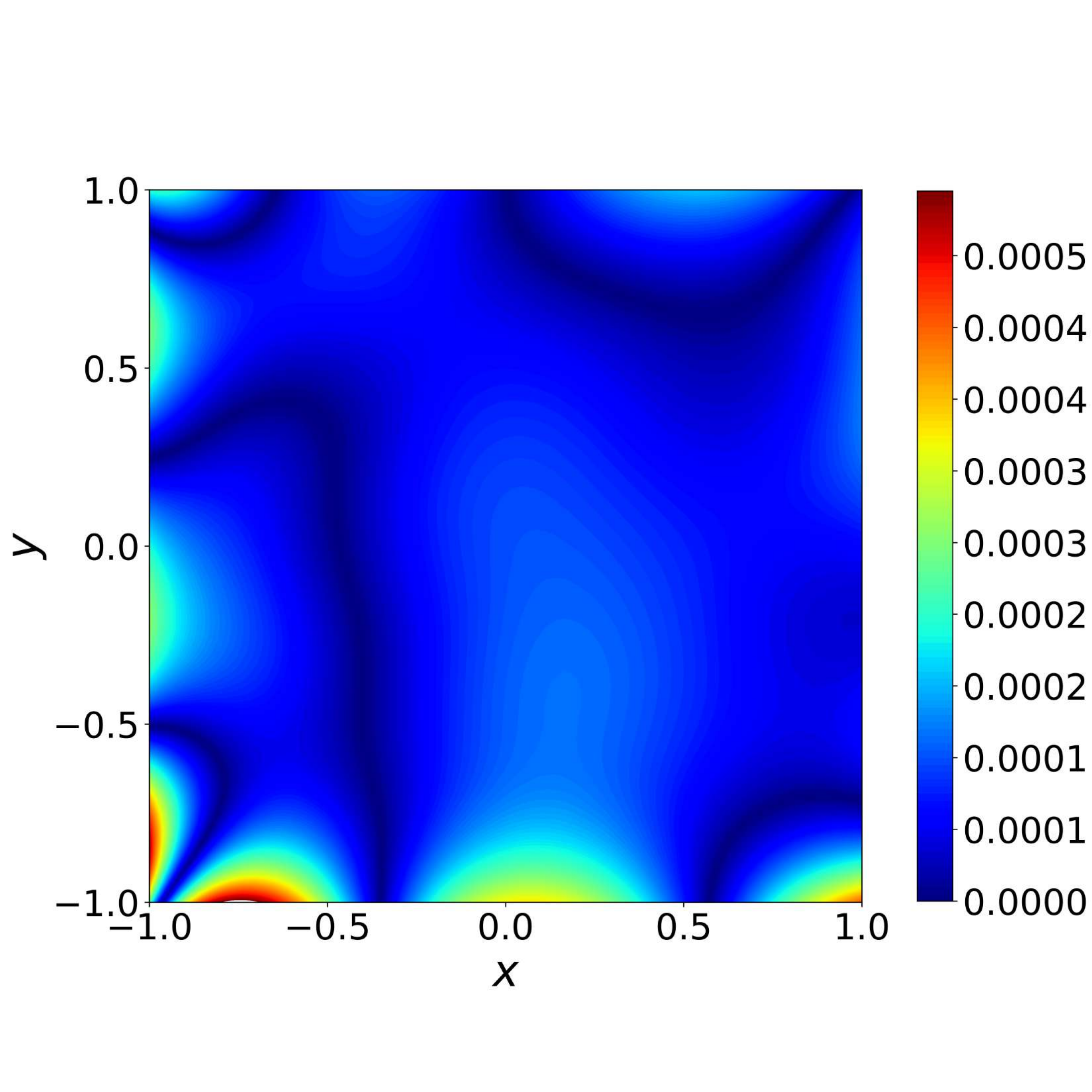}
		
		\\
		\multicolumn{3}{c}{{\scriptsize \quad \textit{hp}-VPINN: $\prescript{(2)}{}{\mathcal{R}} $ formulation}} \\  [-1 pt] 
		\hline \\ [-8 pt]
		
        \multicolumn{1}{l}{\,\,\, \textbf{D} \quad\quad {\scriptsize domain decomposition}}
        &\multicolumn{1}{l}{\, \textbf{E} \quad\quad\quad {\scriptsize predicted solution}}
        &\multicolumn{1}{l}{\, \textbf{F} \quad\quad\quad {\scriptsize point-wise error}}\\ [-1 pt]
        
		\includegraphics[clip, trim=0cm 1cm 0cm 3cm, width=0.27\linewidth]{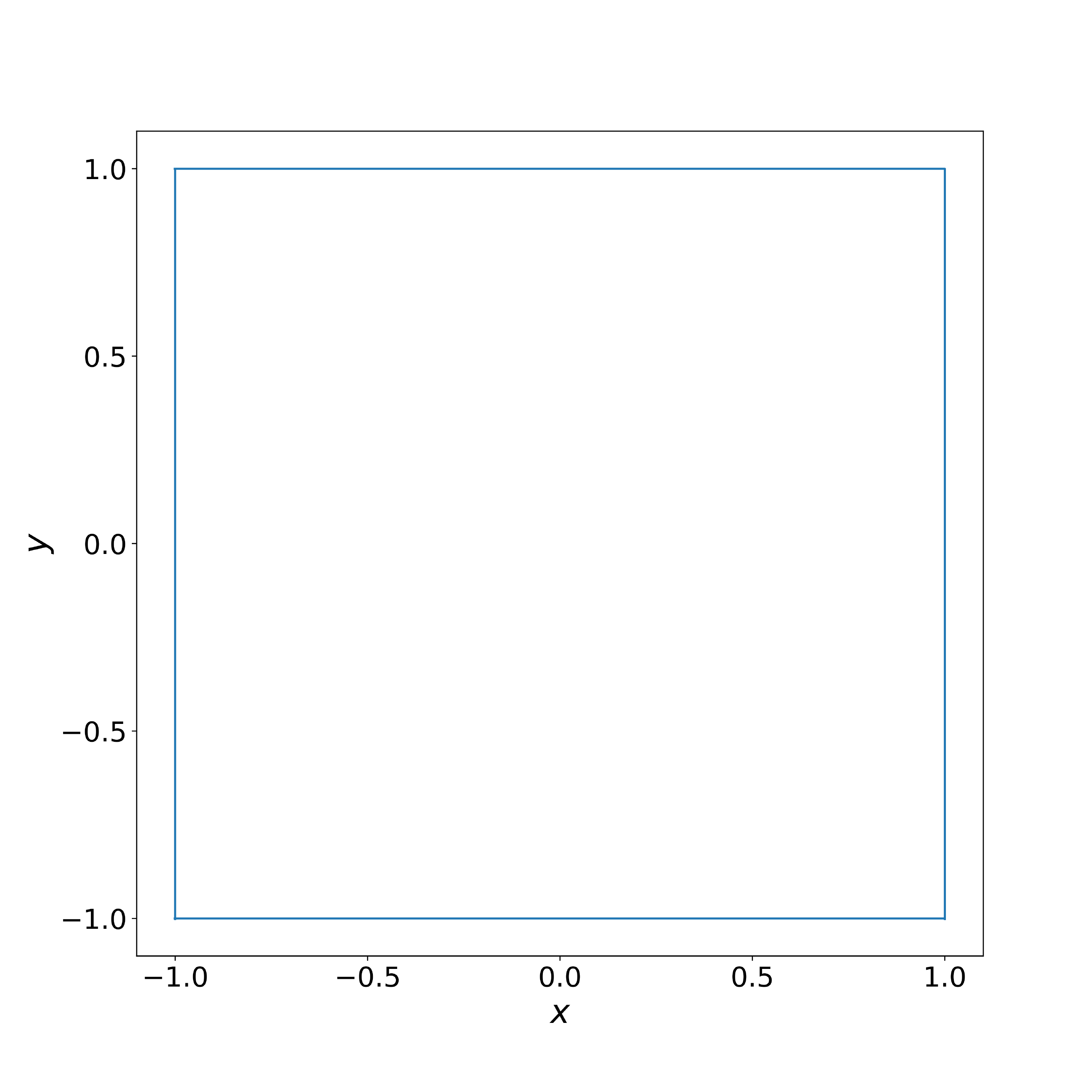}
		&
		\includegraphics[clip, trim=0cm 2cm 0cm 4.25cm, width=0.3\linewidth]{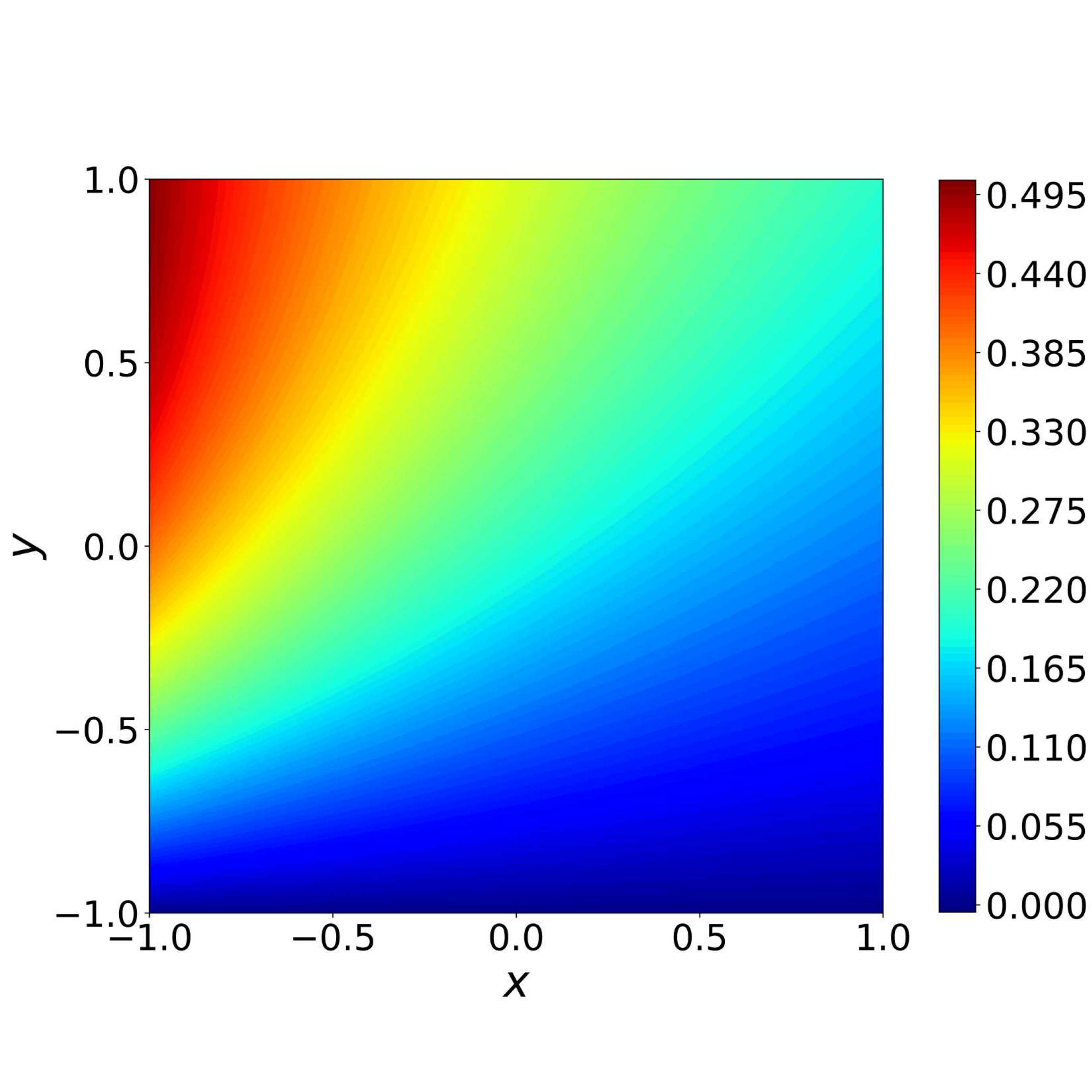}
		&
		\includegraphics[clip, trim=0cm 2cm 0cm 4.25cm, width=0.3\linewidth]{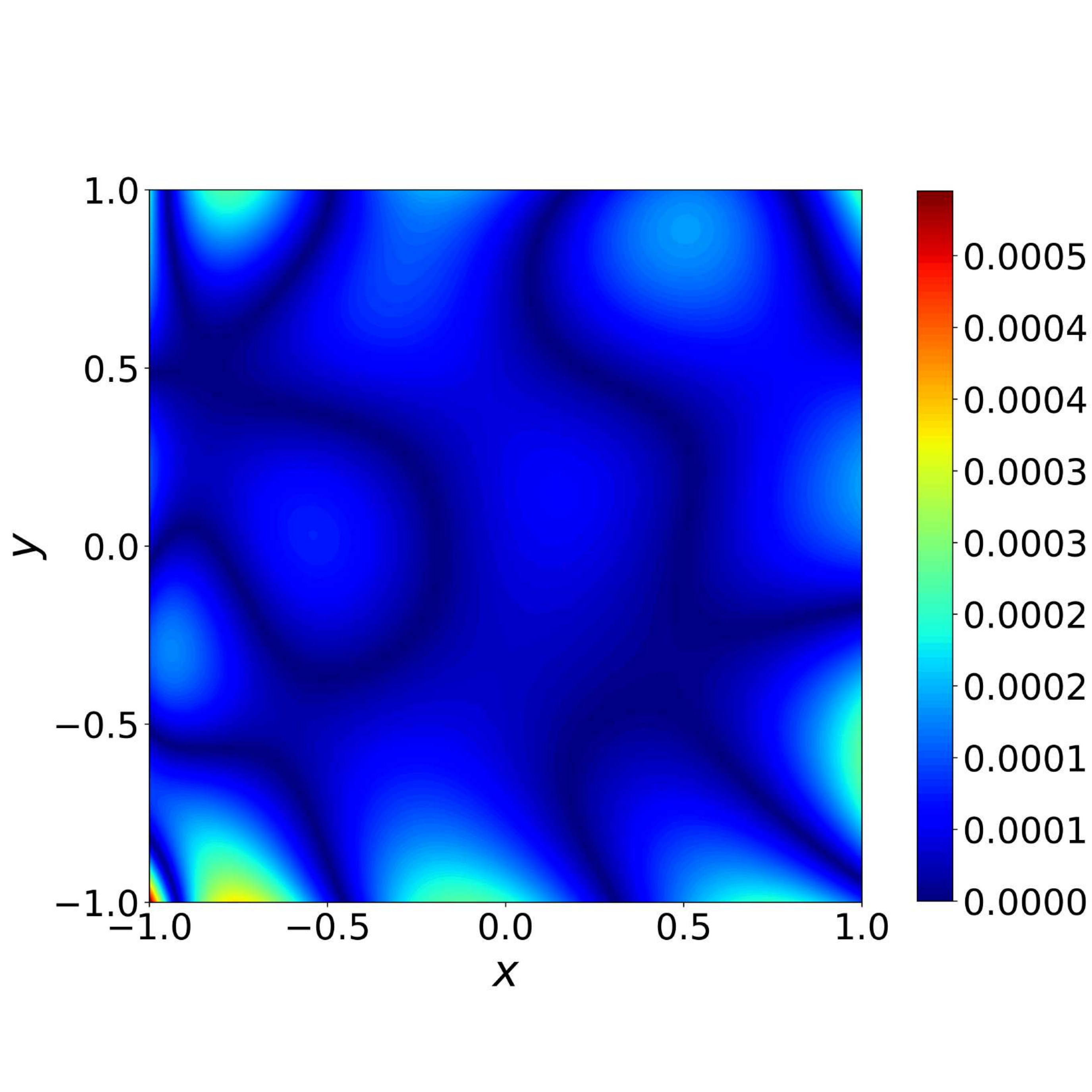}\\

		\includegraphics[clip, trim=0cm 1cm 0cm 3cm, width=0.27\linewidth]{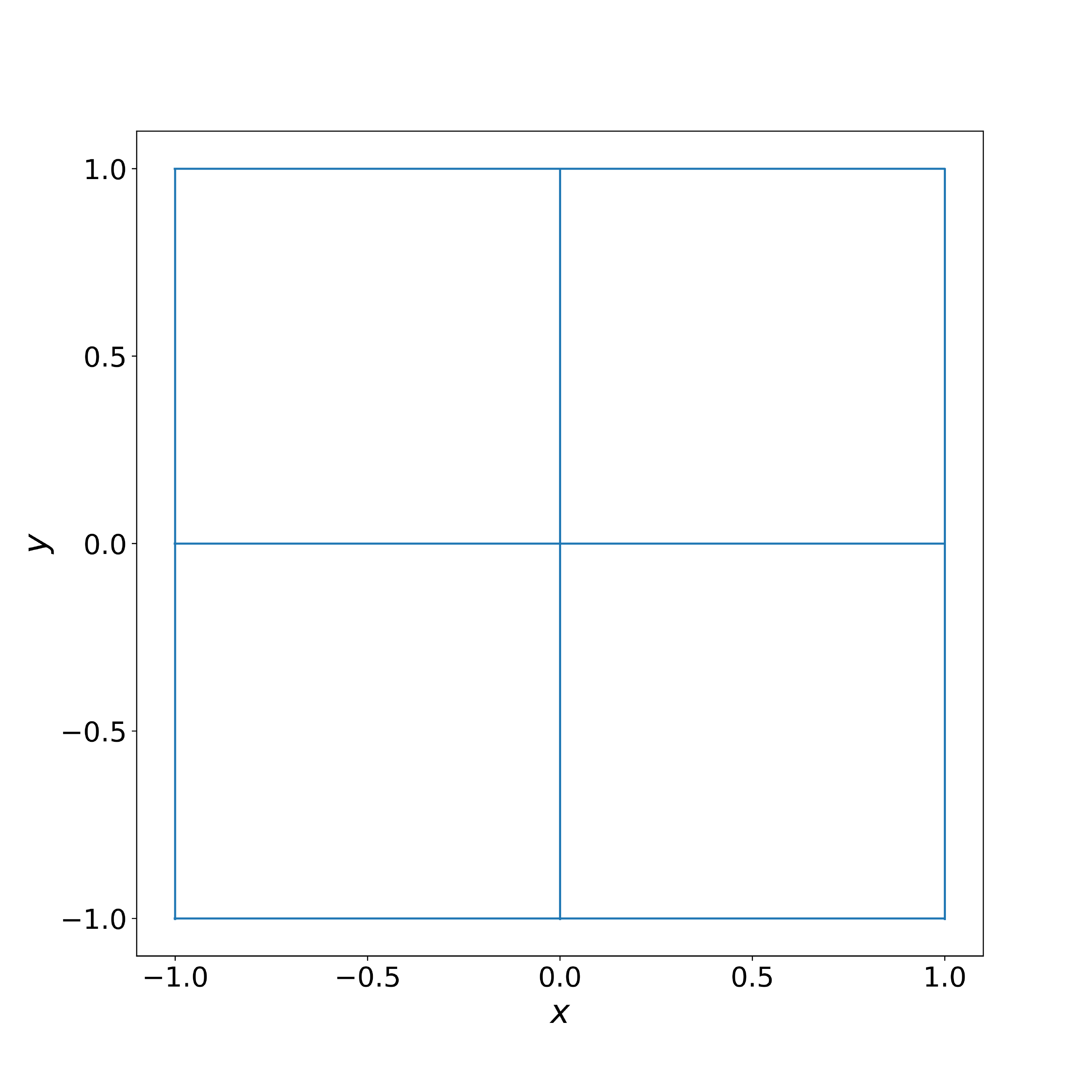}
		&
		\includegraphics[clip, trim=0cm 2cm 0cm 4.25cm, width=0.3\linewidth]{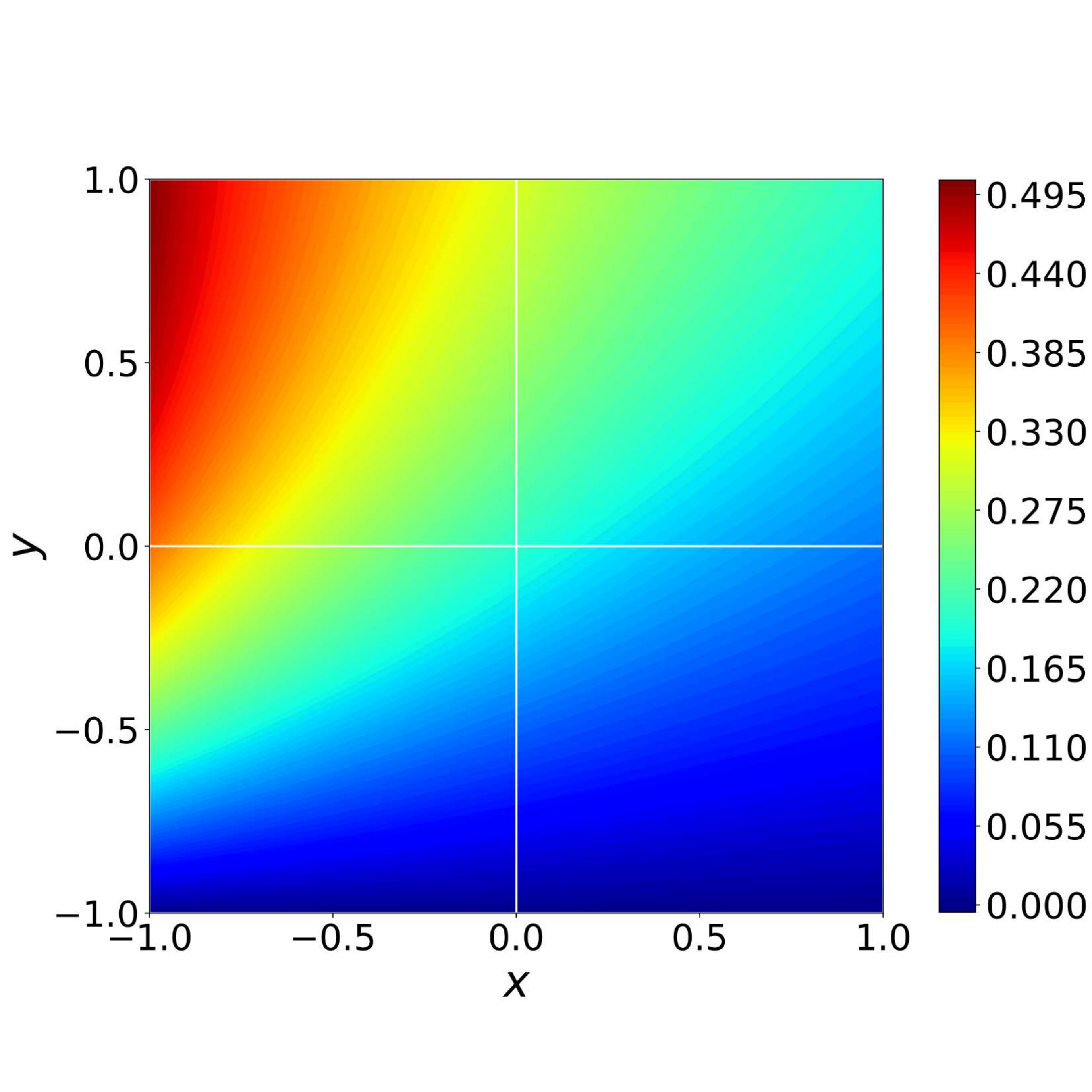}
		&
		\includegraphics[clip, trim=0cm 2cm 0cm 4.25cm, width=0.3\linewidth]{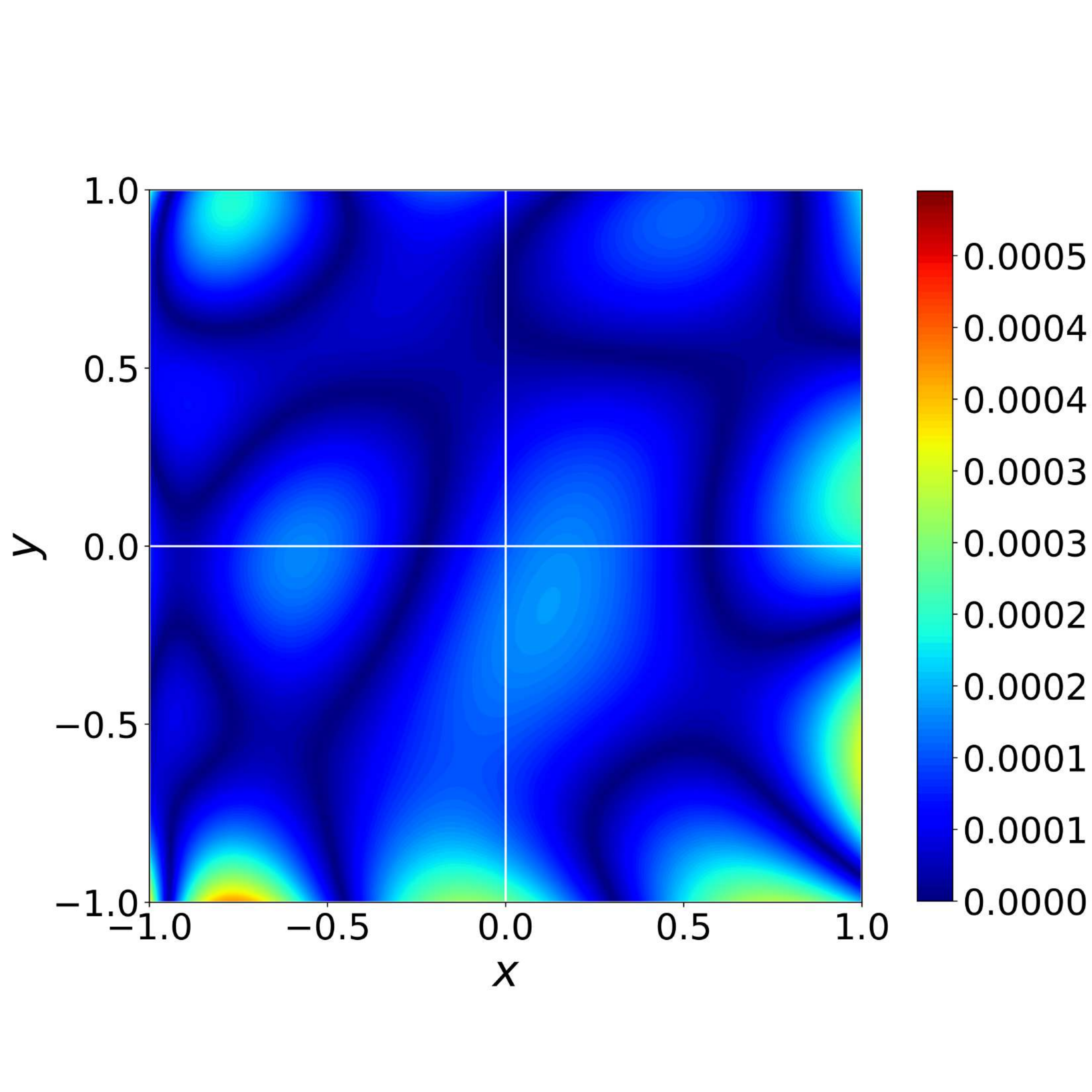}\\
	\end{tabular}
	\vspace{-0.15 in}
	\caption{\scriptsize \label{Fig: Poisson hom 2d vPINN} Two-dimensional homogeneous Poisson's equation. Top panel: (\textbf{A}) Exact solution \eqref{Eq: exact solution hom poisson 2-d}, (\textbf{B}) PINN prediction, and (\textbf{C}) PINN point-wise error. Bottom panel: (\textbf{D} column) \textit{h}-refinement via domain decomposition $N_{el_x} = N_{el_y} =1 $ and $2$, (\textbf{E} column) \textit{hp}-VPINN prediction, and (\textbf{F} column) \textit{hp}-VPINN point-wise error. In all cases, the network is fully connected with $\ell = 3$, $\mathcal{N} = 5$, and tanh activation function. The PINN parameters are $\lbrace N_r = 100, N_b = 80 \rbrace$ random residual and boundary points and $\tau_ b =10$. The \textit{hp}-VPINN parameters are $\lbrace K_1 = K_2 = 5 , Q = 10 \times 10 \rbrace$ in each sub-domain (element), $N_b = 80$ boundary points, and $\tau_ b =10 $. We use Adam optimizer with learning rate $10^{-3}$.
	}
\end{figure}

\vspace{0.2 in}
\begin{exm}
	We solve the homogeneous two-dimensional Poisson's equation, i.e. \eqref{Eq: 2-d Poisson} with $f(x,y) = 0$, over the bi-unit square domain $\Omega = [-1, 1] \times [-1, 1]$. The exact solution is given as 
	\begin{align}
	\label{Eq: exact solution hom poisson 2-d}
	u^{exact}(x, y) =  \frac{2(1+y)}{(3+x)^2 + (1+y)^2}.
	\end{align}
	The results are shown in Fig. \ref{Fig: Poisson hom 2d vPINN}.
\end{exm}

%

In this case, the exact solution is smooth   and thus an accurate approximation can be obtained by using a relatively small network with $\ell = 3$, $\mathcal{N} = 5$, and tanh activation function. We compare the point-wise approximation error of PINN and \textit{hp}-VPINN (with single domain and multiple sub-domains). The PINN formulation uses $N_r = 100$ residual and $N_b = 80$ boundary points randomly drawn from a uniform distribution. In \textit{hp}-VPINN formulation, we use 5 test functions in each direction $x$ and $y$, and employ $10 \times 10$ quadrature points. In this case, the domain decomposition does not improve the approximation and the point-wise error is of order $O(10^{-4})$ in all formulations. However, the domain decomposition can be later used in parallel computation, where each sub-domain can be individually solved in separate computer node, and thus, further improve the total computational costs. We note that the $\prescript{(1)}{}{\mathcal{R}} $ and $\prescript{(2)}{}{\mathcal{R}} $ formulations produce similar error level and we only show the results for latter one. The $\prescript{(3)}{}{\mathcal{R}} $ is not considered here as the boundary terms can cause further complication in the loss function.

\begin{figure}[!ht]
	\center
	\begin{tabular}{c c c}
		\multicolumn{3}{c}{{\scriptsize PINN}} \\  [-3 pt] 
		\hline \\ [-8 pt]
		
        \multicolumn{1}{l}{\, \textbf{A} \quad\quad\quad {\scriptsize exact solution}}
        &\multicolumn{1}{l}{\, \textbf{B} \quad\quad\quad {\scriptsize predicted solution}}
        &\multicolumn{1}{l}{\, \textbf{C} \quad\quad\quad {\scriptsize point-wise error}}\\ [-3 pt]	 
		
		\includegraphics[clip, trim=0cm 2cm 0cm 3.5cm, width=0.3\linewidth]{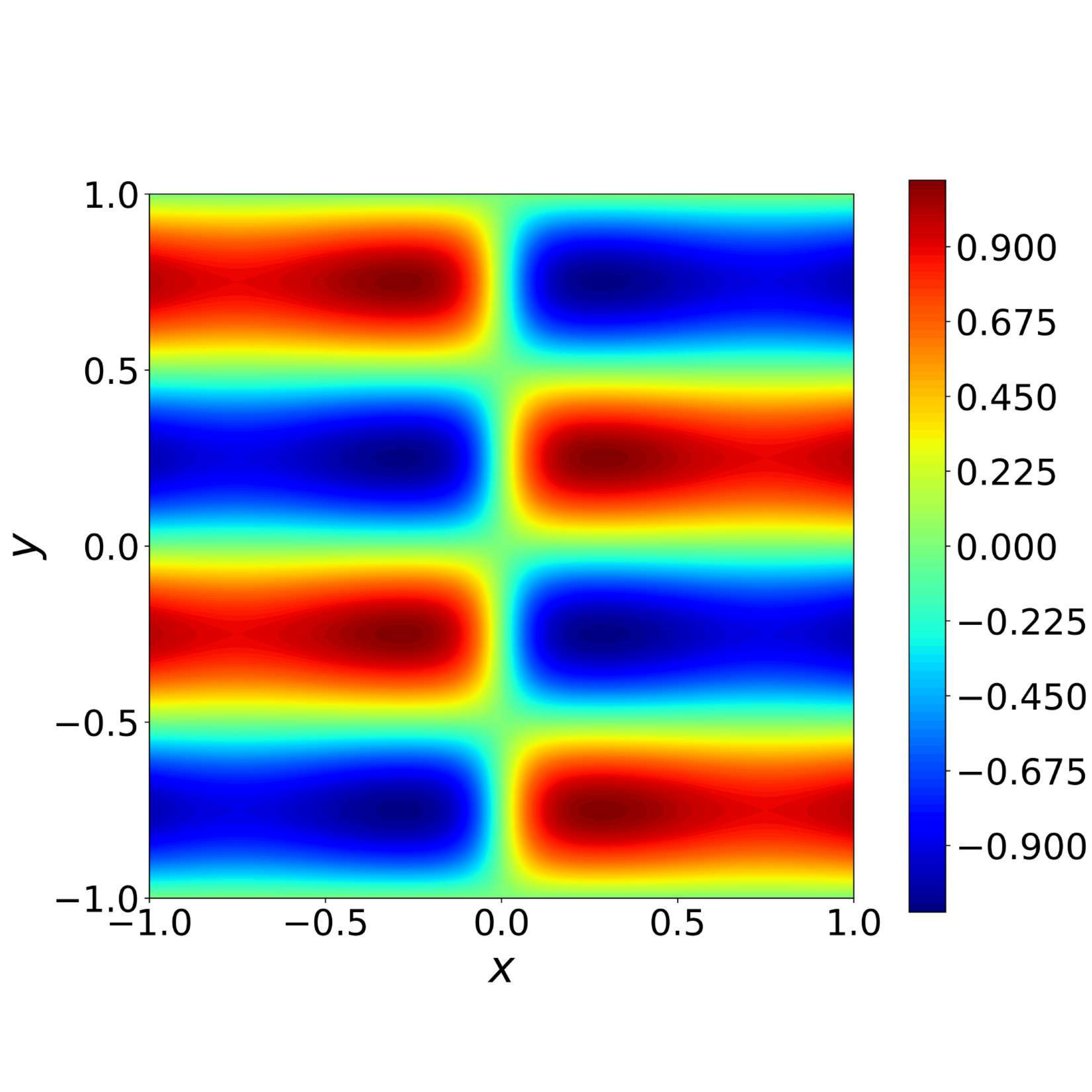}
		&
		\includegraphics[clip, trim=0cm 2cm 0cm 3.5cm, width=0.3\linewidth]{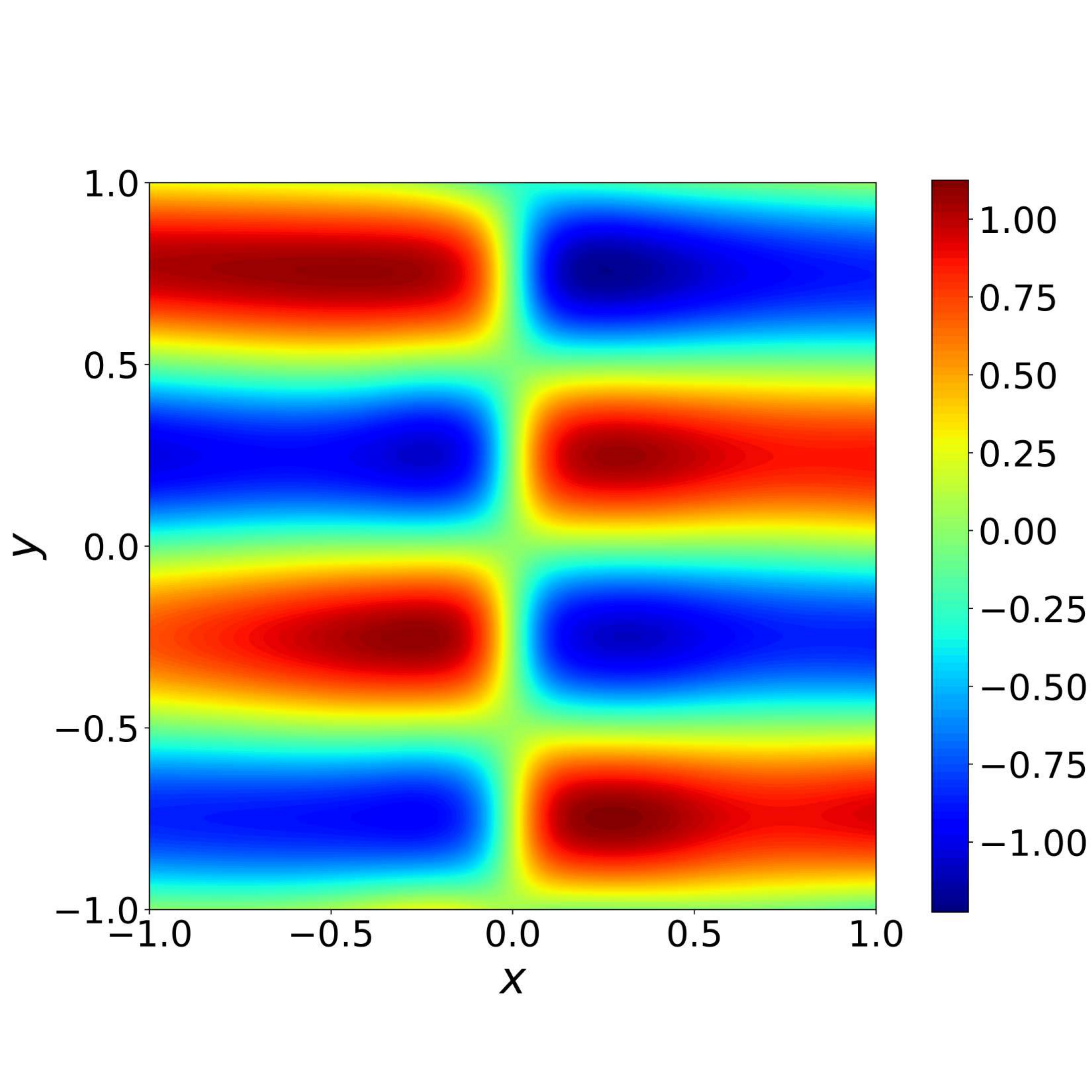}
		&
		\includegraphics[clip, trim=0cm 2cm 0cm 3.5cm, width=0.3\linewidth]{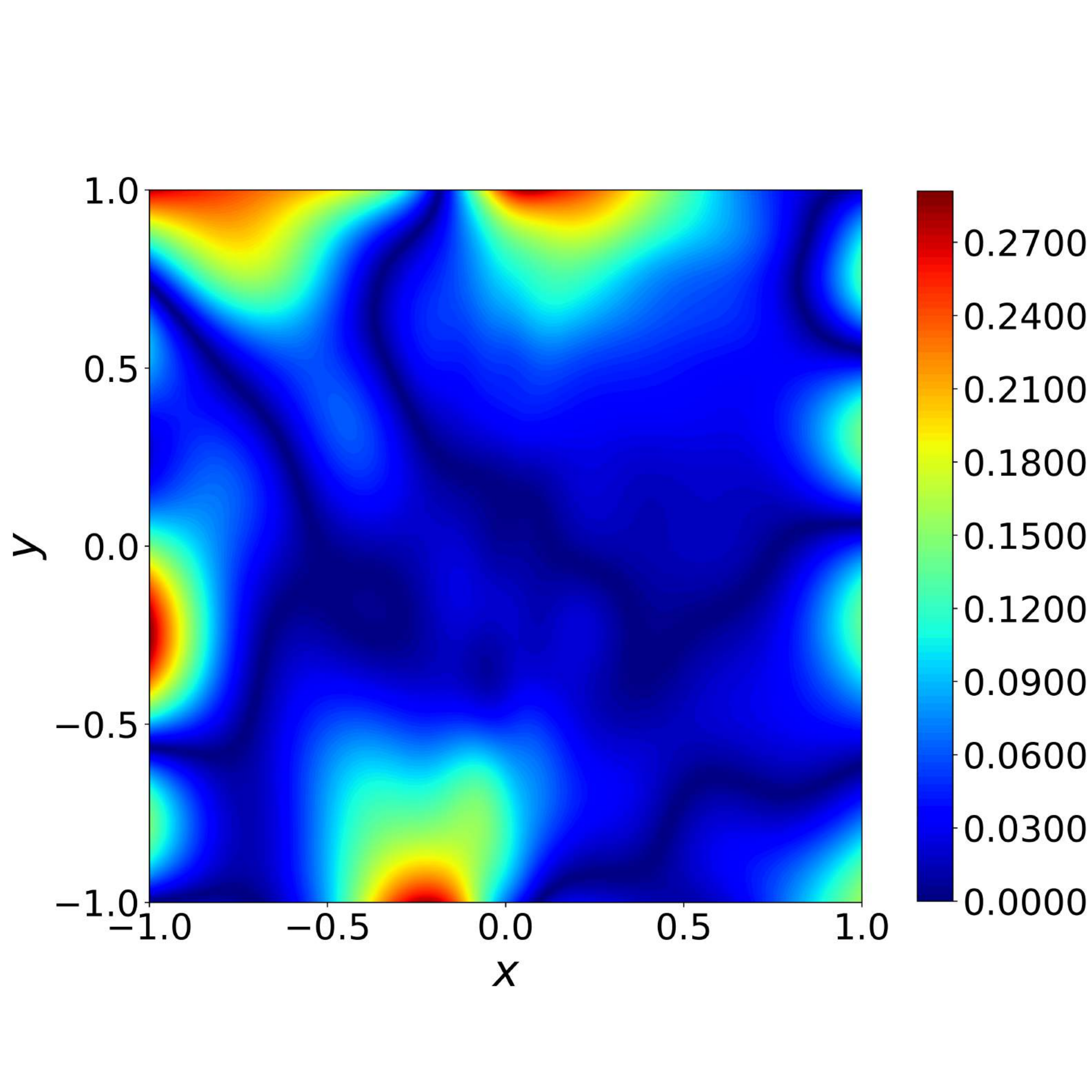}
		\\
		
		\multicolumn{3}{c}{{\scriptsize \quad \textit{hp}-VPINN: $\prescript{(1)}{}{\mathcal{R}} $ formulation}} \\  [-1 pt] 
		\hline \\ [-8 pt]
		
        \multicolumn{1}{l}{\,\,\, \textbf{D} \quad\quad {\scriptsize domain decomposition}}
        &\multicolumn{1}{l}{\, \textbf{E} \quad\quad\quad {\scriptsize predicted solution}}
        &\multicolumn{1}{l}{\, \textbf{F} \quad\quad\quad {\scriptsize point-wise error}}\\ [-1 pt] 
		
		\includegraphics[clip, trim=0cm 1cm 0cm 3cm, width=0.27\linewidth]{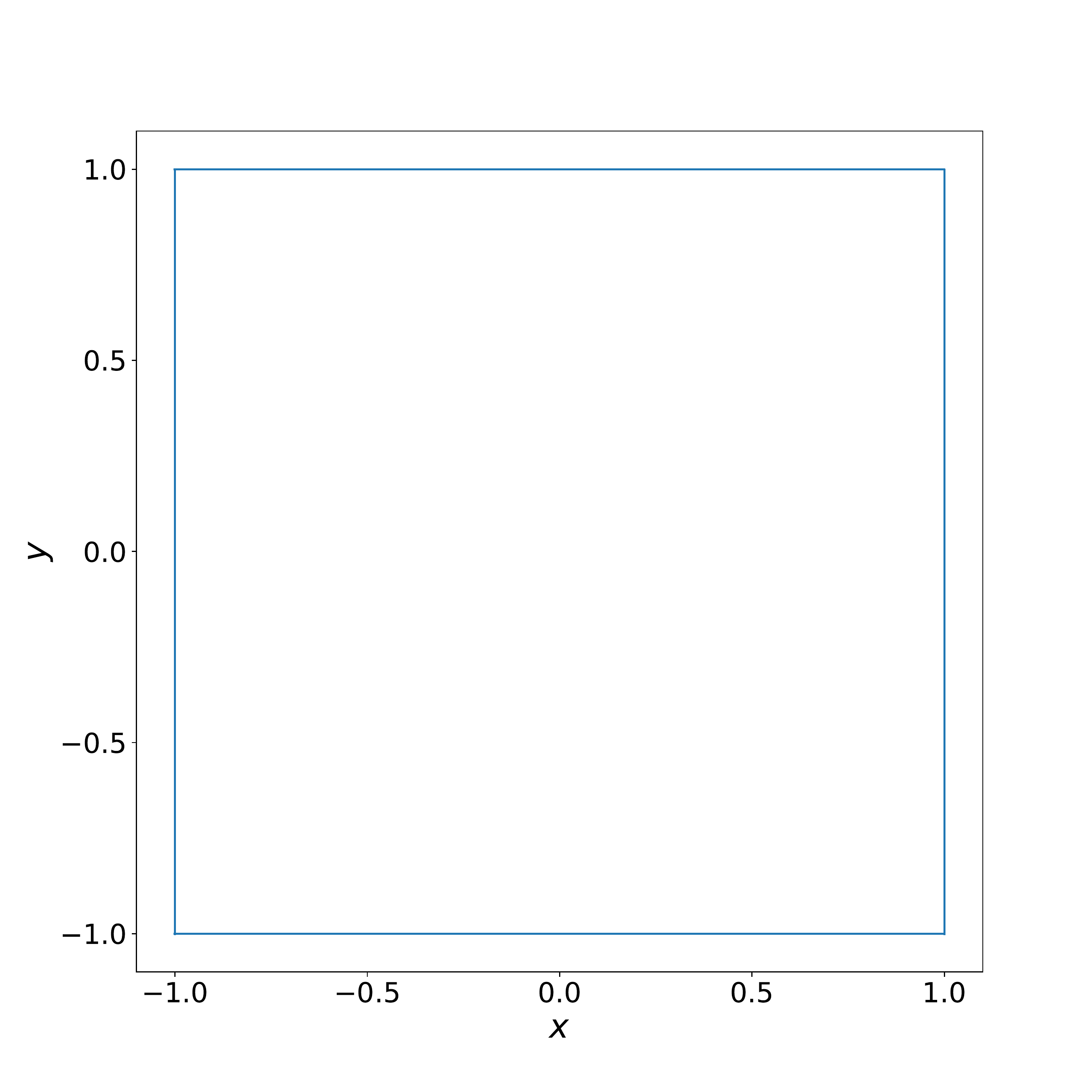}
		&
		\includegraphics[clip, trim=0cm 2cm 0cm 4.25cm, width=0.3\linewidth]{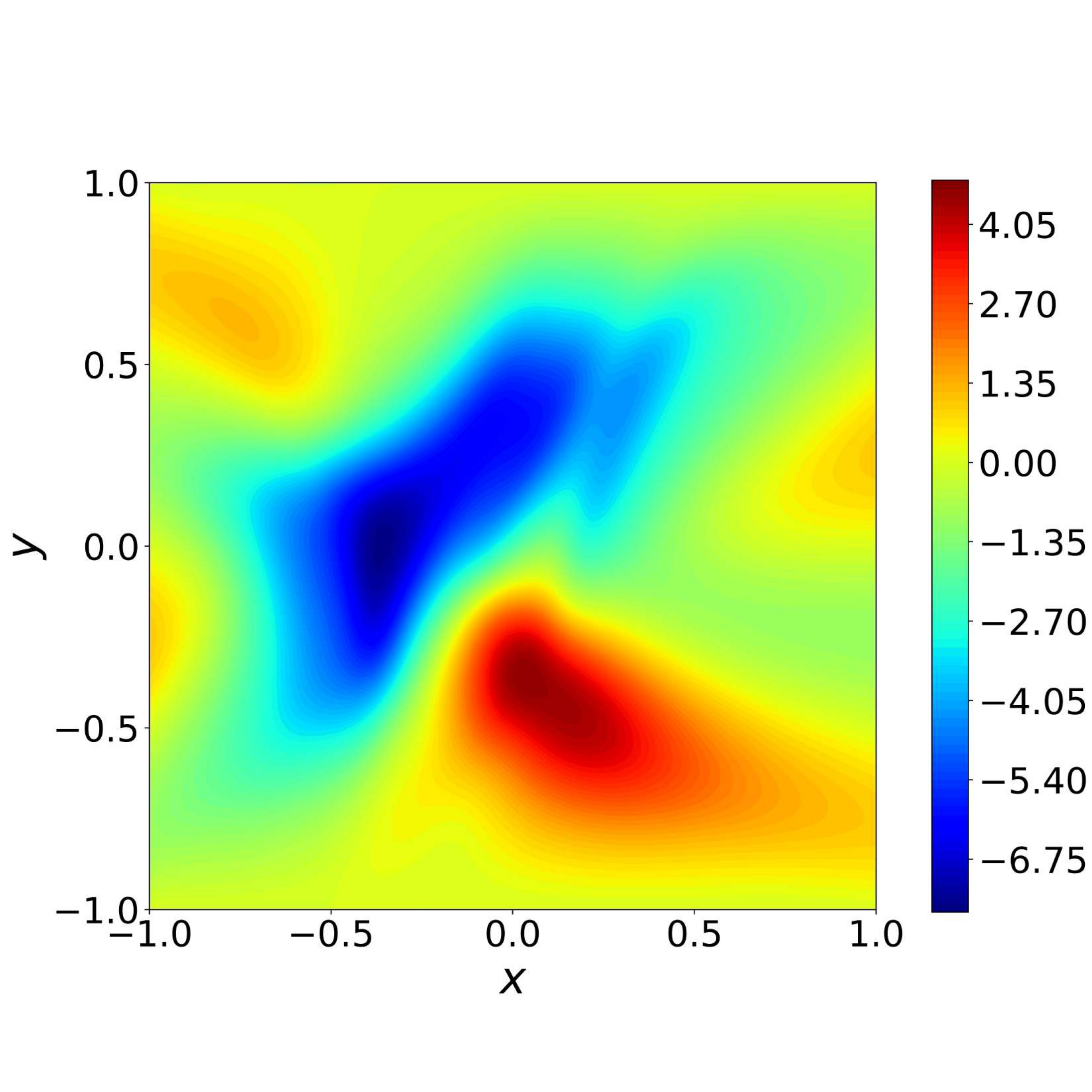}
		&
		\includegraphics[clip, trim=0cm 2cm 0cm 4.25cm, width=0.3\linewidth]{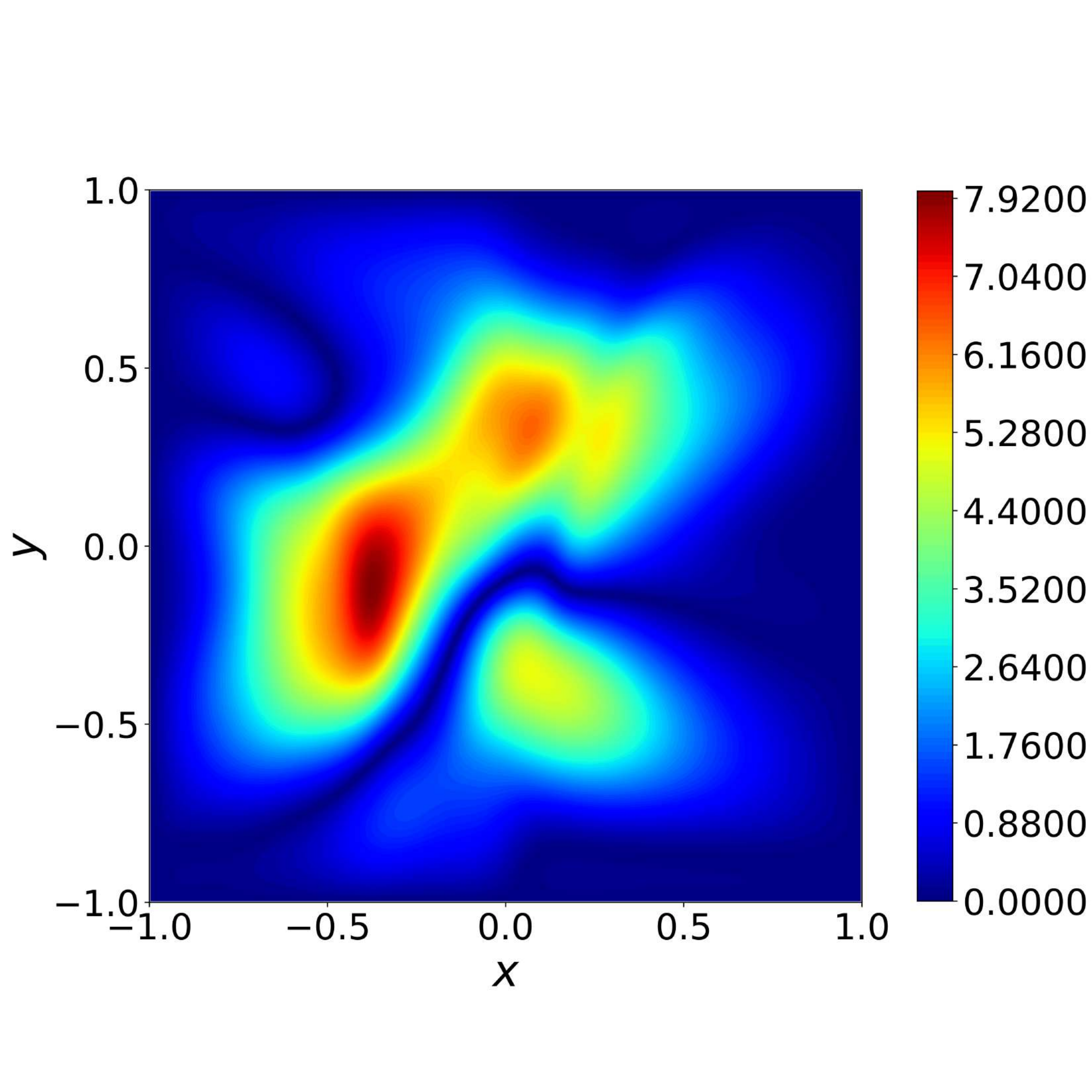}\\
		
		\includegraphics[clip, trim=0cm 1cm 0cm 3cm, width=0.27\linewidth]{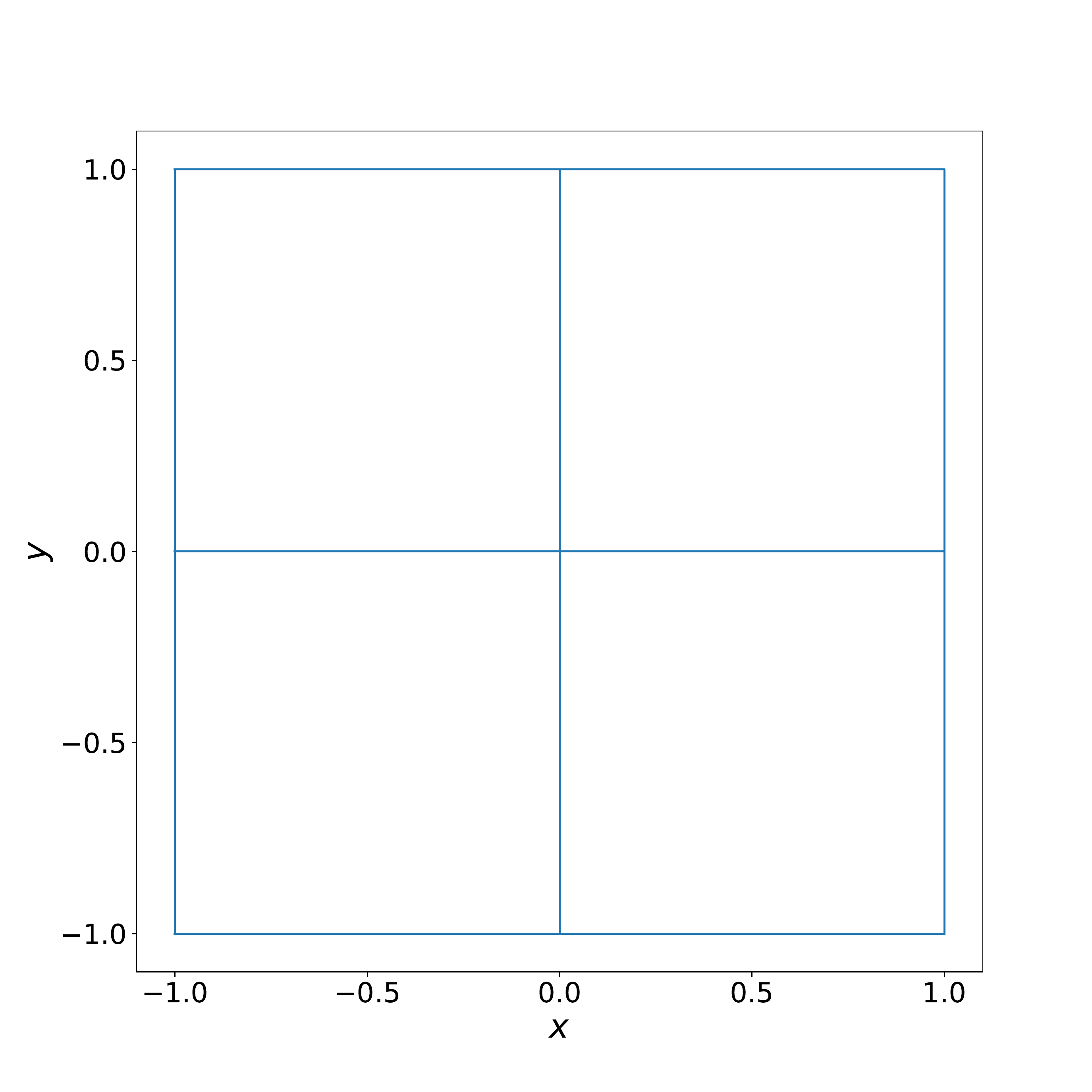}
		&
		\includegraphics[clip, trim=0cm 2cm 0cm 4.25cm, width=0.3\linewidth]{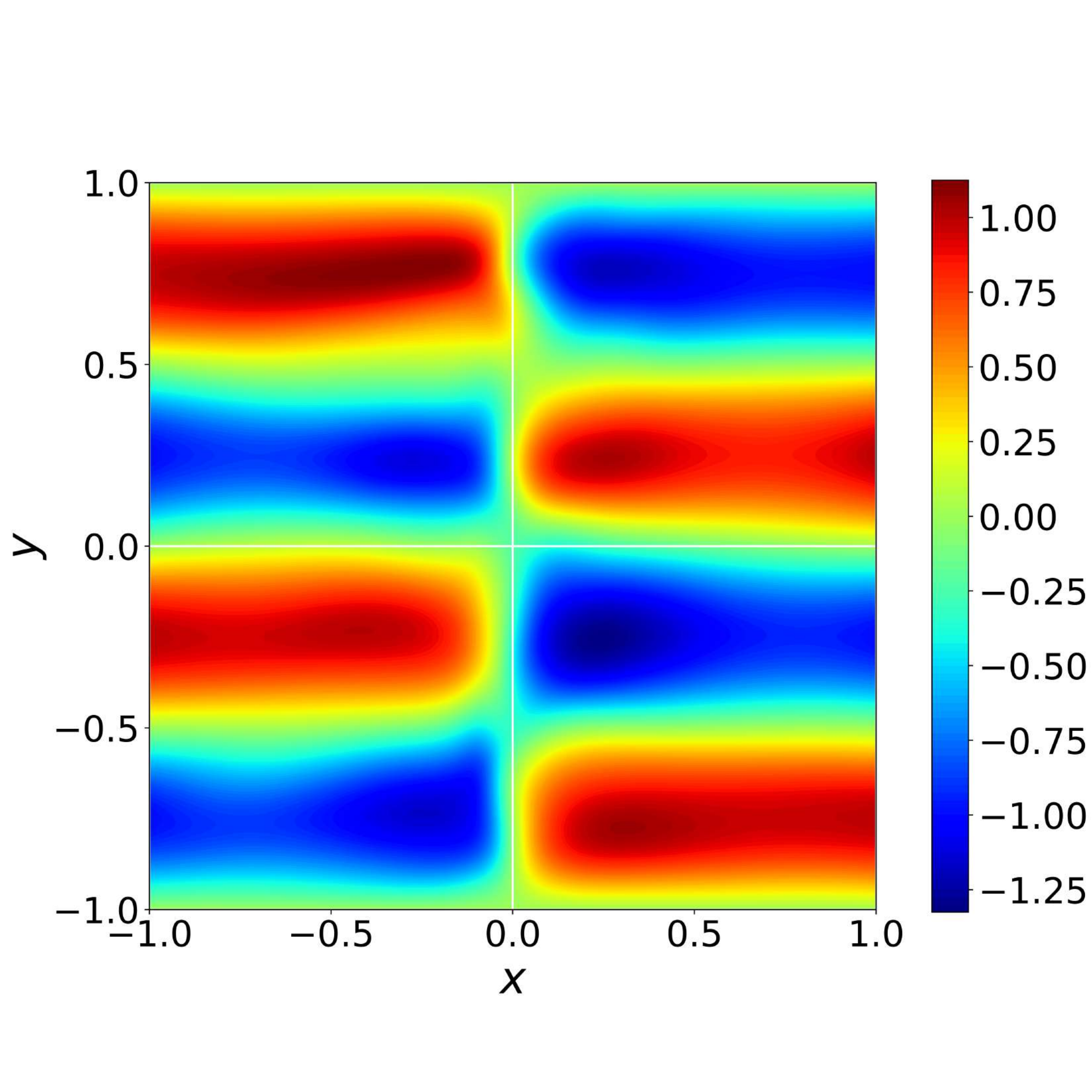}
		&
		\includegraphics[clip, trim=0cm 2cm 0cm 4.25cm, width=0.3\linewidth]{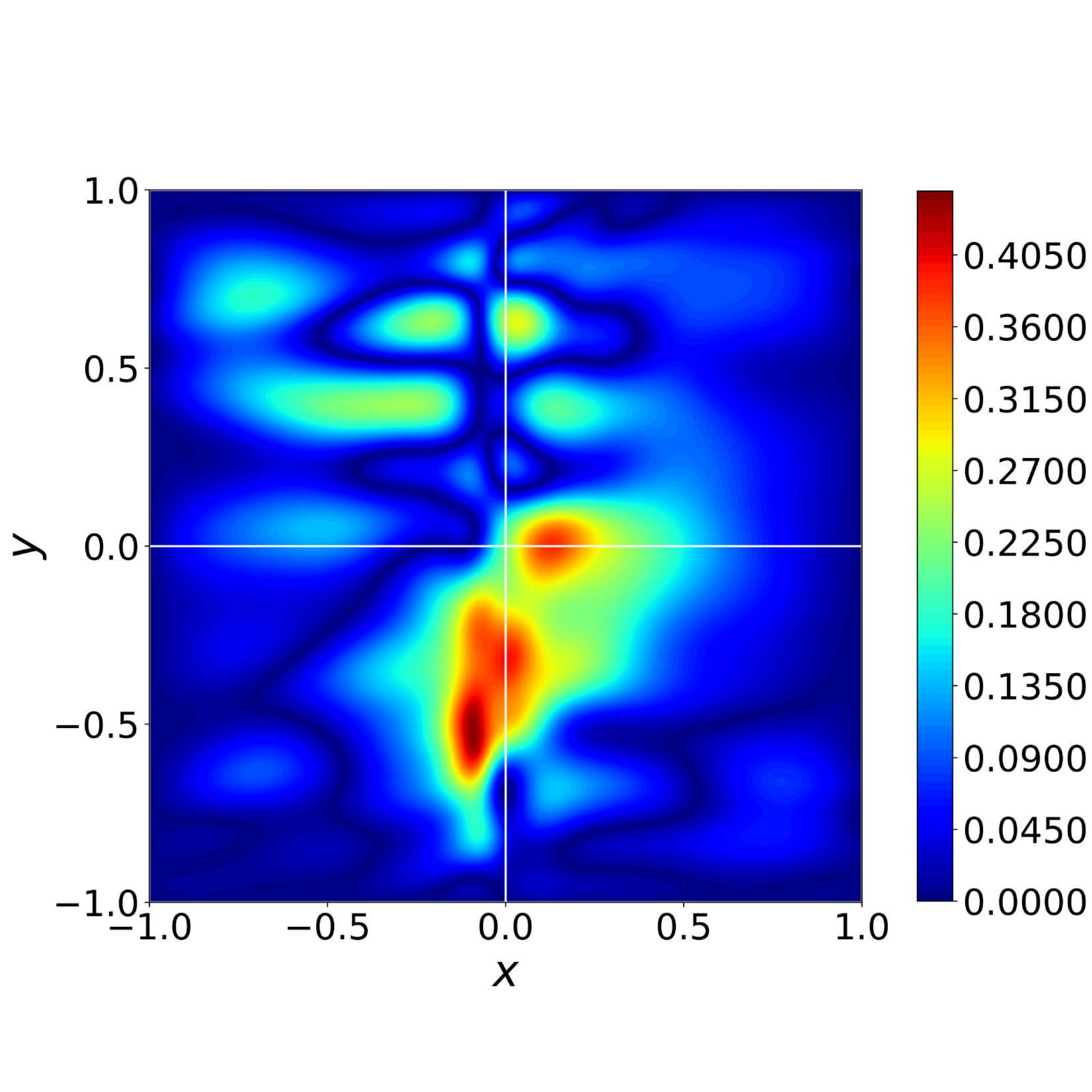}\\
		
		\includegraphics[clip, trim=0cm 1cm 0cm 3cm, width=0.27\linewidth]{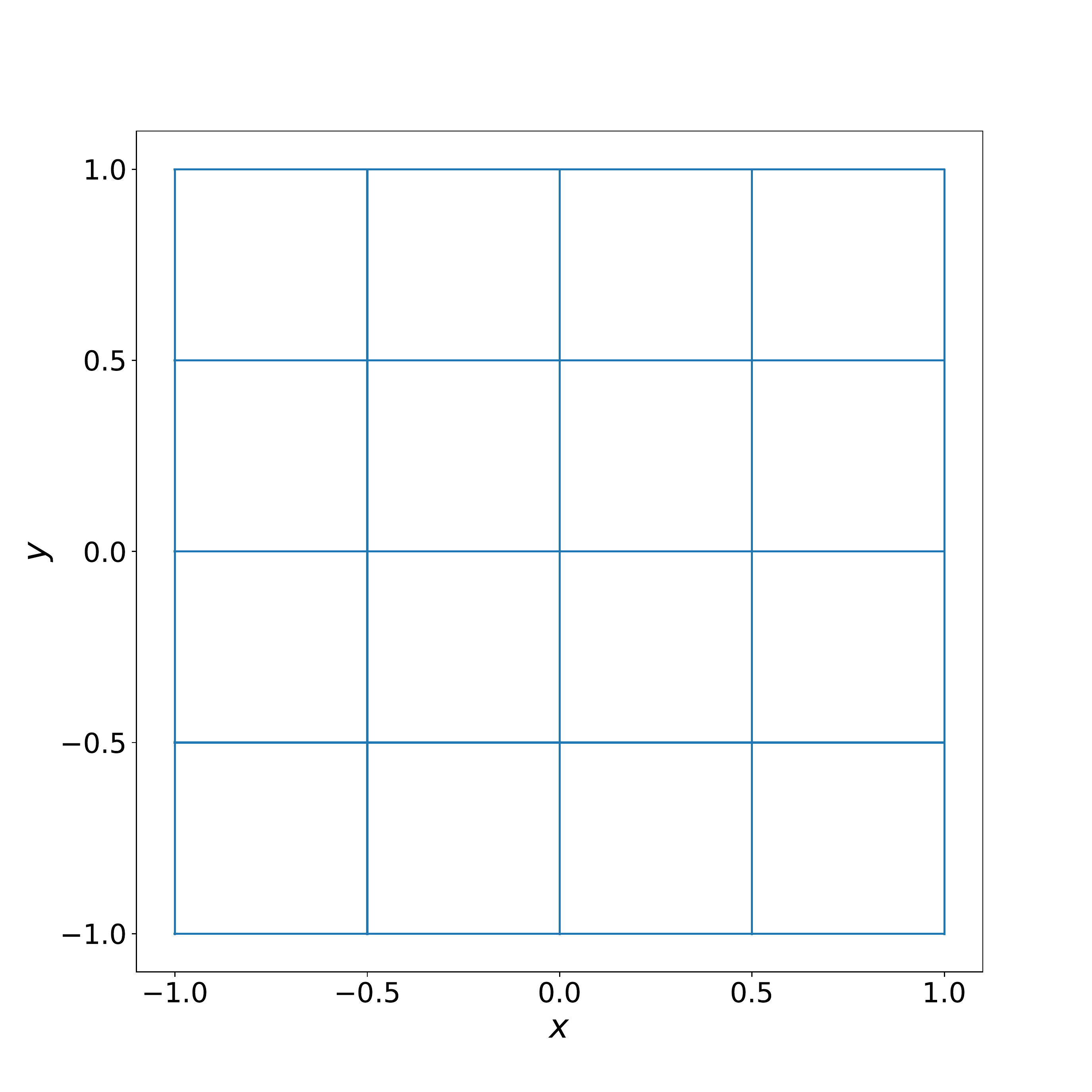}
		&
		\includegraphics[clip, trim=0cm 2cm 0cm 4.25cm, width=0.3\linewidth]{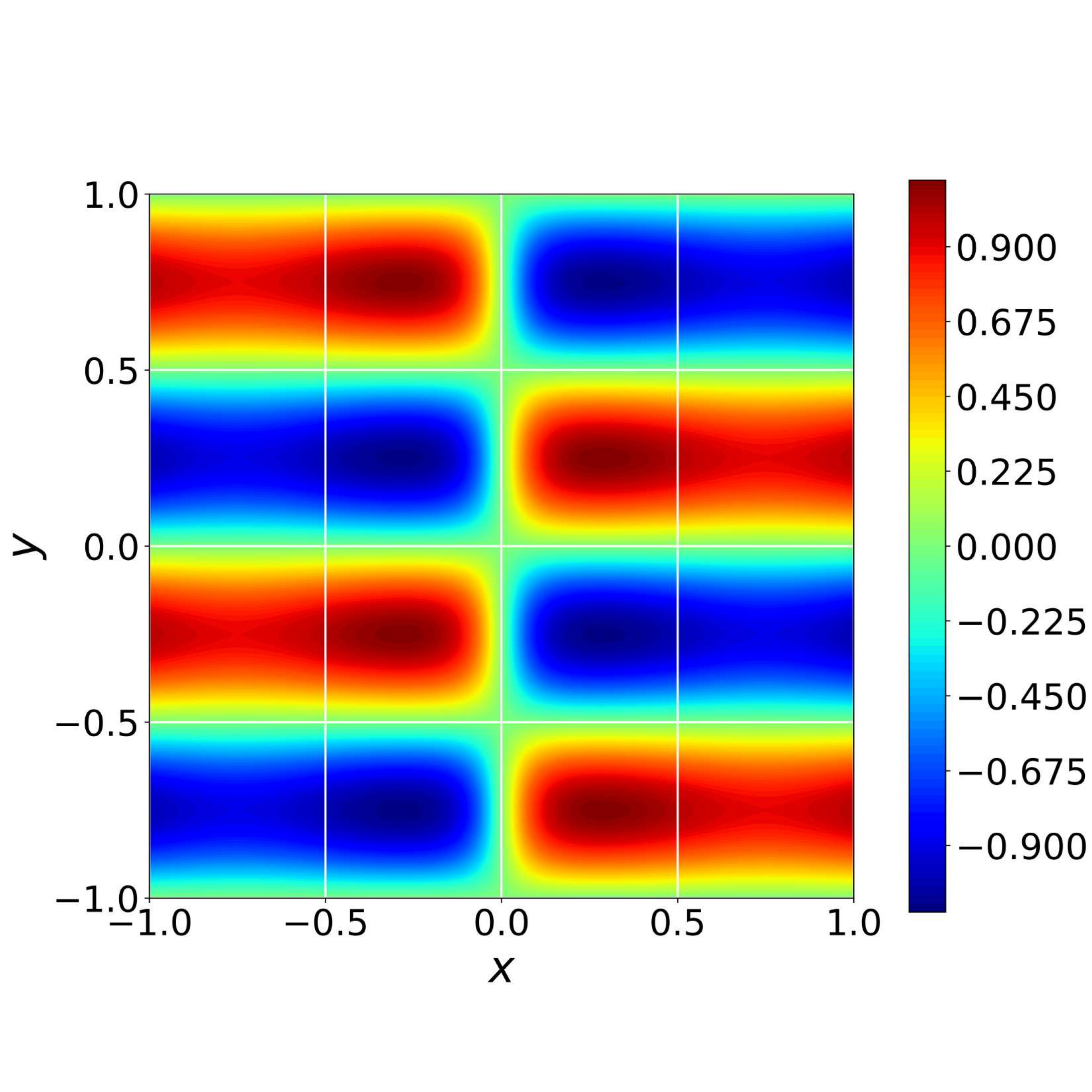}
		&
		\includegraphics[clip, trim=0cm 2cm 0cm 4.25cm, width=0.3\linewidth]{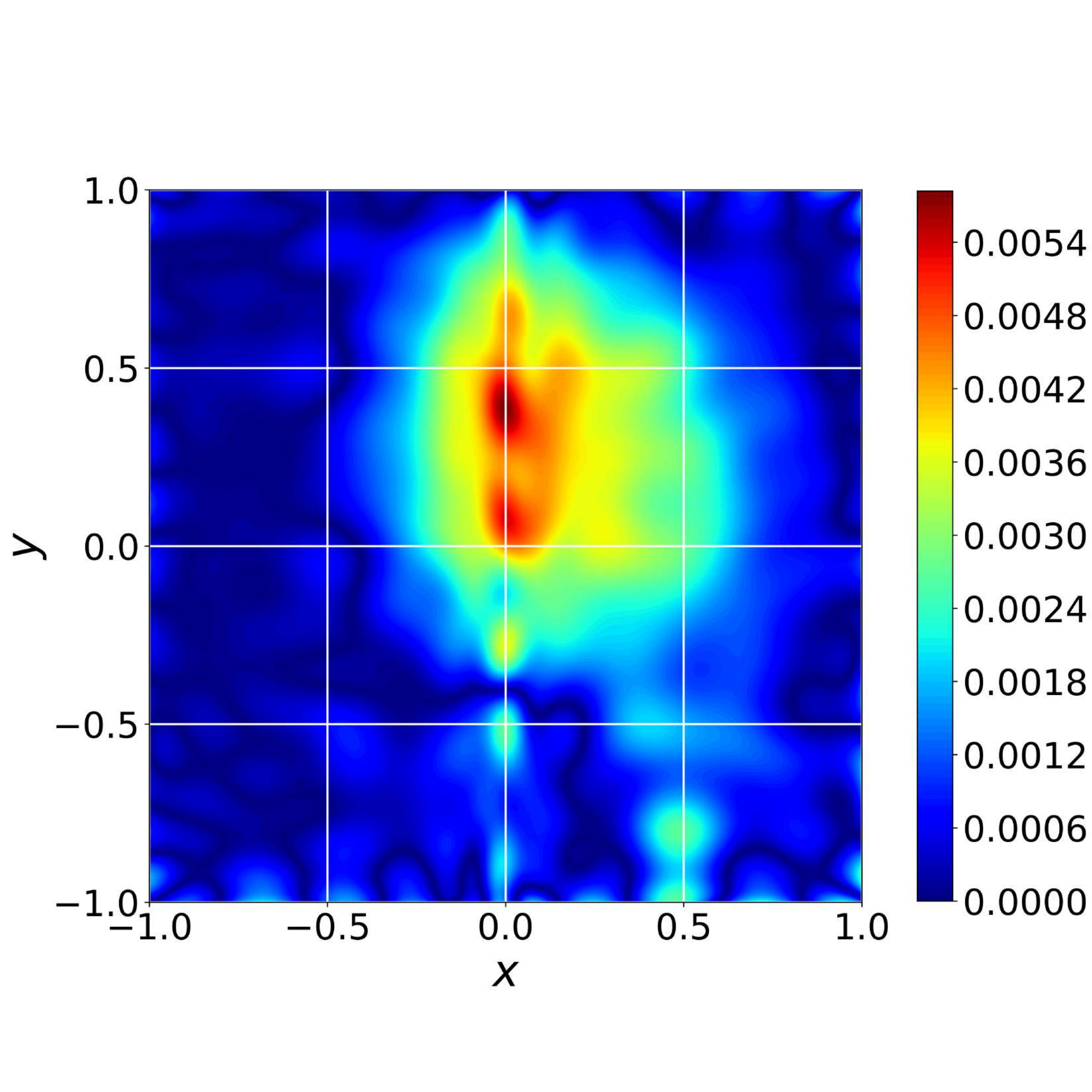}\\
		
		\includegraphics[clip, trim=0cm 1cm 0cm 3cm, width=0.27\linewidth]{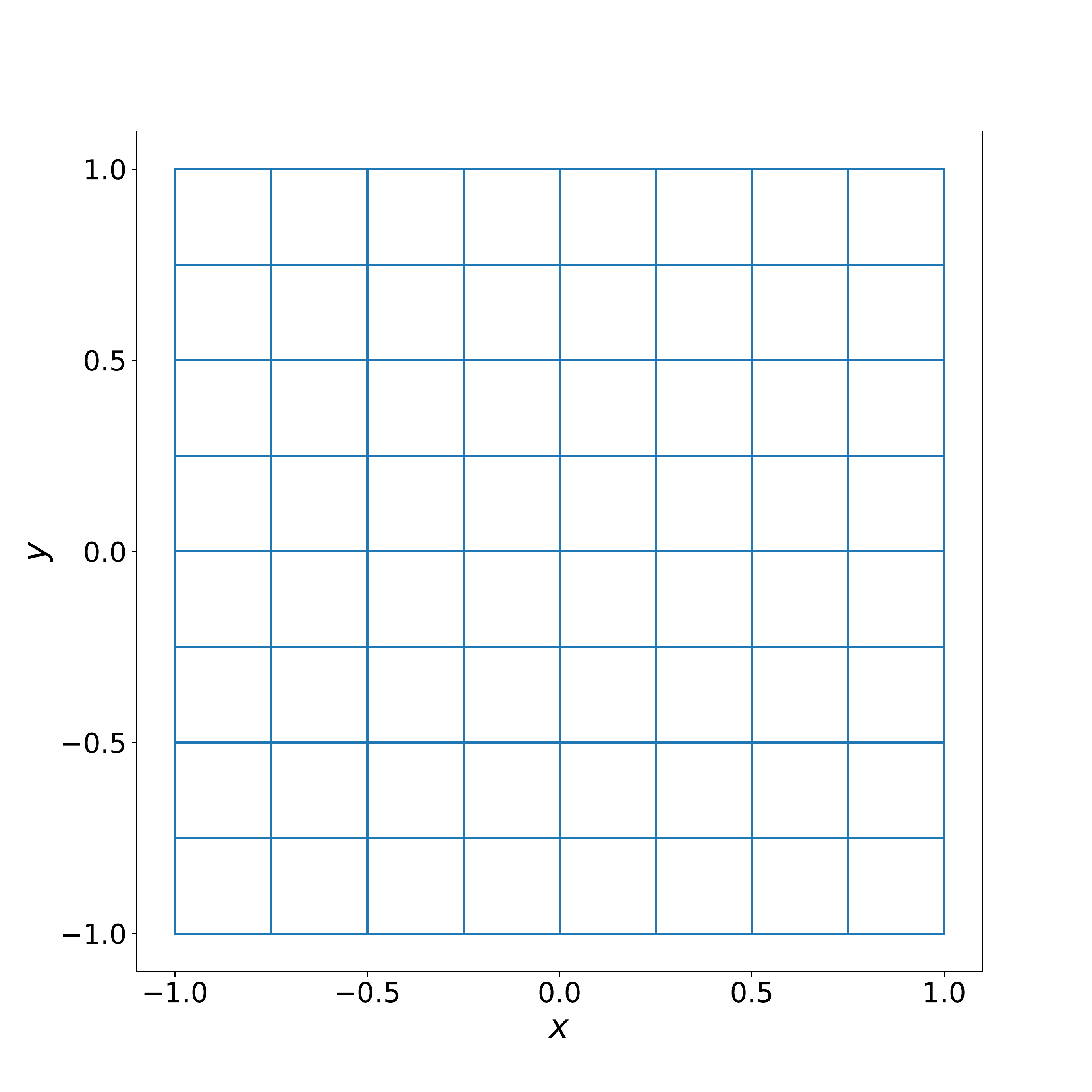}
		&
		\includegraphics[clip, trim=0cm 2cm 0cm 4.25cm, width=0.3\linewidth]{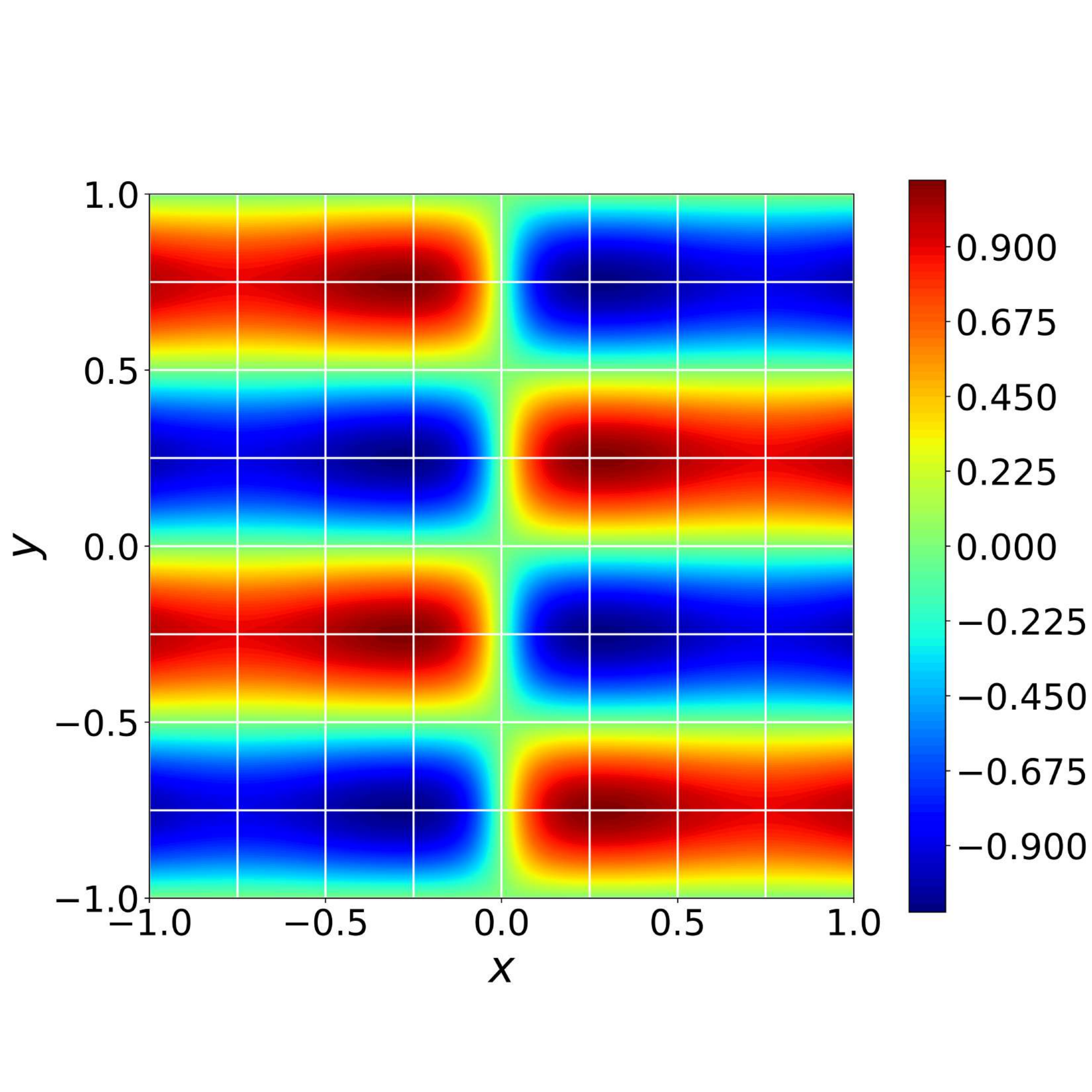}
		&
		\includegraphics[clip, trim=0cm 2cm 0cm 4.25cm, width=0.3\linewidth]{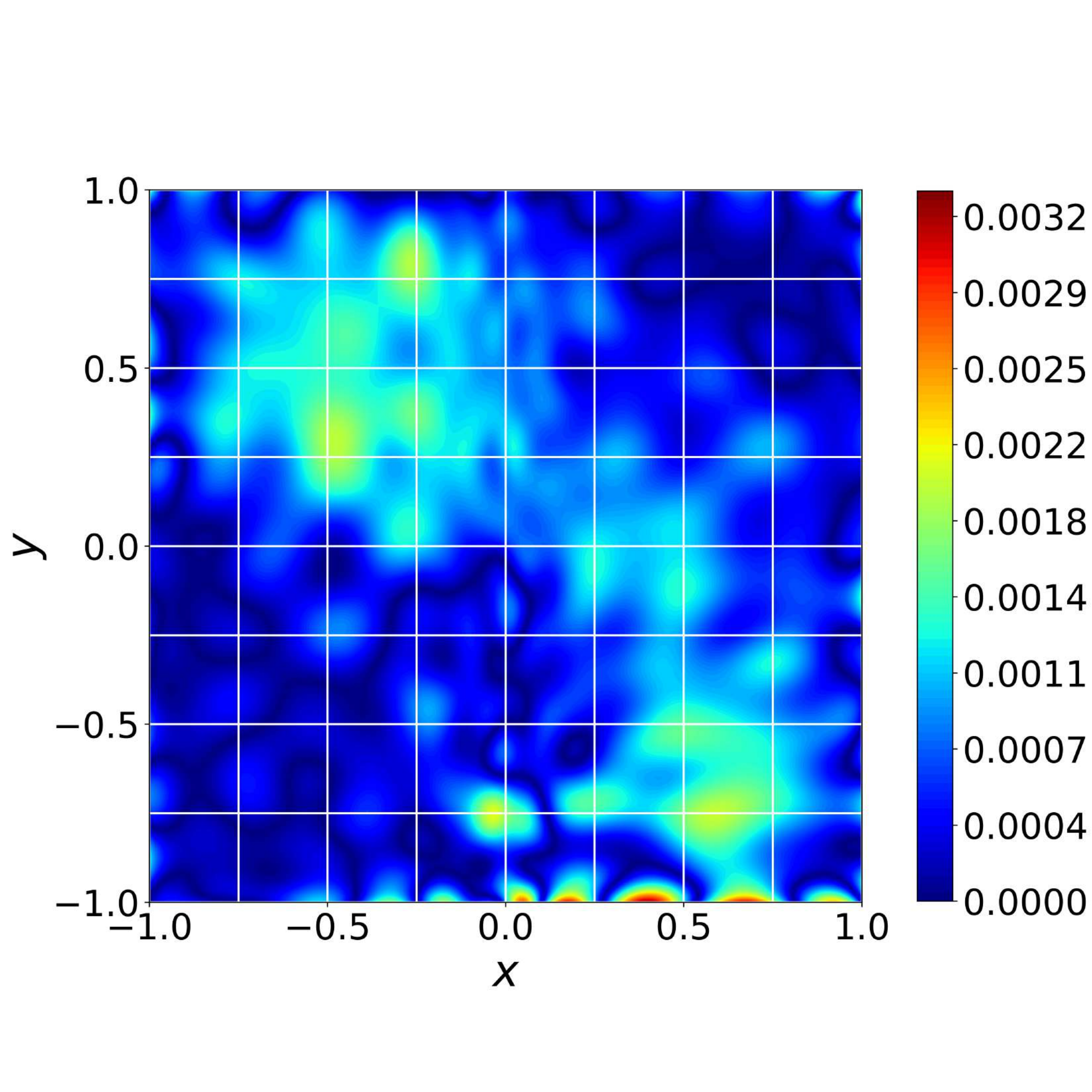}\\

	\end{tabular}
    \vspace{-0.15 in}
	\caption{\scriptsize \label{Fig: Poisson 2d steep vPINN} Two-dimensional Poisson's equation with steep exact solution. Top panel: (\textbf{A}) Exact solution \eqref{Eq: exact solution 2-d}, (\textbf{B}) PINN prediction, and (\textbf{C}) PINN point-wise error. Bottom panel: (\textbf{D} column) \textit{h}-refinement via domain decomposition $N_{el_x} = N_{el_y} =1 $, $2$, $4$, and $8$, (\textbf{E} column) \textit{hp}-VPINN prediction, and (\textbf{F} column) \textit{hp}-VPINN point-wise error. In all cases, the network is fully connected with $\ell = 3$, $\mathcal{N} = 5$, and tanh activation function. The PINN parameters are $\lbrace N_r = 1000, N_b = 80 \rbrace$ random residual and boundary points and $\tau_ b =10$. The \textit{hp}-VPINN parameters are $\lbrace K_1 = K_2 = 5 , Q = 10 \times 10 \rbrace$ in each sub-domain (element), $N_b = 80$ boundary points, and $\tau_ b =10 $. We use Adam optimizer with learning rate $10^{-3}$.}
\end{figure}

\vspace{0.2 in}
\begin{exm}
	We solve the two-dimensional Poisson's equation with the following exact solution with a steep change along x direction and a sinusoidal behavior in y direction as
	\begin{align}
	\label{Eq: exact solution 2-d}
	u^{exact}(x, y) =  \left( 0.1 \sin(2 \pi x) + \tanh(10 x) \right) \times \sin(2 \pi y),
	\end{align}
	where the force function is obtained by substituting the exact solution in \eqref{Eq: 2-d Poisson}. The results are shown in Fig. \ref{Fig: Poisson 2d steep vPINN}.
	
\end{exm}

%

In this case, we use a wider network with $\ell = 3$, $\mathcal{N} = 20$, and tanh activation function to accurately capture the steep change at $x = 0$. We study the error convergence in the \textit{hp}-VPINN formulation by successively increasing the number of sub-domains in the domain decomposition. In each sub-domain, we use 5 test functions in each direction $x$ and $y$, and employ $10 \times 10$ quadrature points. Figure \ref{Fig: Poisson 2d steep vPINN} shows the convergence of error as we increase the number of division along $x$ and $y$ axes. For $N_{el_x} = N_{el_y} = 8$, the error is of order $O(10^{-3})$.

\begin{figure}[!ht]
	\center
	\begin{tabular}{c c c}
		\multicolumn{3}{l}{\qquad\qquad\quad\quad \textbf{A} \quad\quad\quad {\scriptsize reference solution} \qquad\qquad\qquad  \textbf{B} \quad\quad {\scriptsize PINN point-wise error }} \\  [-2 pt]

		\multicolumn{3}{c}{
			\includegraphics[clip, trim=0cm 2cm 0cm 3.5cm, width=0.3\linewidth]{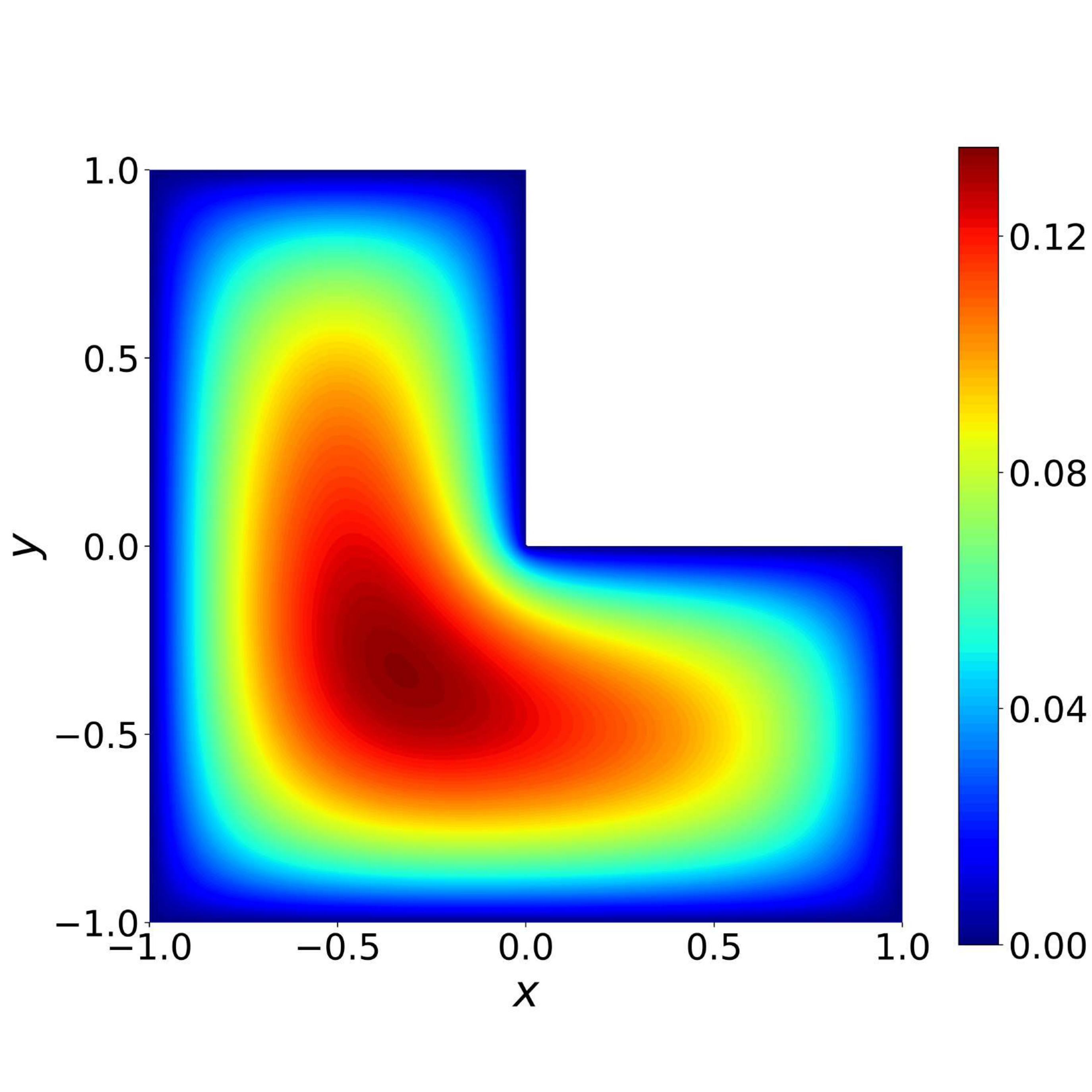}
			\quad
			\includegraphics[clip, trim=0cm 1cm 0cm 2.5cm, width=0.3\linewidth]{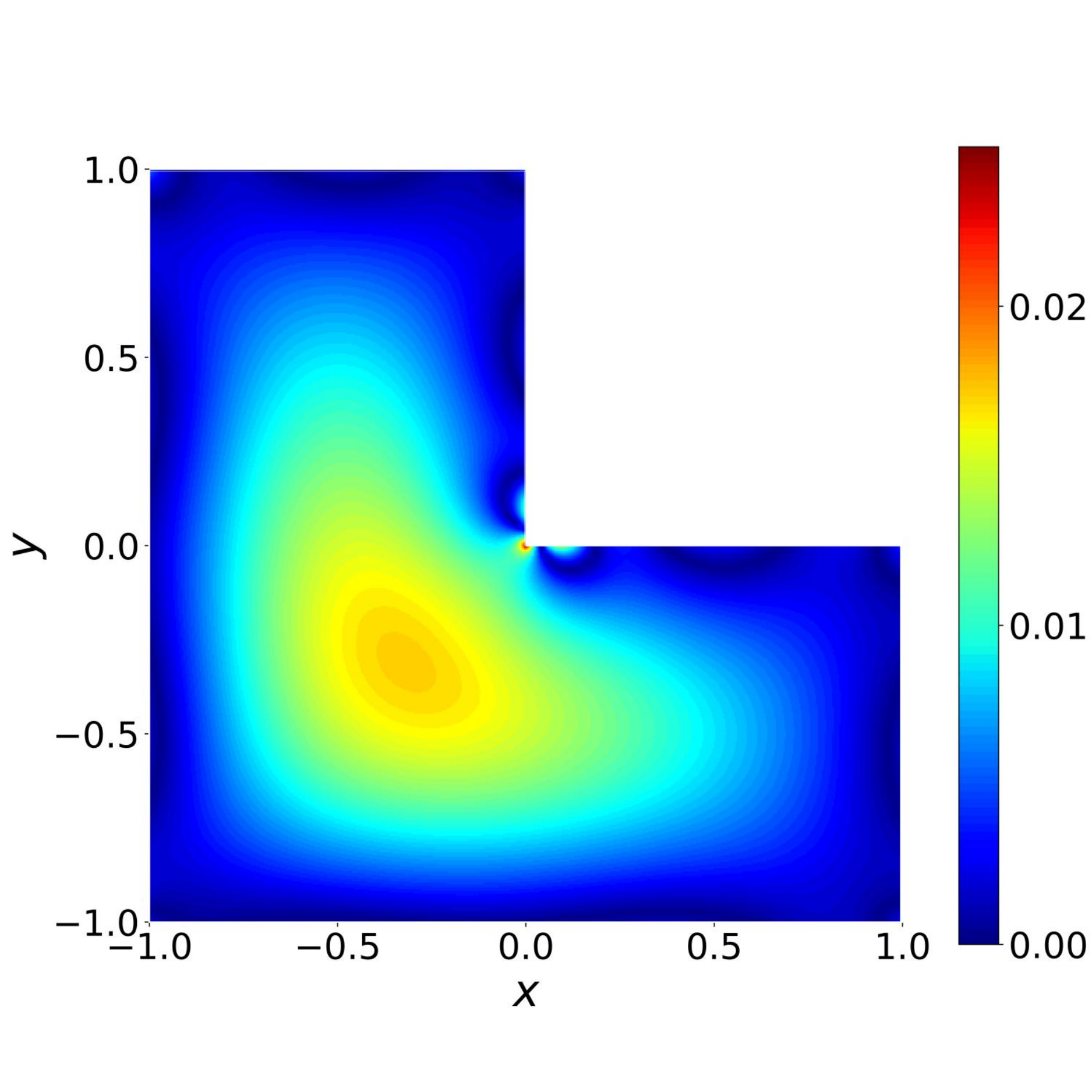}
		}
		\\
		\multicolumn{3}{c}{{\scriptsize \quad \textit{hp}-VPINN: $\prescript{(1)}{}{\mathcal{R}} $ formulation}} \\  [-1 pt] 
		\hline \\ [-8 pt]

        \multicolumn{1}{l}{\, \textbf{C} \quad\quad {\scriptsize domain decomposition}}
        &\multicolumn{1}{l}{\, \textbf{D} \quad\quad\quad {\scriptsize point-wise error}}
        &\multicolumn{1}{l}{\, \textbf{E} \qquad\quad\quad\quad {\scriptsize loss}}\\  [-2 pt]
		\includegraphics[clip, trim=0cm 0cm 0cm 0cm, width=0.24\linewidth]{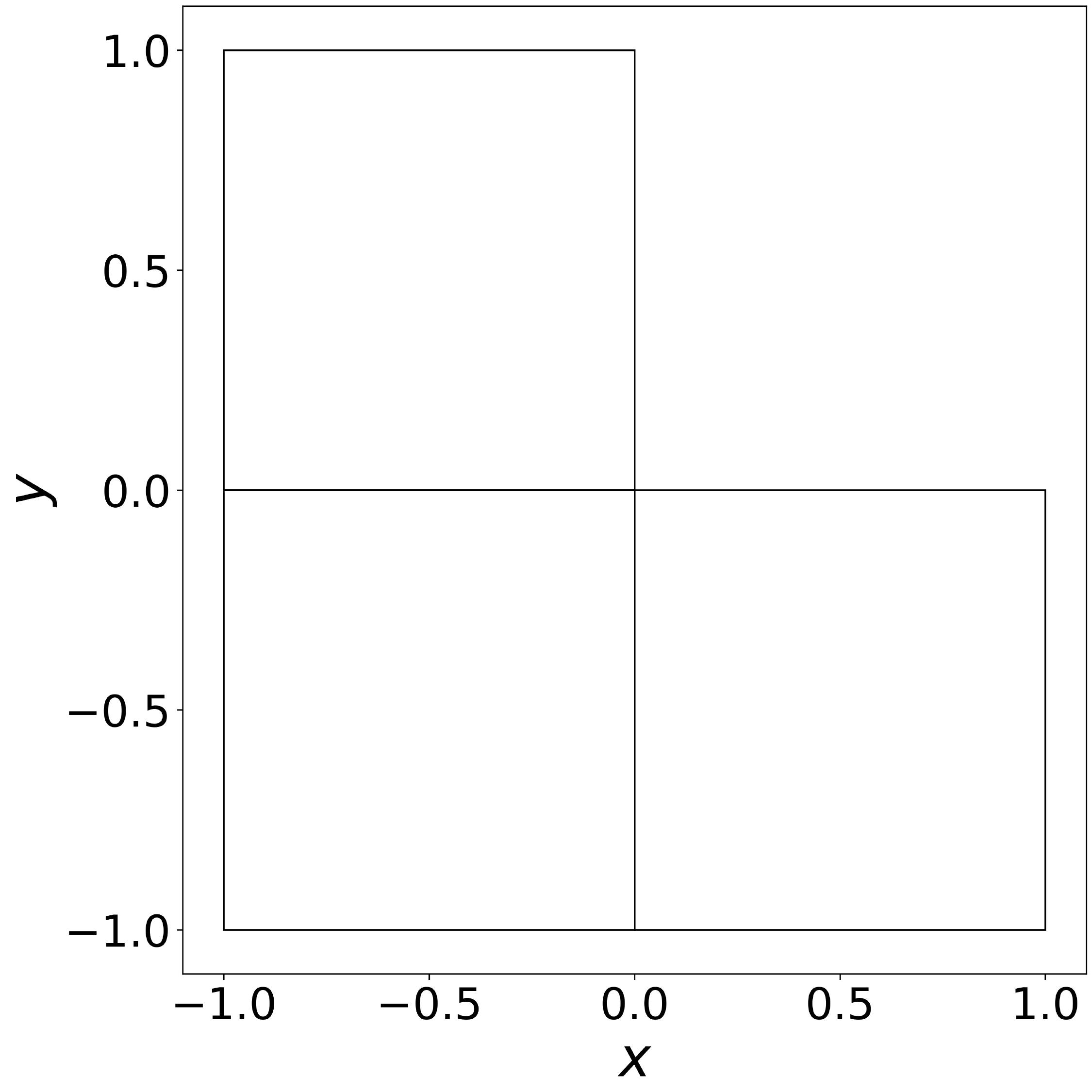}
		&
		\includegraphics[clip, trim=0cm 2cm 0cm 3cm, width=0.3\linewidth]{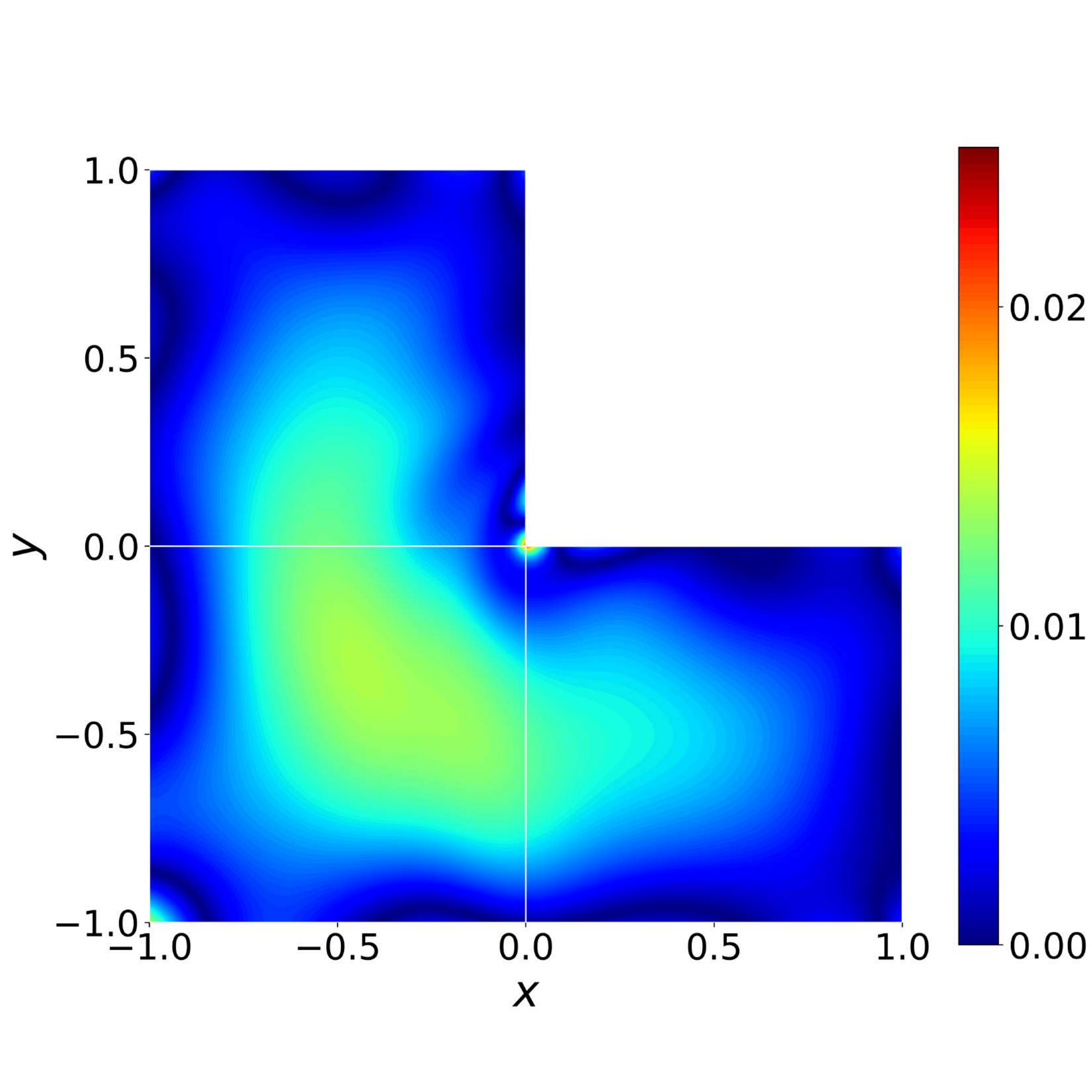}
		&
		\includegraphics[clip, trim=0cm 1cm 0cm 3cm, width=0.28\linewidth]{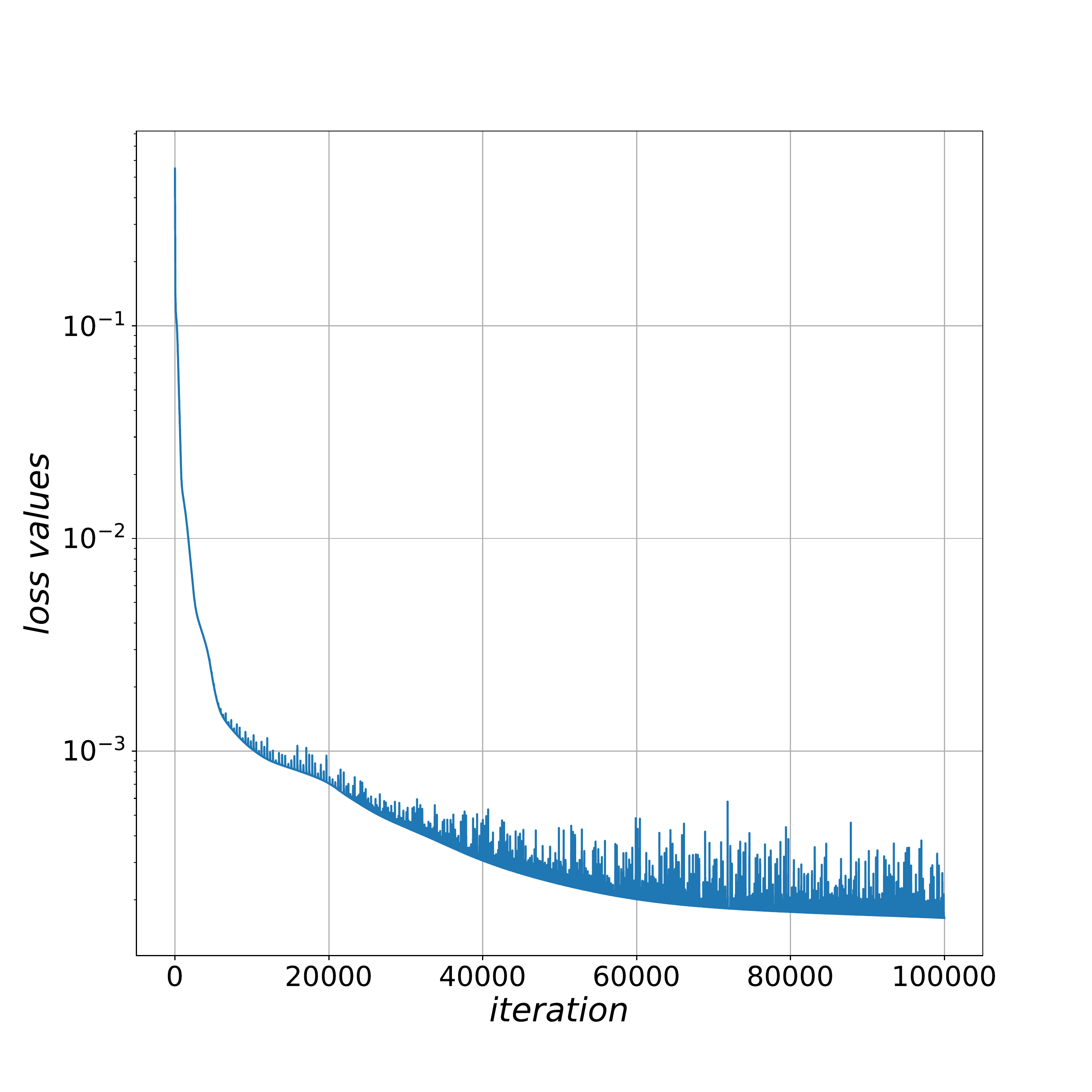}\\

		\includegraphics[clip, trim=0cm 0cm 0cm 0cm, width=0.24\linewidth]{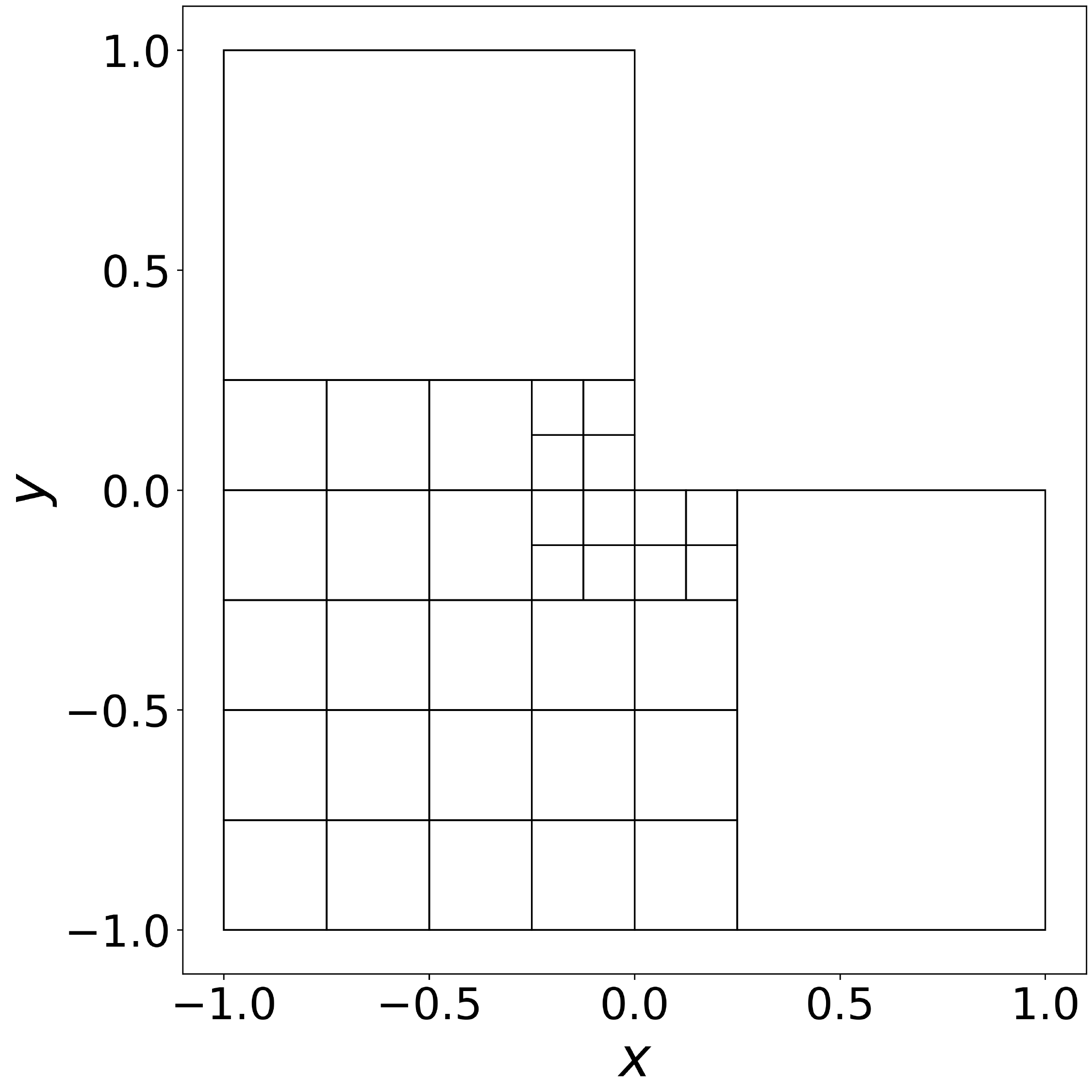}
		&
		\includegraphics[clip, trim=0cm 2cm 0cm 3cm, width=0.3\linewidth]{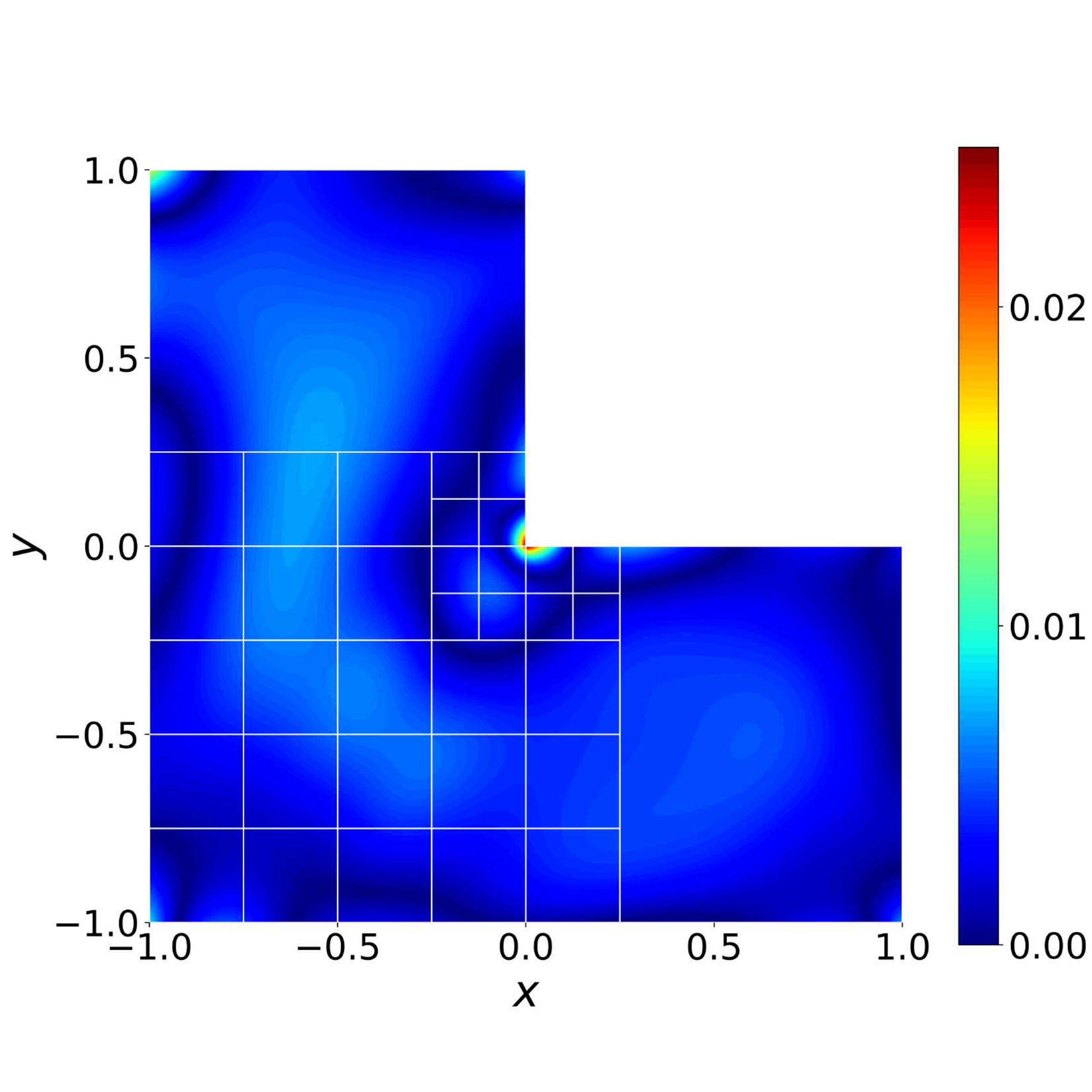}
		&
		\includegraphics[clip, trim=0cm 1cm 0cm 3cm, width=0.28\linewidth]{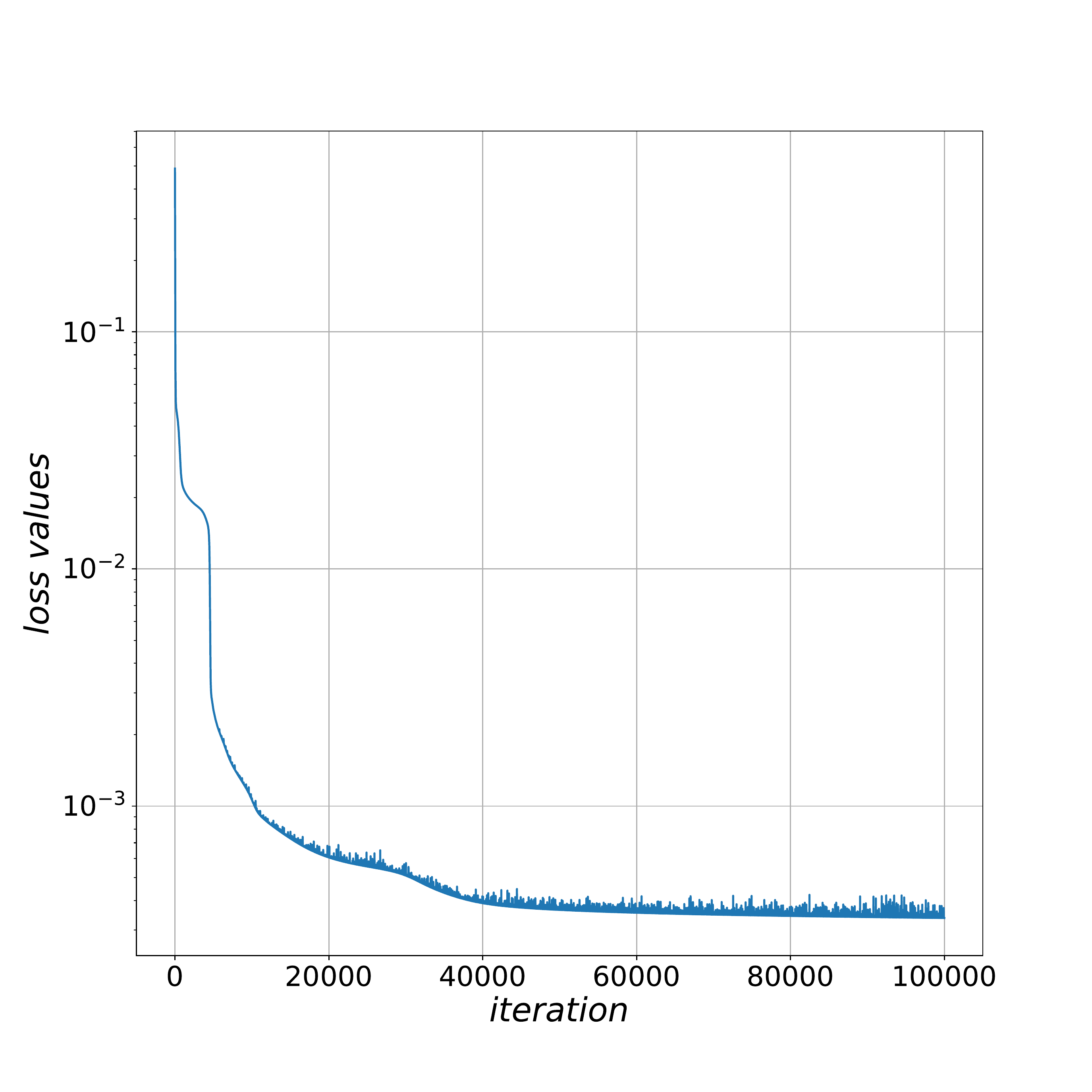}

	\end{tabular}
    \vspace{-0.15 in}
	\caption{\scriptsize \label{Fig: Poisson hom 2d Lshape vPINN} Two-dimensional homogeneous Poisson's equation in L-shaped domain. Top panel: (\textbf{A}) The reference solution $u_{\text{SEM}}$, and (\textbf{B}) PINN point-wiser error $|u_{NN} - u_{\text{SEM}}|$. Bottom panel: (\textbf{C} column) Adaptive \textit{h}-refinement via domain decomposition, (\textbf{D} column) \textit{hp}-VPINN point-wise error $|u_{NN} - u_{\text{SEM}}|$, and (\textbf{E} column) loss value versus training iterations. The \textit{hp}-VPINN is fully connected with $\ell = 3$, $\mathcal{N} = 5$, tanh activation function, and with parameters $\lbrace K_1 = K_2 = 5 , Q = 10 \times 10 \rbrace$ in each sub-domain (element) and $\tau_ b =10$. We use Adam optimizer with learning rate $10^{-3}$.
	}
\end{figure}

\vspace{0.2 in}
\begin{exm}
	We solve the homogeneous two-dimensional Poisson's equation, i.e. \eqref{Eq: 2-d Poisson} with $f(x,y) = 0$, over the L-shaped domain $\Omega $. The \textit{hp}-VPINN results are shown in Fig. \ref{Fig: Poisson hom 2d Lshape vPINN}. For comparison, we also present the numerical solution obtained by using the spectral element method (SEM) \cite{karniadakis2013spectral}. The solution and comparison of PINN with SEM is given in \cite{lu2019deepxde}.  
\end{exm}

%

The exact solution is not available in this case and thus we consider the SEM solution \cite{karniadakis2013spectral, lu2019deepxde} as a benchmark solution (the SEM uses total 12 equal elements with degree of polynomial $10 \times 10$ in x and y directions). The difficulty in this example is to accurately approximate the solution at the sharp edge $x = y = 0$. We see in \cite{lu2019deepxde} that the PINN formulation produces the largest error at this vertex while preserving better accuracy over the rest of the domain; it uses $N_r = 1200$ residual and $N_b = 120$ boundary points randomly drawn from a uniform distribution. We observe similar behavior in the \textit{hp}-VPINN formulation too, however, we report a better accuracy over the interior domain by refining the domain decomposition. In a coarse decomposition, we divide the domain into three sub-domains of equal sizes as shown in Fig. \ref{Fig: Poisson hom 2d Lshape vPINN}. Then, in a fine decomposition, we divide the domain into total of 35 sub-domains with different sizes. In both cases, we use 5 test functions in each direction $x$ and $y$, and employ $10 \times 10$ quadrature points in each sub-domain. We can see that in the fine domain decomposition, the interior domain error decreases, while the error at the sharp edge is still dominant. In all \textit{hp}-VPINN formulations, we use a fully connected network with $\ell = 3$, $\mathcal{N} = 20$, and tanh activation function. We also note that $\prescript{(1)}{}{\mathcal{R}} $ and $\prescript{(2)}{}{\mathcal{R}} $ formulations produce similar error level in this test case.

%
\section{Advection Diffusion Equation (Inverse and Forward Problems)}
\label{Sec: VPINN Diff examples}
%
Let $u(t, x): \Omega \rightarrow \mathbb{R}$, where $\Omega = [0,1] \times [-1,1]$. We consider the (1+1)-dimensional advection diffusion equation (ADE)
\begin{align}
\label{Eq: ADE}
&\frac{\partial u}{\partial t} + v \frac{\partial u}{\partial x} = \kappa \frac{\partial^2 u}{\partial x^2} , 
\\ \nonumber
&u(-1,t) = u(1,t) = 0,
\\ \nonumber
&u(x,0) = -\sin(\pi x),
\end{align}
where the constant coefficients $v = 1$ and $\kappa = 0.1/\pi$ are the advection velocity and the diffusivity coefficient, respectively. When the diffusion coefficients $\kappa$ is small the advection becomes dominant, which complicates the solution close to the right boundary $x = 1$ as the no slip boundary condition is imposed. The analytical solution of the ADE problem \eqref{Eq: ADE} is given in \cite{mojtabi2015one} in terms of infinite series summation. We use 800 number of terms to compute the analytical solution and compare with our proposed method.

\begin{figure}[!ht]
	\center
	\begin{tabular}{c c c}
	\multicolumn{1}{l}{}
	&\multicolumn{1}{l}{\quad \textbf{A}}
	&\multicolumn{1}{l}{}\\ [-1 pt]
	
	\multicolumn{3}{c}{
		\includegraphics[clip, trim=0cm 0cm 0cm 0cm, width=0.3\linewidth]{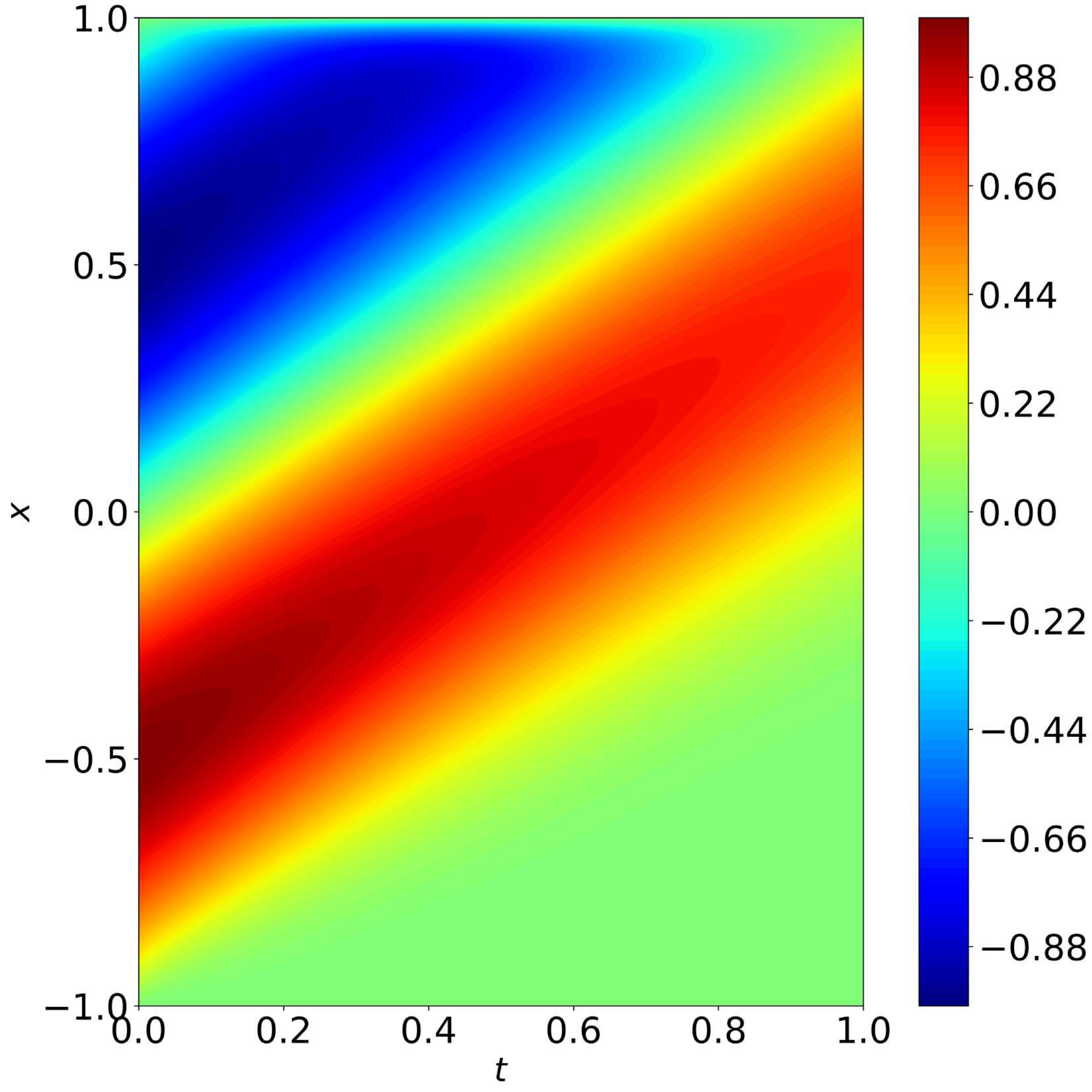}
	}
	
	\\
	\multicolumn{3}{c}{{\scriptsize \quad \textit{hp}-VPINN $\prescript{(1)}{}{\mathcal{R}} $ formulation: point-wise error}} \\  [-1 pt] 
	\hline \\ [-8 pt]
	
	\multicolumn{1}{l}{\quad \textbf{B}}
	&\multicolumn{1}{l}{\quad \textbf{C}}
	&\multicolumn{1}{l}{\quad \textbf{D}}\\ [-1 pt]
	
	\includegraphics[clip, trim=0cm 0cm 0cm 0cm, width=0.3\linewidth]{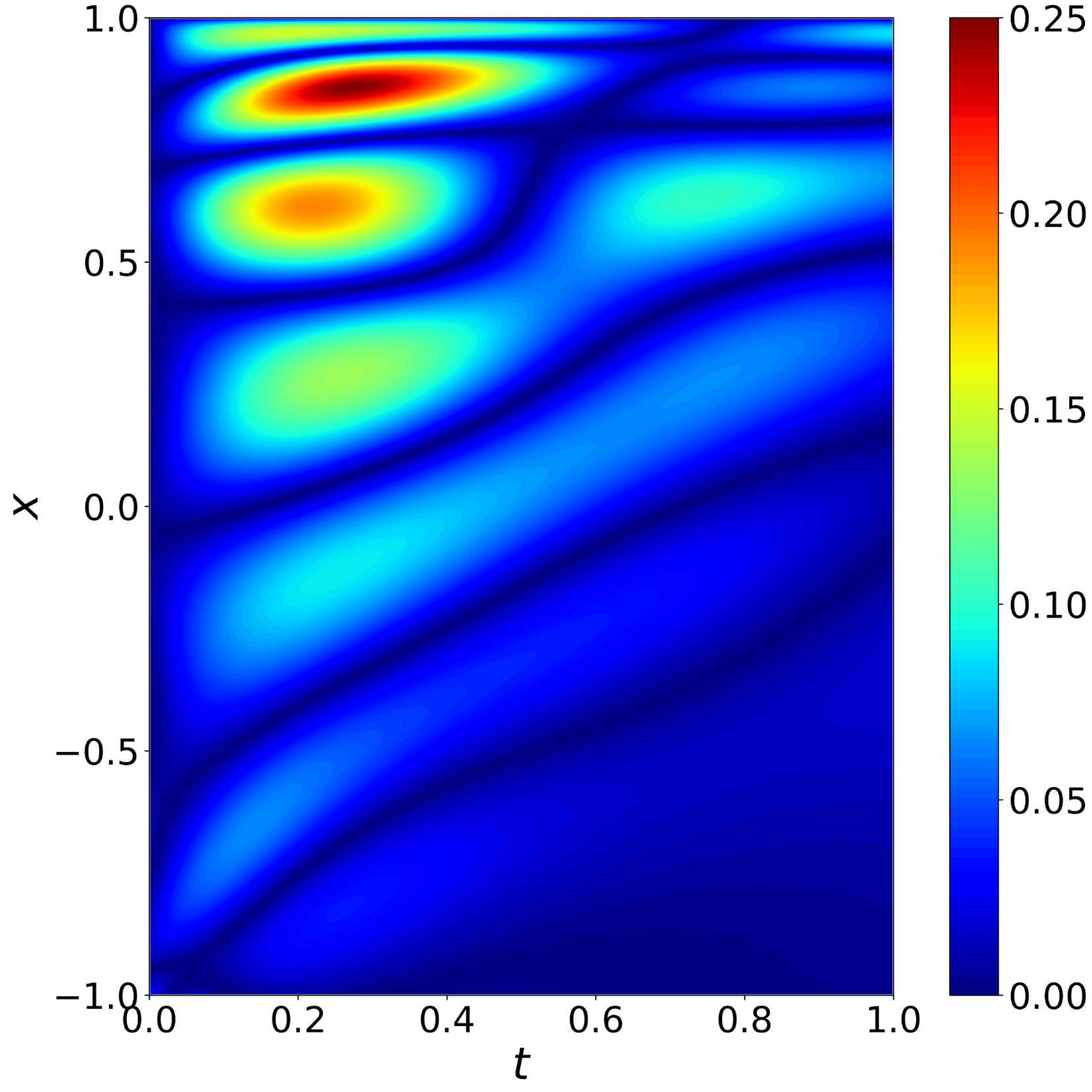}
	&
	\includegraphics[clip, trim=0cm 0cm 0cm 0cm, width=0.3\linewidth]{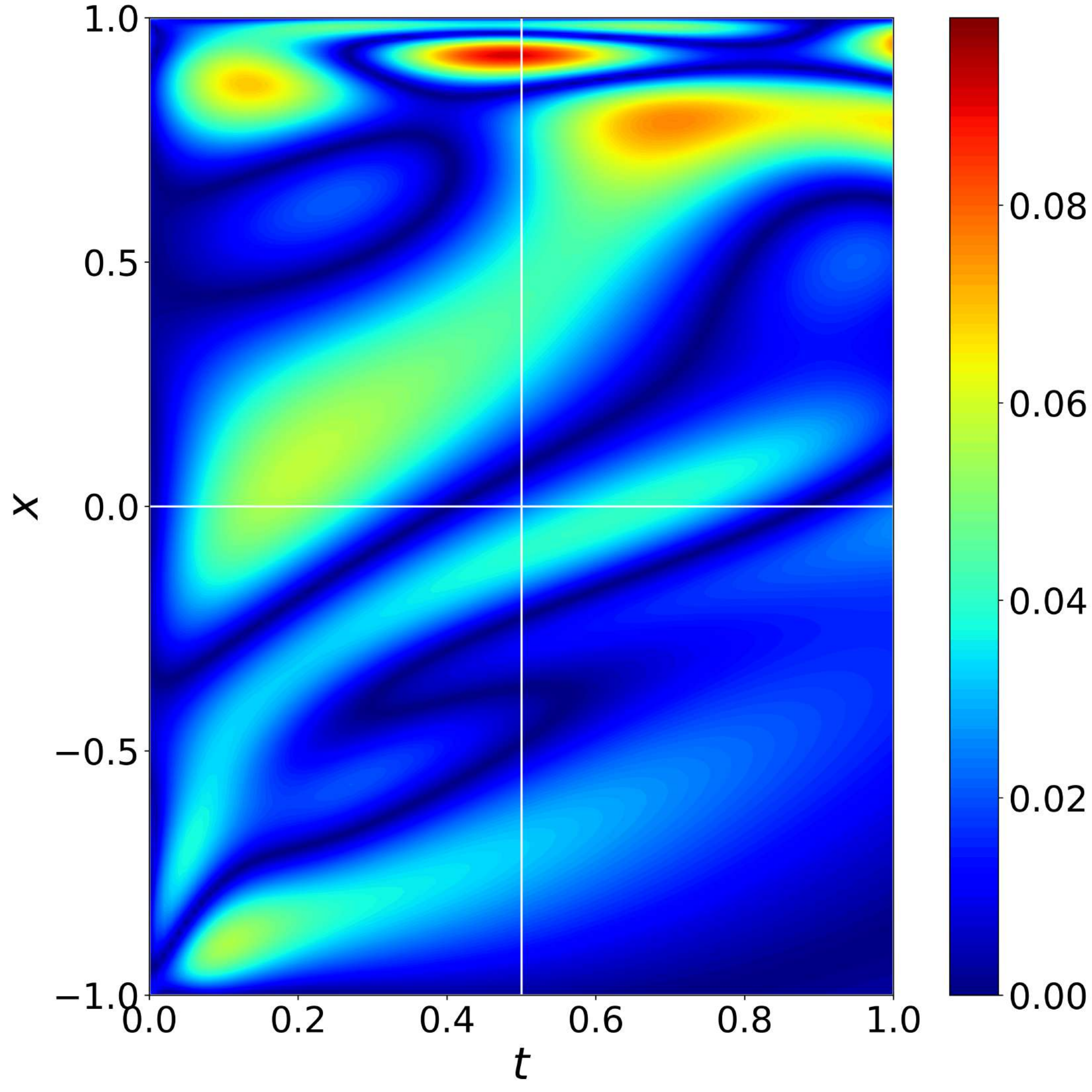}
	&
	\includegraphics[clip, trim=0cm 0cm 0cm 0cm, width=0.3\linewidth]{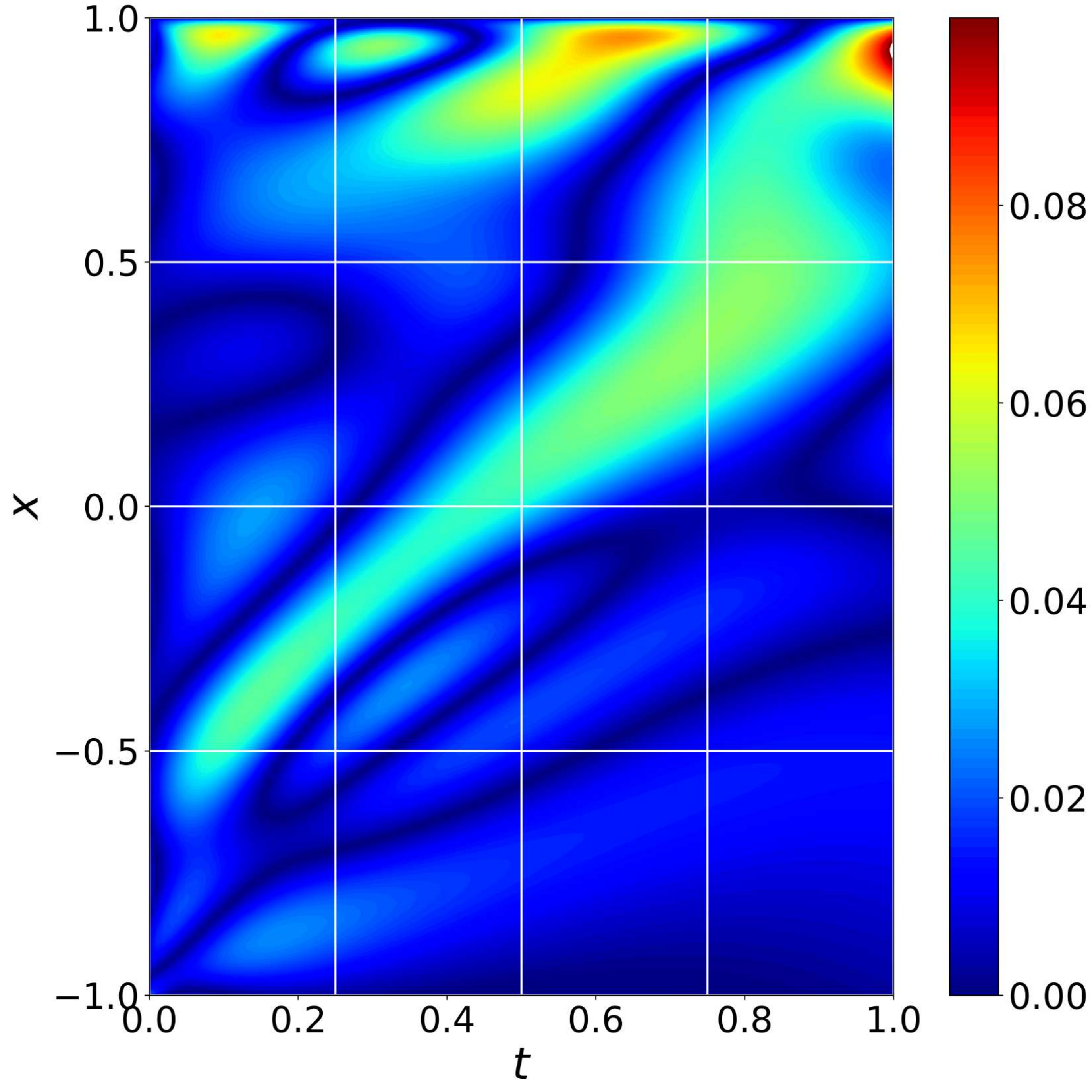}\\
	\end{tabular}
    \vspace{-0.15 in}
	\caption{\scriptsize \label{Fig: ADE vPINN} (1+1)-dimensional advection diffusion equation. (\textbf{A}) the exact solution. (\textbf{B}, \textbf{C}, and \textbf{D}) The \textit{h}-refinement of \textit{hp}-VPINN with $\prescript{(1)}{}{\mathcal{R}}$ formulation for $N_{el_t} = N_{el_x} = 1$, $2$, and $4$. In all cases, the network is fully connected with $\ell = 3$, $\mathcal{N} = 5$, and tanh activation function. The \textit{hp}-VPINN parameters are $\lbrace K_1 = K_2 = 5 , Q = 10 \times 10 \rbrace$ in each sub-domain (element), $N_b = 80$ boundary points, and $\tau_ b =10 $. We use Adam optimizer with learning rate $10^{-3}$.
	}
\end{figure}

We let the approximate solution be $u(t, x) \approx \tilde u(t, x) = u_{NN}(t, x)$ and note that in the transient problem, time can be thought of as another dimension and thus the formulation of variational residuals become similar to the previous examples. The space-time domain $\Omega$ is decomposed into $N_{el_t} \times N_{el_x}$ structured non-overlapping sub-domains (elements) $\Omega_{e_t e_x} = [t_{e_t-1},t_{e_t}] \times [x_{e_x-1}, x_{e_x}], \,\, e_t = 1,2,\cdots, N_{el_t}, \,\, e_x = 1,2,\cdots, N_{el_x}$ via constructing the temporal and spatial grids $\lbrace 0 = t_0, t_1, \cdots, t_{N_{el_t}} = 1   \rbrace$ and $\lbrace -1 = x_0, x_1, \cdots, x_{N_{el_x}} = 1   \rbrace$, respectively. Figure \ref{Fig: ADE vPINN} shows the \textit{h}-refinement of the \textit{hp}-VPINN method by considering different domain decompositions, i.e. $N_{el_t} = N_{el_x} = 1$, $2$, and $4$. The point-wise error is shown based on the $\prescript{(1)}{}{\mathcal{R}} $ formulation. We also report similar point-wise error for the $\prescript{(2)}{}{\mathcal{R}} $ formulation.

%

\noindent $\bullet$ \textbf{Diffusivity Estimation.}
The inverse problem is defined as: given a (sparse observation/measurement of) solution of mathematical model, one seeks to obtain an accurate estimation of the corresponding model parameters. 
It has been shown in \cite{raissi2017machine, raissi2018hidden} that the PINN formulation can incorporate the model parameters into the neural network parameters. This provides a framework to take the advantage of iterative methods in the context of neural network by letting the training algorithm to simultaneously optimize neural network and model parameters. By following a similar approach, we use VPINNs to solve the inverse problem for parameter estimation. We note that the VPINN formulation does not necessarily perform more accurately when compared to PINN formulation in solving the inverse problem; yet, we intent to show that it has this capability.

\begin{figure}[!ht]
	\center
	\begin{tabular}{c c c}
		%
    	\multicolumn{1}{l}{\quad \textbf{A}}
    	&\multicolumn{1}{l}{\quad \textbf{B}}
    	&\multicolumn{1}{l}{\quad \textbf{C}}\\ [-1 pt]
		\includegraphics[clip, trim=6cm 0cm 0cm 0cm, width=0.3\linewidth]{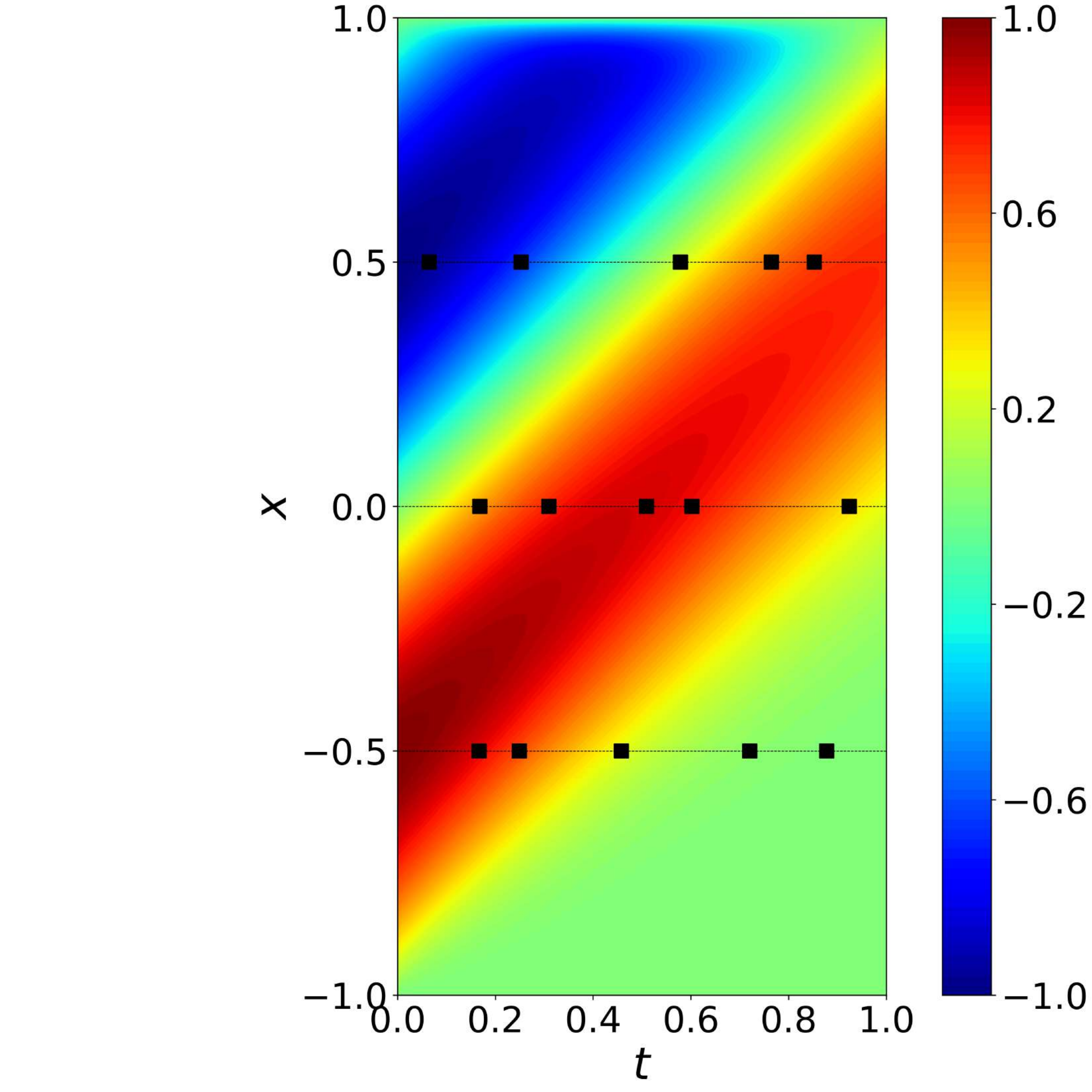}
		&
		\includegraphics[clip, trim=6cm 0cm 0cm 0cm, width=0.3\linewidth]{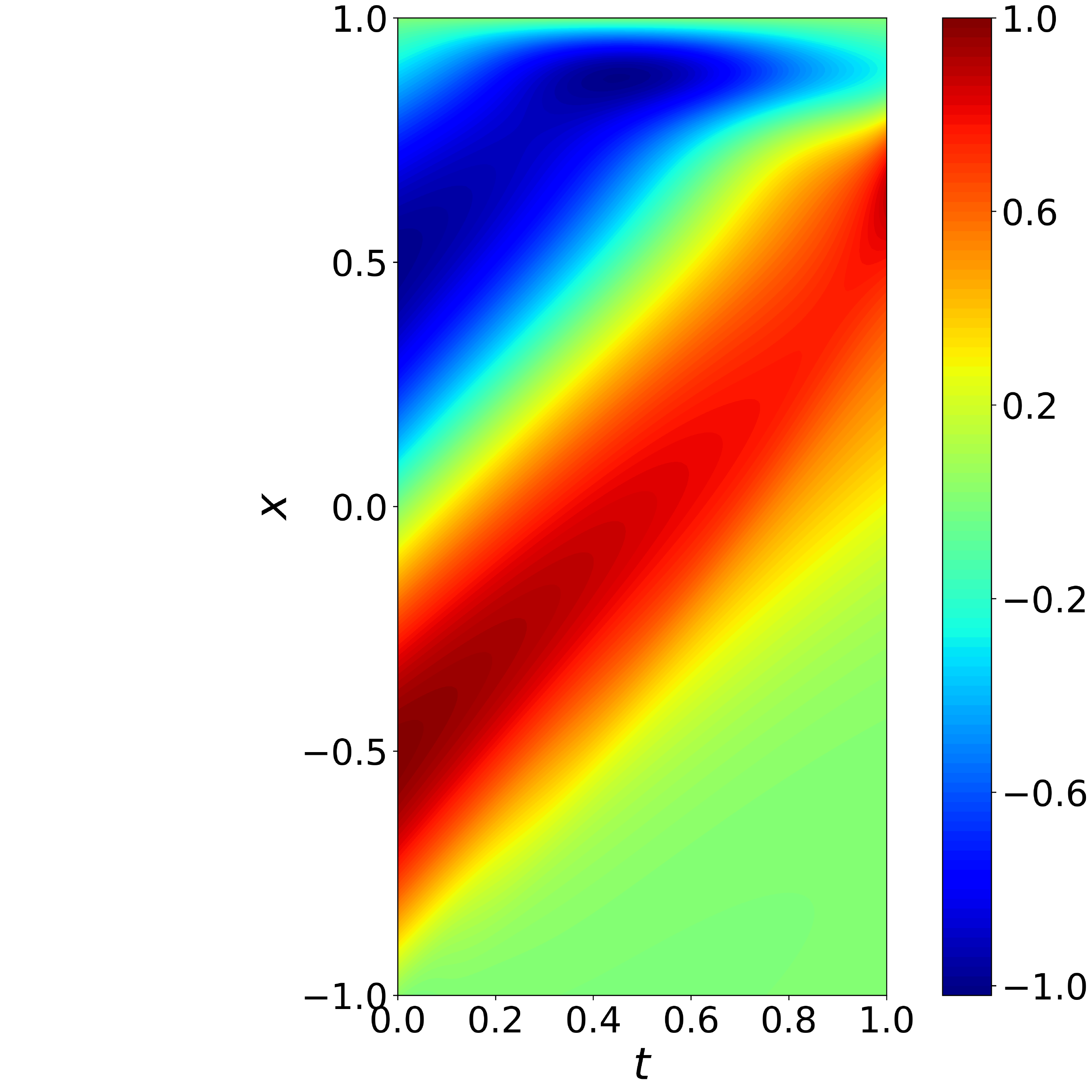}
		&
		\includegraphics[clip, trim=6cm 0cm 0cm 0cm, width=0.3\linewidth]{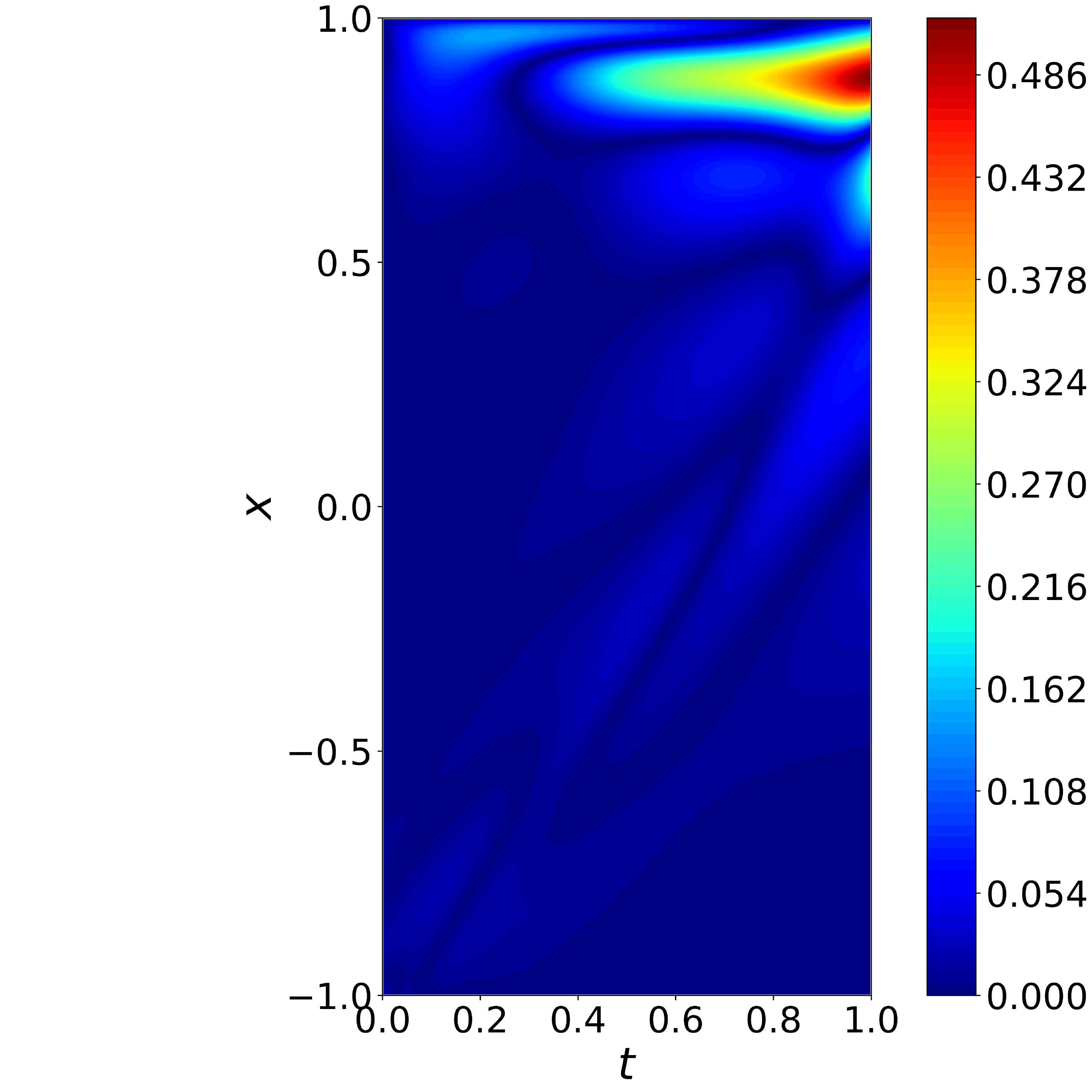}\\

    	\multicolumn{3}{l}{\quad\quad \textbf{D} \hspace{2 in}\qquad\qquad\qquad\, \textbf{E}}\\ [-3 pt]

		\multicolumn{3}{c}{
			\includegraphics[clip, trim=0cm 0cm 1cm 1cm, width=0.45\linewidth]{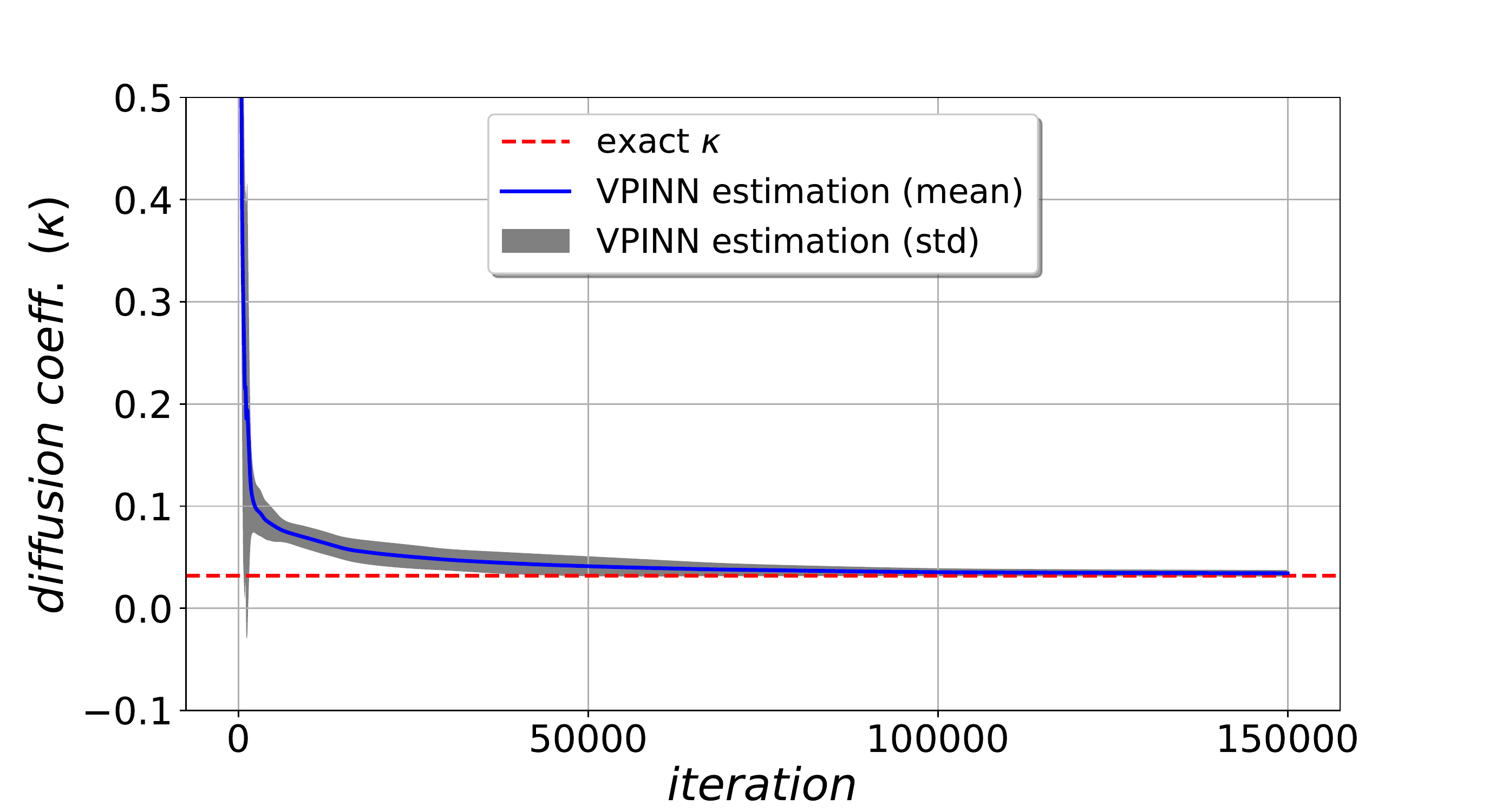}
			
			\includegraphics[clip, trim=0cm 0cm 1cm 1cm, width=0.45\linewidth]{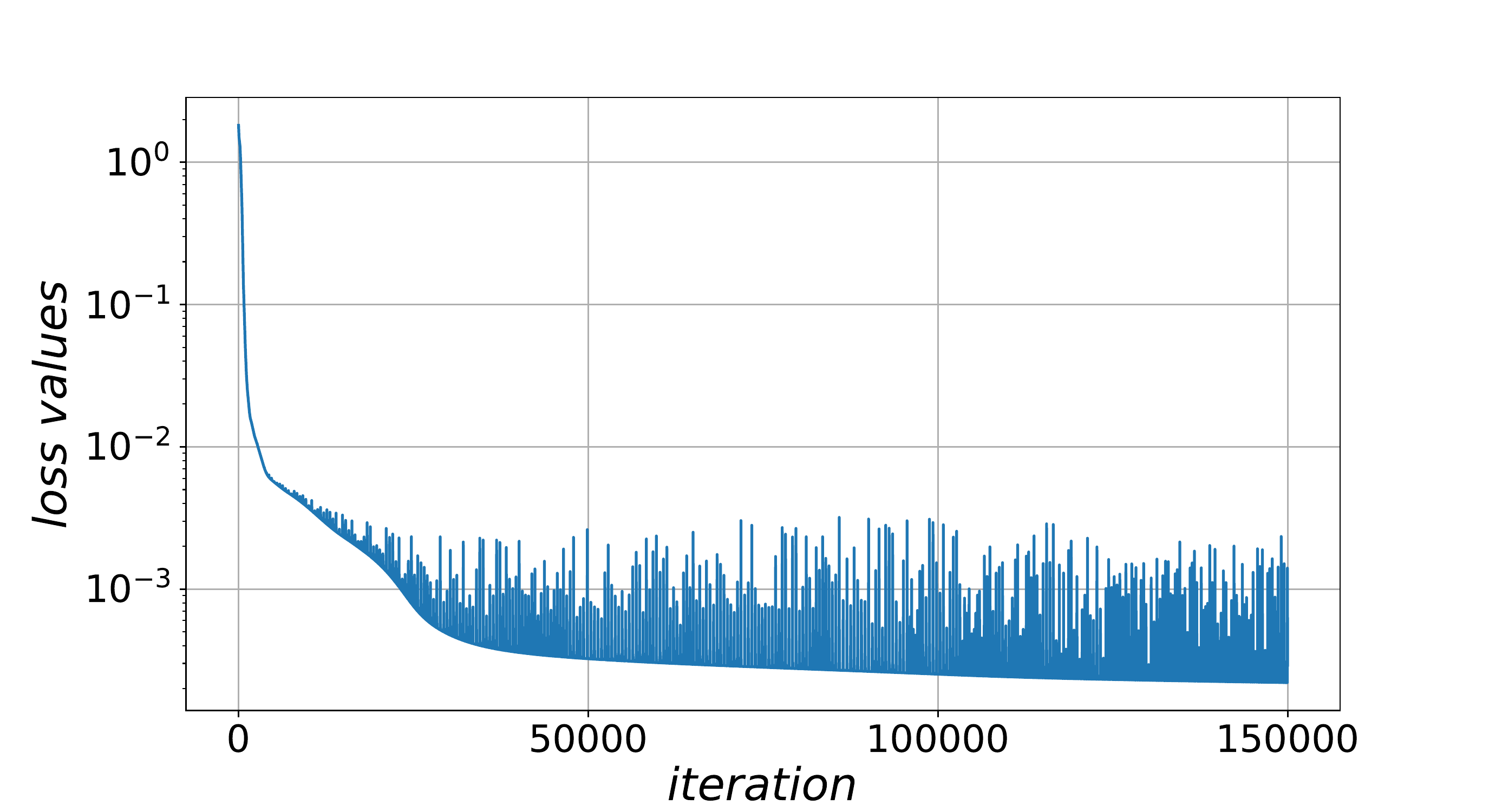}
		}
	\end{tabular}
    \vspace{-0.15 in}
	\caption{\scriptsize \label{Fig: ADE identification} Diffusivity estimation in (1+1)-dimensional ADE. (\textbf{A}) exact solution and one realization of $N^{\star} = 15$ randomly selected pints at the sensor locations. (\textbf{B}) VPINN prediction. (\textbf{C}) VPINN point-wise error. (\textbf{D}) convergence of diffusivity coefficient and (\textbf{E}) loss values versus training iterations. }
\end{figure}

\vspace{0.2 in}
\begin{exm}[Diffusivity Estimation]
	We consider the ADE \eqref{Eq: ADE} and let $\textbf{q} = \lbrace v, \kappa \rbrace$ be the set of model parameters, where the advection velocity is known to be a constant $v = 1$ and the diffusion coefficient $\kappa$ is unknown. Although in this case we have the analytical solution, we assume that the (observed/measured) values of exact solution $u^{\star}(t^{\star},x^{\star})$ is only available as time series at three (sensor) locations along the x axis, i.e., $x^{\star} = \lbrace -0.5, 0, 0.5 \rbrace$. We randomly select 5 data points at each sensor, and thus in total 15 measurements all over the whole domain; an example of these points is shown as black squares in Fig. \ref{Fig: ADE identification}. We pose the inverse problem diffusivity estimation as follows:
	\begin{align*}
	&\text{given the measurement set $\lbrace t^{\star}_i,x^{\star}_i, u^{\star}_i(t^{\star}_i,x^{\star}_i) \rbrace_{i=1}^{N^{\star}}$, estimate the diffusion coefficient $\kappa$}
	\\
	&\text{in ADE \eqref{Eq: ADE}}.  
	\end{align*}
	The results are shown in Fig. \ref{Fig: ADE identification}.
\end{exm}

The additional data points from given measurements/observations in the inverse problem are added as the following extra term in the variational loss function \eqref{Eq: loss weak - 1}
\begin{align}
\label{Eq: loss weak - inverse}
&
\tau^{\star} \, \frac{1}{N^{\star}} \sum_{i = 1}^{N^{\star}} \Big| u_{NN}(t^{\star}_i, x^{\star}_i) - u^{\star}(t^{\star}_i, x^{\star}_i) \Big|^2.
\end{align}
We use the VPINN formulation by constructing a fully connected neural network with tanh activation function, parameters $ \ell = 3, \mathcal{N} = 5, K_1 = K_2 = 5, Q = 10 \times 10, \tau_b = \tau_0 = \tau^{\star} = 10, N_b = 160, N_i = 80, N^{\star} = 15 $, and Legendre test functions in both space and time direction. We recall that the parameters $K_1$ and $ K_2$ are the number of test functions in space and time. We also note that here we use the $\prescript{(1)}{}{\mathcal{R}} $ formulation in the VPINN.

The unknown diffusivity coefficient $\kappa$ is initialized by one and as the network learns its parameter, the value of $\kappa$ converges to its exact value. The estimation is averaged over 10 different cases of randomly selected $N^{\star}$ points. The convergence of mean value of $\kappa$, its standard deviation, and also values of loss function are shown in Fig. \ref{Fig: ADE identification}. We observe that after convergence of diffusion coefficients, the point-wise error has only a large magnitude close to the right boundary at $t= 1$.

%

\section{Summary}
We developed the \textit{hp}-VPINN formulation to solve differential equations in the context of sub-domain Petrov-Galerkin method. The trial space is the space of neural networks and test space is the space of localized non-overlapping high order polynomials. We formulated the method in a general form, following the method of weighted residuals, where different choices of test functions lead to different numerical methods. We showed the efficiency and accuracy of \textit{hp}-VPINNs, compared to other methods, in several examples of function approximation and solving differential equations. We developed the method in detail for one- and two-dimensional problems and obtained the corresponding variational loss functions. We discussed the \textit{hp}-refinement and convergence of solution in solving equations with non-smooth solution. Moreover, we examined the efficiency of \textit{hp}-VPINNs in solving the inverse problem of parameter estimation in advection diffusion equation. For time-dependent problems and long-time integration, it may be more efficient to develop a discrete in time version of hp-VPINN as was done in \cite{raissi2019physics}.

\section*{Acknowledgement}
This work was supported by the Applied Mathematics Program within the Department of Energy on the PhILMs project (DE-SC0019453).

\bibliographystyle{elsarticle-num}
\bibliography{BIB_Temp}

\end{document}